\documentclass[final,1p,times,authoryear]{elsarticle}

\usepackage{amsmath}
\usepackage{amssymb}
\usepackage{graphicx}
\usepackage{tikz}
\usetikzlibrary{arrows.meta,positioning, calc}
\usepackage[hyphens]{url}
\usepackage{xurl} 
\usepackage[hidelinks]{hyperref}
\usepackage{orcidlink}
\usepackage{threeparttable}
\usepackage{booktabs}
\usepackage{array}
\usepackage{subcaption}
\usepackage[table]{xcolor}
\usepackage{tabularx}
\usepackage{multirow}
\newcolumntype{Y}{>{\raggedright\arraybackslash}X}
\definecolor{rowgray}{gray}{0.94}

\usepackage{adjustbox}
\journal{Ocean Engineering}

\begin{document}

\begin{frontmatter}

\title{Critic-Free Deep Reinforcement Learning for Maritime Coverage Path Planning on Irregular Hexagonal Grids}

\author[uai,ach]{Carlos S. Sepúlveda, \orcidlink{0000-0003-2426-7634}, \corref{cor1}}
\ead{carlos.sepulveda@alumnos.uai.cl}

\author[uai,sodas,phyto]{Gonzalo A. Ruz, \orcidlink{0000-0001-7740-9865}}
\ead{gonzalo.ruz@uai.cl}

\affiliation[uai]{organization={Facultad de Ingeniería y Ciencias, Universidad Adolfo Ibáñez},
                  addressline={Av. Diagonal las Torres 2640},
                  city={Santiago},
                  country={Chile}}

\affiliation[ach]{organization={Dirección de Programas, Investigación y Desarrollo, Armada de Chile},
                  addressline={General del Canto 398},
                  city={Valparaíso},
                  country={Chile}}

\affiliation[sodas]{organization={Millennium Nucleus for Social Data Science (SODAS)}, city={Santiago}, country={Chile}}

\affiliation[phyto]{organization={Millennium Nucleus in Data Science for Plant Resilience (PhytoLearning)}, city={Santiago}, country={Chile}}

\cortext[cor1]{Corresponding author.}

\begin{abstract}
Maritime surveillance missions, such as search and rescue and environmental monitoring, rely on the efficient allocation of sensing assets over vast and geometrically complex areas. Traditional Coverage Path Planning (CPP) approaches depend on decomposition techniques that struggle with irregular coastlines, islands, and exclusion zones, or require computationally expensive re-planning for every instance. We propose a Deep Reinforcement Learning (DRL) framework to solve CPP on hexagonal grid representations of irregular maritime areas. Unlike conventional methods, we formulate the problem as a neural combinatorial optimization task where a Transformer-based pointer policy autoregressively constructs coverage tours. To overcome the instability of value estimation in long-horizon routing problems, we implement a critic-free Group-Relative Policy Optimization (GRPO) scheme. This method estimates advantages through within-instance comparisons of sampled trajectories rather than relying on a value function. Experiments on 1{,}000 unseen synthetic maritime environments demonstrate that a trained policy achieves a 99.1\% Hamiltonian success rate, more than double the best heuristic (46.0\%), while producing paths 7\% shorter and up to 24.1\% fewer heading changes than the closest heuristic baseline. All three inference modes (greedy, stochastic sampling, and sampling with 2-opt refinement) operate under 50~ms per instance on a laptop GPU, confirming feasibility for real-time onboard deployment.
\end{abstract}


\end{frontmatter}

\section{Introduction}
\label{sec:intro}

Maritime surveillance (MS) underpins safety- and security-critical missions such as search and rescue (SAR), protection of critical infrastructure, prevention of illicit activities, and environmental monitoring \citep{Boraz2009}. Maritime transport carries the majority of global merchandise trade by volume, making the maritime domain a key component of national and international resilience \citep{united2024review}. Consequently, dense traffic, constrained chokepoints, and the strategic relevance of offshore infrastructure increase the operational demand for persistent Maritime Domain Awareness (MDA) \citep{bueger2024securing,bischof2018untangling}.

Achieving MDA typically relies on heterogeneous sensing assets whose capabilities must be allocated over large areas of interest (AOIs) with limited endurance. This operational requirement gives rise to planning problems formalized as Coverage Path Planning (CPP) \citep{choset2001coverage,Galceran2013,cabreira2019,Fevgas2022}. CPP seeks trajectories that systematically observe an AOI under constraints on motion, endurance, and obstacles.

However, maritime environments present specific geometric challenges that complicate classical CPP. Realistic AOIs are rarely convex polygons; they frequently involve irregular coastlines, exclusion zones, islands, and navigational hazards \citep{Nielsen2019, Mier2023}. Standard sweep patterns (e.g., boustrophedon paths) typically require the exact cellular decomposition of the target area into simpler, sweepable sub-regions \citep{choset2001coverage, Galceran2013}. This process relies heavily on the geometric morphology of the environment; irregular coastlines and dense exclusion zones generate numerous critical points that severely fragment the area. Consequently, routing a vehicle through these disjointed sub-regions becomes computationally intensive and kinematically inefficient, forcing excessive transit flights and sharp heading reversals \citep{cabreira2019, Nielsen2019}. Furthermore, exact optimization-based approaches such as Mixed-Integer Linear Programming (MILP) or evolutionary algorithms typically solve each instance from scratch \citep{Choi2019, Choi2020, Azad2017}. Recent USV-based bathymetric survey and cooperative coverage studies further confirm this pattern in coastal surveys, offshore mapping, island environments, and fault-tolerant multi-USV missions \citep{zhao2024optimal,zhao2024joint,shen2025multiple,ma2026fault}. This lack of generalization prevents the reuse of computational effort across different mission geometries, limiting their utility for rapid onboard replanning.

Reinforcement Learning (RL) offers a paradigm to learn policies that generalize across problem instances \citep{Sutton2018,Kaelbling1996}. In particular, attention-based Neural Combinatorial Optimization (NCO) has demonstrated that Transformer-based policies can learn to construct high-quality solutions for routing problems such as the Traveling Salesman Problem (TSP) \citep{Kool2018,berto2025rl4co}. This paradigm is highly compatible with CPP over discretized AOIs, where the agent must select a sequence of nodes to visit. We adopt a hexagonal discretization due to its favorable geometric properties, such as equidistant neighbors and reduced anisotropy compared to grid maps \citep{Boots1999,Kadioglu2019,Cho2021}.

In this work, we propose a Transformer-based pointer policy that operates on a graph representation of the hexagonal AOI. The policy constructs valid coverage tours by selecting feasible moves via dynamic action masking. A key innovation in our approach is the use of a critic-free Group-Relative Policy Optimization (GRPO) scheme. While Actor-Critic methods like Proximal Policy Optimization (PPO) are standard, learning an accurate value function for sparse-reward combinatorial problems remains challenging. GRPO stabilizes training by estimating advantages from within-instance comparisons of multiple sampled trajectories \citep{shao2024deepseekmath}, avoiding the bias and instability of a learned critic.

At a scientific level, this work addresses the problem of whether strict coverage routing on sparse, irregular maritime graphs can be learned as an amortized neural construction policy rather than repeatedly solved as a per-instance optimization problem. The central research question is: \emph{can a critic-free, group-relative training scheme learn to construct valid, single-visit, kinematically efficient coverage paths on geometrically irregular AOI graphs, while generalizing to unseen mission geometries without instance-specific re-solving?}

The main contributions of this work are:
\begin{itemize}
\item \textbf{Problem formulation.} We formulate maritime coverage path planning over geometrically irregular AOIs that may contain internal obstacles as a constrained Hamiltonian path problem on sensor-sized hexagonal grids. This formulation captures sparse local adjacency, strict single-visit coverage, path-length minimization, and kinematic turn costs, while avoiding the need for exact decomposition into sweepable subregions.

\item \textbf{Critic-free learning for maritime CPP.} We adapt Group-Relative Policy Optimization (GRPO) to long-horizon neural combinatorial coverage routing. Instead of learning a value-function critic, the proposed training scheme estimates \emph{outcome-level} relative advantages from groups of trajectories sampled on the same AOI instance, applied uniformly across the construction steps in line with the one-step MDP view of neural constructive routing \citep{berto2025rl4co}, addressing the instability of critic estimation under sparse terminal rewards and heterogeneous graph topologies.

\item \textbf{Feasibility-aware neural construction policy.} We design a Transformer-based pointer policy that constructs coverage paths autoregressively over sparse hexagonal graphs. Dynamic action masking enforces valid self-avoiding moves, while early dead-end detection provides sharper credit assignment by terminating trajectories when complete coverage becomes unreachable.

\item \textbf{Real-time generalization and benchmarking.} We empirically show that a single trained GRPO policy generalizes to 1{,}000 unseen maritime geometries, achieving 99.1\% Hamiltonian feasibility and producing shorter, smoother paths than a broad suite of classical coverage heuristics, without instance-specific retraining or per-instance optimization. We further benchmark the policy against a budgeted exact CP-SAT/MILP solver and a memetic genetic algorithm, showing that it attains near-optimal path quality at inference latencies roughly four orders of magnitude lower. The reported greedy, best-of-$K$, and 2-opt-refined inference modes quantify the latency--quality trade-off and demonstrate real-time onboard planning performance.
\end{itemize}

\paragraph{Impact}
From an operational perspective, the proposed framework supports real-time onboard mission planning and replanning for autonomous surface and aerial assets operating over irregular maritime AOIs. By producing single-visit routes with reduced path length and fewer heading changes, the method can reduce redundant transit and support a more efficient use of limited platform endurance, particularly in surveillance, search and rescue, and environmental monitoring missions. From a research perspective, this study shows that critic-free group-relative optimization, originally developed in the context of language-model reasoning, can be transferred to long-horizon, sparse-reward neural combinatorial optimization. The result is a reusable methodological bridge between attention-based routing policies, constrained graph coverage, and maritime CPP.

The remainder of the paper is organized as follows. Section~\ref{sec:related_work} reviews related work. Section~\ref{sec:problem} formalizes the problem. Section~\ref{sec:method} presents the proposed method and training scheme. Section~\ref{sec:experiments} describes the experimental setup, Section~\ref{sec:results} presents the results and discussion, and Section~\ref{sec:conclusion} concludes.

\section{Related Work}
\label{sec:related_work}

This section reviews the lines of work most relevant to our formulation: (i)~maritime surveillance and persistent monitoring, (ii)~coverage path planning and hexagonal tessellations, and (iii)~deep reinforcement learning for combinatorial routing. We close by positioning our approach within this landscape (visualized in Fig.~\ref{fig:taxonomy}).

\subsection{Maritime surveillance and persistent monitoring}

Maritime Domain Awareness (MDA) is a strategic requirement for safety and security missions, including search and rescue (SAR) and critical infrastructure protection \citep{Boraz2009, bueger2024securing}. Traditional surveillance relies on heterogeneous sensor constellations like coastal radars, Automatic Identification System (AIS), and space-based assets, whose data must be fused into a consistent situational picture \citep{Soldi2021, Dogancay2021}. As national-level policy frameworks increasingly mandate persistent monitoring of vast maritime zones \citep{programaocea}, the efficient scheduling and routing of mobile assets such as UAVs and USVs becomes operationally critical \citep{LiSavkin2021}.

From an optimization perspective, maritime surveillance has been modeled as routing and mission-planning problems since at least the late 1990s: SAR mission planning as TSP variants \citep{Panton1999, John2001}, regional platform routing for surface surveillance \citep{Grob2006}, and persistent aerial surveillance as mixed-integer linear programs (MILP) maximizing information gathered under endurance constraints \citep{Zuo2020}. Comprehensive surveys on UAV routing and trajectory optimization further highlight the complexity of integrating realistic vehicle dynamics, endurance constraints, and heterogeneous sensing objectives into tractable planning models \citep{Coutinho2018, Otto2018}. These contributions underline that, even before considering learning-based approaches, surveillance planning is inherently a high-dimensional combinatorial optimization problem.

\subsection{Coverage path planning and hexagonal tessellations}

CPP seeks a trajectory that visits every point in a region of interest while optimizing criteria such as path length or time \citep{Galceran2013}. In maritime and aerial robotics, standard sweep patterns (e.g., boustrophedon) require decomposing complex areas into simpler sub-regions \citep{Choi2019, Nielsen2019}, a process that is highly sensitive to the irregular morphology of coastlines and exclusion zones.

Hexagonal tessellations address this limitation: they minimize overlap and provide better isotropy than square grids, making motion costs direction-agnostic \citep{Boots1999, Azpurua2018}. Hexagonal decompositions have been formulated in MILP models for multi-UAV maritime SAR \citep{Cho2021, Cho2021b, Kadioglu2019}, and the existence of Hamiltonian circuits in hexagonal grid graphs has been studied theoretically \citep{islam2007hamilton}, establishing connectivity conditions directly relevant to complete-coverage feasibility. These results position hexagonal grids as a principled compromise between representational fidelity and algorithmic tractability.

Closely related are \emph{sweep} and \emph{barrier coverage} formulations, which focus on periodic traversal or boundary monitoring of target zones \citep{Li2011, Benahmed2019}. While historically addressed with static sensor deployments, these problems converge with CPP when mobile platforms (UAVs, USVs) are deployed to dynamically maintain coverage \citep{Li2019}.

A recent and closely related line of work targets USV-based coverage for bathymetric survey and cooperative multi-vehicle coverage. \citet{zhao2024optimal} combine coastline approximation with an evolutionary sequencing algorithm and validate it in real lake trials, while \citet{zhao2024joint} cast large-scale offshore survey coverage as an integer program solved by a hierarchical optimizer. At the fleet level, \citet{shen2025multiple} address cooperative coverage and task allocation among multiple unmanned sailboats in island environments, and \citet{ma2026fault} introduce fault tolerance for multi-USV coverage via cooperative-game task reallocation coupled with deterministic finite-state-machine control. These works confirm the operational relevance of maritime CPP, but they remain \emph{instance-specific}: each new AOI requires re-running a decomposition, integer program, or game-theoretic allocation from scratch. In contrast, we learn a single amortized policy that constructs feasible tours on unseen geometries without per-instance optimization, and we additionally benchmark against budgeted exact and evolutionary solvers (Section~\ref{subsec:baselines} and~\ref{subsec:exact_comparison}).

\subsection{Reinforcement learning and neural combinatorial optimization}

Deep Reinforcement Learning (DRL) has been applied to enhance barrier and sweep coverage with mobile UAVs cooperating with sensor networks \citep{Li2022}, and more broadly to local ship maneuvering and collision avoidance \citep{Gao2022}, UAV path optimization under threat \citep{Alpdemir2022}, and dynamic maritime CPP using grid-based Markov decision processes \citep{Ai2021, wu2024autonomous}. However, these controllers often rely on simple discretizations and small action spaces, struggling to scale to large, irregularly weighted grids.

A complementary paradigm is Neural Combinatorial Optimization (NCO), which treats DRL as a solver for routing problems. Starting with Pointer Networks \citep{Vinyals2015}, attention-based models have evolved into highly competitive solvers for TSP and Vehicle Routing Problems \citep{Kool2018, Nazari2018, berto2025rl4co}. These graph RL policies operate directly on graph-structured inputs, learning construction heuristics that match exact solvers in quality while operating orders of magnitude faster at inference \citep{darvariu2024graph}. Key advances include invalid-action masking for structured discrete spaces \citep{Huang2020}, and shared-instance baseline mechanisms such as POMO \citep{kwon2020pomo}, which drastically reduce gradient variance by comparing multiple rollouts from the same problem instance, an idea closely related to our critic-free GRPO training scheme \citep{shao2024deepseekmath}.

\subsection{Positioning of the present work}

While CPP has been extensively studied \citep{cabreira2019} and maritime surveillance heavily relies on routing models \citep{Zuo2020, Grob2006}, few works explicitly address large-scale maritime CPP on hexagonal grids using learning-based combinatorial solvers. Existing barrier and sweep heuristics lack DRL integration, and attention-based NCO has rarely been applied to coverage topologies with geometric obstacles and sparse adjacency constraints.

To make the positioning explicit, Table~\ref{tab:positioning} compares representative maritime CPP, optimization-based, DRL, and NCO approaches along six dimensions that define the scope of this study: hexagonal representation, irregular AOI handling, strict single-visit routing, amortized inference, explicit turn-cost modeling, and real-time execution. These dimensions separate the proposed contribution from both per-instance maritime planners and canonical neural routing policies.

\begin{table*}[!t]
\footnotesize
\centering
\setlength{\tabcolsep}{2.3pt}
\renewcommand{\arraystretch}{1.18}
\caption{Positioning of representative approaches to maritime and graph-based
coverage planning. \checkmark: explicitly addressed; $\sim$: partially
addressed or addressed under different assumptions; $\times$: not a primary
focus.}
\label{tab:positioning}
\begin{tabularx}{\textwidth}{@{}
Y
c
c
Y
c
c
c
c
@{}}
\toprule
\textbf{Approach (family / representative works)} &
\textbf{Hex} &
\shortstack{\textbf{Irreg.}\\\textbf{AOI}} &
\textbf{Main method} &
\shortstack{\textbf{Single}\\\textbf{visit}} &
\shortstack{\textbf{Amort.}\\\textbf{policy}} &
\shortstack{\textbf{Turn}\\\textbf{cost}} &
\shortstack{\textbf{RT}\\\textbf{infer.}} \\
\midrule

Classical CPP decomposition \citep{choset2001coverage,Galceran2013,cabreira2019} &
$\times$ & $\sim$ & Cellular decomposition / sweep &
$\times$ & $\times$ & $\sim$ & $\sim$ \\

Hex-based SAR / coverage optimization \citep{Cho2021,Cho2021b,Kadioglu2019} &
\checkmark & $\sim$ & Hex decomposition + MILP / graph optimization &
$\sim$ & $\times$ & $\times$ & $\times$ \\

USV coastal bathymetric survey \citep{zhao2024optimal} &
$\times$ & \checkmark & Coastline approximation + evolutionary sequencing &
$\times$ & $\times$ & $\sim$ & $\times$ \\

USV offshore bathymetric survey \citep{zhao2024joint} &
$\times$ & \checkmark & Integer programming / hierarchical planning &
$\times$ & $\times$ & $\sim$ & $\times$ \\

Multi-USV cooperative coverage \citep{shen2025multiple,ma2026fault} &
$\times$ & \checkmark & Task allocation / cooperative game + FSM &
$\times$ & $\times$ & $\sim$ & $\sim$ \\

Grid-MDP maritime CPP with DRL \citep{Ai2021,wu2024autonomous} &
$\times$ & $\sim$ & DRL over grid states &
$\times$ & $\sim$ & $\sim$ & $\sim$ \\

Attention-based NCO routing \citep{Kool2018,kwon2020pomo,berto2025rl4co} &
$\times$ & $\times$ & Neural construction policy for TSP/VRP &
\checkmark & \checkmark & $\times$ & \checkmark \\

\midrule
\textbf{This work} &
\checkmark & \checkmark & Critic-free DRL-NCO policy on sparse hex graphs &
\checkmark & \checkmark & \checkmark & \checkmark \\

\bottomrule
\end{tabularx}
\end{table*}

The comparison highlights two complementary gaps. First, attention-based NCO already provides single-visit routing, amortized policies, and fast inference, but has mainly been developed for canonical dense routing problems such as TSP and VRP rather than sparse, irregular maritime coverage graphs with obstacle-induced dead-end risk. Second, exact and metaheuristic maritime solvers can handle irregular operational areas and, in suitably constrained formulations, can enforce single-visit feasibility, but they solve each instance from scratch and therefore do not provide an amortized policy for real-time replanning. To the best of our knowledge, no prior family in this literature simultaneously addresses all six axes. The present work targets this gap by combining a hexagonal, irregular-AOI, single-visit, turn-aware CPP formulation with a critic-free amortized policy that performs real-time inference. The budgeted exact (CP-SAT/MILP) and evolutionary (memetic GA) solvers are therefore used as quantitative per-instance comparators in Section~\ref{subsec:exact_comparison}, rather than as amortized planning methods.

\begin{figure*}[!htb]
\centering
\resizebox{\textwidth}{!}{%
\begin{tikzpicture}[
  >=Latex,
  font=\footnotesize,
  node distance=2mm,
  box/.style={
    draw,
    rounded corners=2pt,
    align=center,
    inner xsep=3pt,
    inner ysep=3pt,
    line width=0.4pt
  },
  root/.style={box, fill=gray!15, text width=9cm, font=\footnotesize\bfseries},
  familyCov/.style={box, fill=blue!15,  text width=4.5cm, font=\footnotesize\bfseries},
  familyPat/.style={box, fill=green!15, text width=4.5cm, font=\footnotesize\bfseries},
  familyRout/.style={box, fill=orange!15,text width=4.5cm, font=\footnotesize\bfseries},
  covSub/.style={box, fill=blue!5,   text width=3.9cm},
  patSub/.style={box, fill=green!5,  text width=3.9cm},
  routSub/.style={box, fill=orange!5,text width=3.9cm},
  arrow/.style={->, line width=0.4pt}
]

\node[root] (root) at (0,0)
  {Surveillance-related\\coverage, patrolling and routing problems};

\node[familyCov]  (cov)  at (-5.0,-1.6) {Coverage of \\static regions};
\node[familyPat]  (pat)  at ( 0.0,-1.6) {Patrolling \&\\persistent surveillance};
\node[familyRout] (rout) at ( 5.0,-1.6) {Routing \&\\task assignment};

\coordinate (middle) at ($(root.south)+(0,-0.3)$);

\coordinate (covSplit)  at (cov.north  |- middle);
\coordinate (patSplit)  at (pat.north  |- middle);
\coordinate (routSplit) at (rout.north |- middle);

\draw[line width=0.4pt] (root.south) -- (middle);

\draw[line width=0.4pt] (covSplit) -- (routSplit);

\draw[arrow] (covSplit)  -- (cov.north);
\draw[arrow] (patSplit)  -- (pat.north);
\draw[arrow] (routSplit) -- (rout.north);

\node[covSub, below=3mm of cov] (cpp) 
  {\textit{Region coverage (CPP)}\\
   e.g., \citep{choset2001coverage,Galceran2013,cabreira2019,Fevgas2022,Kumar2023}};
\node[covSub, below=of cpp] (sweep)
  {\textit{Sweep coverage}\\
   e.g., \citep{Li2011,Gorain2014,Li2020,liu2021agent}};
\node[covSub, below=of sweep] (barrier)
  {\textit{Barrier coverage}\\
   e.g., \citep{Benahmed2019,Nguyen2018,Li2019,Kong2016}};
\node[covSub, below=of barrier] (wsn)
  {\textit{Sensor deployment \& WSN coverage}\\
   e.g., \citep{Li2011,Nguyen2018,Chen2015}};

\draw[arrow] (cov.south) -- (cpp.north);

\node[patSub, below=3mm of pat] (patcov)
  {\textit{Patrolling \& persistent\\area coverage}\\
   e.g., \citep{Nigam2009,Zelenka2020,Karapetyan2019,Savkin2019,Yanez2020,Luis2021b}};
\node[patSub, below=of patcov] (revisit)
  {\textit{Target / region revisit\\scheduling \& info-gain models}\\
   e.g., \citep{Zuo2020,Karasakal2016,Siew2022,siew2022cislunar}};

\draw[arrow] (pat.south) -- (patcov.north);

\node[routSub, below=3mm of rout] (vrp)
  {\textit{Routing (VRP/TSP/\\orienteering variants)}\\
   e.g., \citep{Panton1999,John2001,Grob2006,Coutinho2018,Otto2018,Cho2021,Cho2021b}};
\node[routSub, below=of vrp] (crowd)
  {\textit{Spatial crowdsourcing \&\\task assignment}\\
   e.g., \citep{Wu2019,Bhatti2021,Tong2020,ZhouZhen2019,Chen2020}};
\node[routSub, below=of crowd] (nco)
  {\textit{Learning-based CO\\(DRL, attention-based NCO)}\\
   e.g., \citep{Vinyals2015,Bello2016,Nazari2018,Kool2018,Xin2021,Li2021a,Li2021b,berto2025rl4co,darvariu2024graph}};

\draw[arrow] (rout.south) -- (vrp.north);

\end{tikzpicture}
}
\caption{Taxonomy of surveillance-related coverage, patrolling, and routing problems 
underpinning our CPP formulation on hexagonal grids. The three columns mirror 
the structure of the \textit{Related work} section, while Table~\ref{tab:taxonomy} provides a concise summary of the corresponding problem families.}
\label{fig:taxonomy}
\end{figure*}
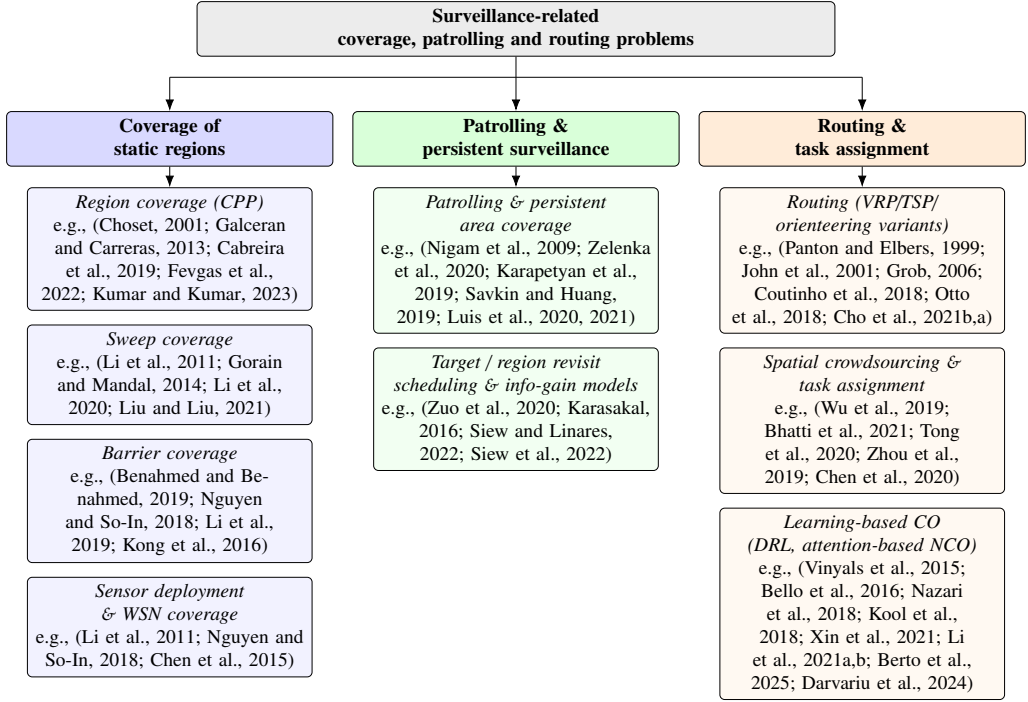

Fig.~\ref{fig:taxonomy} complements this positioning matrix by summarizing the broader taxonomy of coverage, patrolling, and routing problem families reviewed in this section. A detailed breakdown of objectives and representative literature per category is provided in Table~\ref{tab:taxonomy} (Appendix~\ref{app:taxonomy}).

\section{Problem Formulation}
\label{sec:problem}

\subsection{Area of Interest and Coverage}
\label{subsec:aoi_coverage_model}

Let the \emph{area of interest} (AOI) be a planar polygonal region
$\mathcal{P}\subset\mathbb{R}^2$ with a set of polygonal holes/obstacles
$\{\mathcal{H}_k\}_{k=1}^{K}$.
The feasible surveillance region is therefore
\begin{equation}
\mathcal{A} \;=\; \mathcal{P}\setminus\bigcup_{k=1}^{K}\mathcal{H}_k.
\end{equation}
We consider a single mobile sensing platform whose planar position at discrete time
$t\in\{0,\dots,T\}$ is $\mathbf{p}_t\in\mathcal{A}$.\footnote{The formulation naturally extends to multiple platforms via task partitioning; this extension is discussed in Section~\ref{sec:conclusion}.}

Coverage path planning (CPP) is defined as generating a trajectory that guarantees \emph{complete} or \emph{persistent} sensing \slash visiting of a region under platform and sensors constraints \citep{choset2001coverage, Galceran2013, cabreira2019, Fevgas2022, TanChee2021}.

In wide-area surveillance and reconnaissance, a standard abstraction is to model the instantaneous sensor footprint as a compact region around the platform position (often approximated by a disk), which aligns with classical geometric coverage models (e.g., unit-disk-type covering problems)
\cite{Biniaz2017}.

Accordingly, we model the effective footprint as a closed disk of radius $r_s>0$,
\begin{equation}
\mathcal{F}(\mathbf{p}_t) \;=\; \left\{\mathbf{q}\in\mathbb{R}^2:\|\mathbf{q}-\mathbf{p}_t\|_2\le r_s\right\}.
\end{equation}
A point $\mathbf{q}\in\mathcal{A}$ is covered by a trajectory $\{\mathbf{p}_t\}_{t=0}^{T}$
if $\mathbf{q}\in \bigcup_{t=0}^{T}\mathcal{F}(\mathbf{p}_t)$.
Directly optimizing trajectories over the continuous set $\mathcal{A}$ is generally difficult,
especially when $\mathcal{A}$ is non-convex and includes holes; therefore CPP pipelines typically adopt
cellular discretization to obtain a finite representation amenable to routing/optimization
\cite{choset2001coverage,Galceran2013,cabreira2019,Fevgas2022}. We adopt a hexagonal tessellation and graph construction pipeline detailed in Sections~\ref{subsec:hex_tessellation}--\ref{subsec:graph_representation}.

\subsection{Why Approximate Discretizations}
\label{subsec:why_approx}

CPP literature distinguishes \emph{exact} cellular decompositions (that preserve geometry with algorithmic guarantees)
from \emph{approximate} decompositions (grid/tessellation-based) \cite{choset2001coverage,Galceran2013,cabreira2019}.
Exact decompositions are attractive when one can exploit geometric structure, but they become cumbersome as
the AOI grows in complexity (multiple holes, narrow passages) and when additional operational constraints
must be incorporated (energy limits, risk maps, multi-vehicle coordination), often requiring repeated
re-computation and complicated bookkeeping \cite{cabreira2019,Fevgas2022,TanChee2021}.
Approximate discretizations provide a controllable resolution parameter (cell size) that trades geometric
fidelity for computational tractability, while enabling uniform neighborhood relations and direct mapping to
graph-based routing formulations \cite{cabreira2019,Fevgas2022,TanChee2021}.
This trade-off is particularly relevant in maritime surveillance and other large-area missions where scalability
and robustness are first-order requirements \cite{Grob2006,cabreira2019}.

\subsection{Hexagonal Tessellation of the AOI}
\label{subsec:hex_tessellation}

We discretize $\mathcal{A}$ by a hexagonal tessellation. In practice, we generate the grid in an oriented bounding box (OBB) frame to reduce boundary artifacts; the resulting tessellation-and-filtering sequence is illustrated in Fig.~\ref{fig:process_sequence}(a)--(c).

Let $\mathcal{G}_h$ denote a regular hexagonal grid with characteristic size (e.g., circumradius) $r_h$.
Hexagonal tessellations are a canonical class of spatial tessellations \cite{Boots1999} and have been used
for coverage and surveillance planning due to their near-isotropic adjacency and reduced directional bias compared
to square grids, which is beneficial when motion costs should be direction-agnostic \cite{Boots1999}.
Moreover, hexagonal decompositions have been explicitly adopted in UAV coverage and multi-UAV maritime SAR settings
\cite{Kadioglu2019,Azpurua2018,Cho2021}.

Beyond geometric convenience, hexagonal discretization is consistent with disk-like footprint abstractions:
when coverage is approximated by disks, the discretization should avoid anisotropies that can distort effective
coverage radii and neighborhood costs \cite{Biniaz2017,Boots1999}.
While the present work does not claim optimality of any single discretization, the above considerations motivate
a hexagonal grid as a principled compromise between representational fidelity and algorithmic simplicity
\cite{Boots1999,Cho2021,Kadioglu2019,Azpurua2018}. In this pipeline, the hexagon size is selected to be sensor-driven to align discretization resolution with the effective footprint. Specifically, the circumradius $r_h$ of each hexagonal cell is set equal to the sensor 
footprint radius $r_s$, so that a single visit to a cell centroid guarantees coverage of the 
entire cell area. Consequently, the number of cells $|\mathcal{V}|$ is determined not by the 
absolute area of the AOI but by the ratio $\text{Area}(\mathcal{A})/\text{Area}(c)$ between 
the feasible region and the individual cell area, making the graph size scale-invariant.

Let $\mathcal{C}=\{c_1,\dots,c_N\}$ be the set of hexagonal cells whose interiors intersect $\mathcal{A}$, and let $\mathbf{s}_i\in\mathbb{R}^2$ be the centroid of cell $c_i$.
Cells whose centroids fall inside holes are removed, yielding the feasible cell set
\begin{equation}
\mathcal{V} \;=\; \left\{ i\in\{1,\dots,N\} : \mathbf{s}_i\in \mathcal{A} \right\}.
\end{equation}
In practice, obstacles may be defined either as 
explicit polygonal holes $\mathcal{H}_k$ or, 
equivalently, by directly removing selected cells 
from $\mathcal{V}$. Both approaches yield the 
same graph $G$; the latter is used in our dataset 
generation for simplicity (Section~\ref{subsec:dataset}).

\paragraph{Discrete coverage.}
We say cell $c_i$ is \emph{covered} at time $t$ if the sensor footprint centered at $\mathbf{p}_t$
contains its centroid $\mathbf{s}_i$:
\begin{equation}
\label{eq:cell_covered}
\mathbf{1}\{c_i \text{ covered at } t\} \;=\; \mathbf{1}\left\{ \|\mathbf{s}_i-\mathbf{p}_t\|_2 \le r_s \right\}.
\end{equation}
Under the common discretization assumption $\mathbf{p}_t\in\{\mathbf{s}_i\}_{i\in\mathcal{V}}$ (platform moves between cell centroids),
coverage depends only on visited cells and can be represented combinatorially.

\subsection{Graph Representation}
\label{subsec:graph_representation}

\begin{figure*}[!htb]
\centering
\begin{subfigure}[t]{0.32\textwidth}
  \centering
  \includegraphics[width=\linewidth]{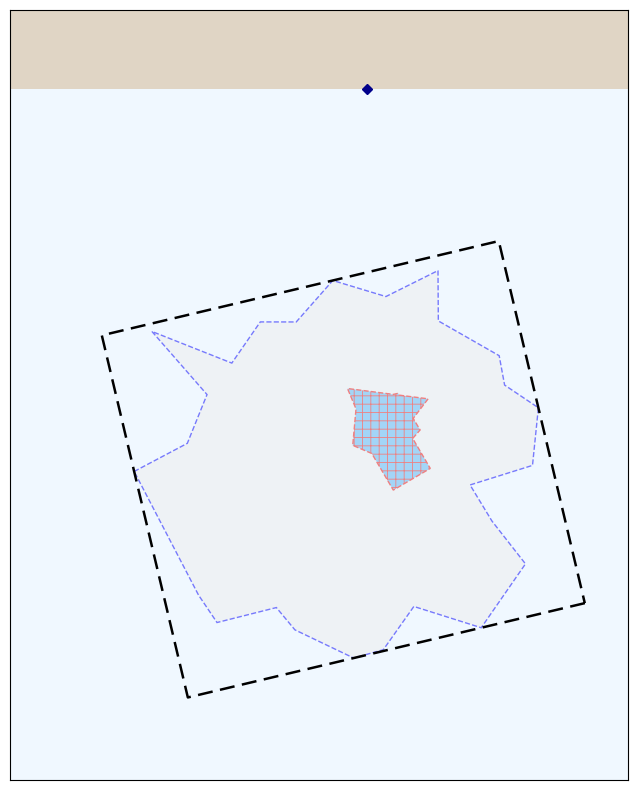}
  \caption{AOI, obstacle, and OBB.}
\end{subfigure}\hfill
\begin{subfigure}[t]{0.32\textwidth}
  \centering
  \includegraphics[width=\linewidth]{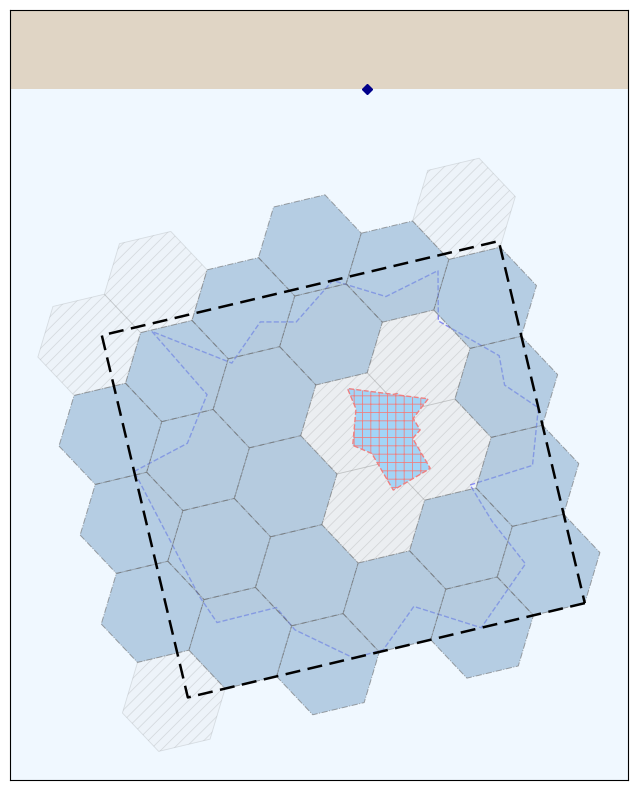}
  \caption{OBB-aligned hex grid (sensor-driven cell size).}
\end{subfigure}\hfill
\begin{subfigure}[t]{0.32\textwidth}
  \centering
  \includegraphics[width=\linewidth]{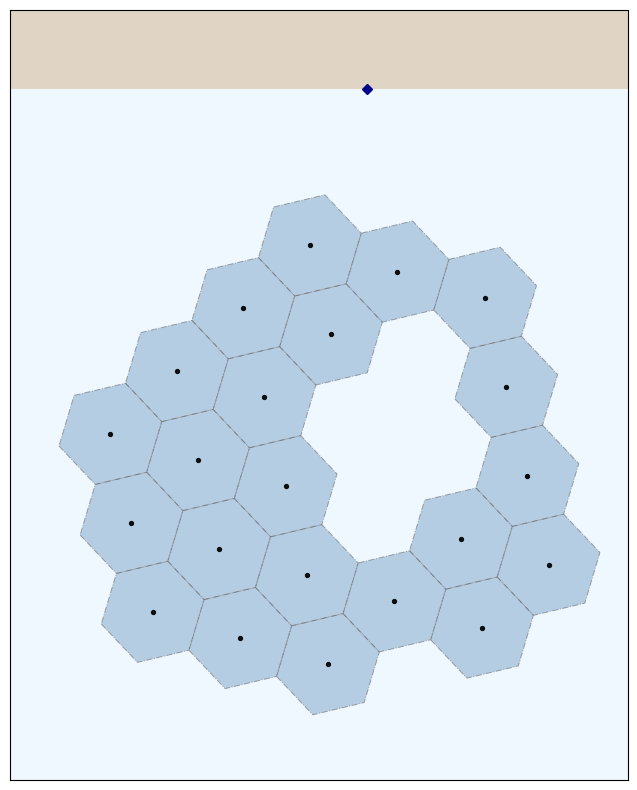}
  \caption{Visitable cells after AOI/obstacle filtering.}
\end{subfigure}

\vspace{4pt}
\begin{subfigure}[t]{0.49\textwidth}
  \centering
  \includegraphics[width=\linewidth]{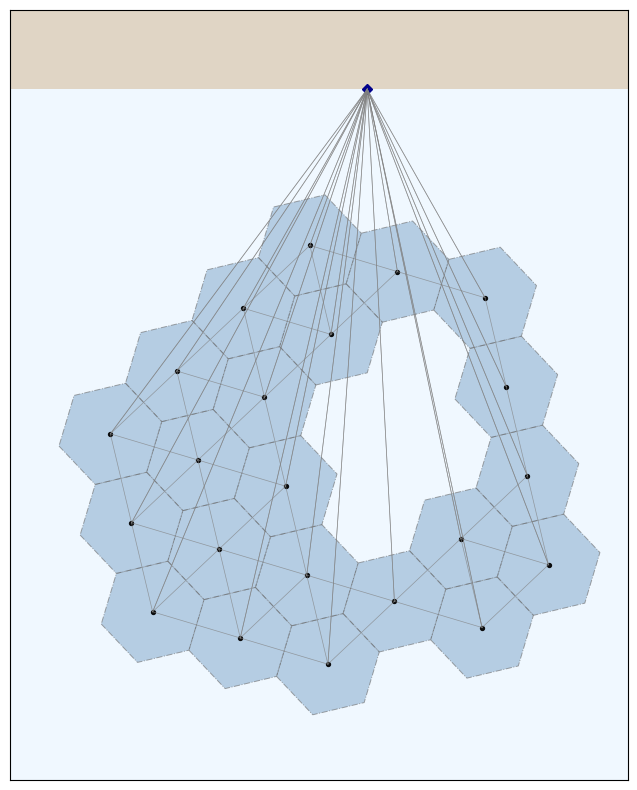}
  \caption{Adjacency graph from cell centers (incl.\ base node).}
\end{subfigure}\hfill
\begin{subfigure}[t]{0.49\textwidth}
  \centering
  \includegraphics[width=\linewidth]{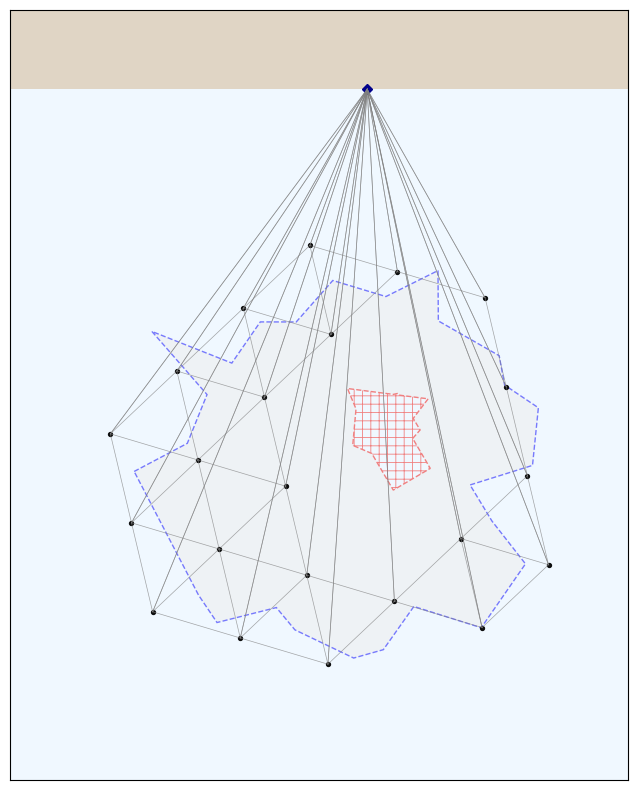}
  \caption{Final AOI graph for coverage planning.}
\end{subfigure}

\caption{Hexagonal tessellation-to-graph pipeline for an irregular AOI with an internal obstacle. Hexagon size is set by the sensor footprint. (a) AOI, obstacle, and OBB. (b) OBB-aligned hex grid. (c) Visitable cells selected by intersection tests. (d) Graph edges defined by hex-neighborhood adjacency using cell centers (including the base node). (e) Final graph used for coverage planning. (The pipeline is illustrated with an explicit polygonal obstacle for visual clarity; in the experimental dataset, obstacles are implemented as individual cell removals, see Section~\ref{subsec:dataset}.)}
\label{fig:process_sequence}
\end{figure*}

The discretized AOI naturally induces a graph.
Define an undirected adjacency relation $\mathcal{N}(i)$ over cells in $\mathcal{V}$ such that
$j\in\mathcal{N}(i)$ if cells $c_i$ and $c_j$ share an edge and the straight-line transition between
$\mathbf{s}_i$ and $\mathbf{s}_j$ does not intersect any hole.\footnote{Additional edge-level constraints (e.g., 
no-go zones, directional restrictions) can be incorporated by removing or redirecting edges; the graph abstraction remains unchanged.}
This yields a graph $G=(\mathcal{V},\mathcal{E})$ with edges
$\mathcal{E}=\{(i,j): j\in\mathcal{N}(i)\}$.
The edge construction from cell centers and the resulting AOI graph are depicted in Fig.~\ref{fig:process_sequence}(d)--(e).
Let $c_{ij}\ge 0$ be the travel cost associated with moving from $i$ to $j$ (e.g., distance, time, or a risk-weighted metric).
Graph abstractions of this form are standard in navigation and enable the use of graph-based reinforcement learning
and combinatorial optimization techniques \cite{darvariu2024graph,zweig2020neural}.

A discrete path is a sequence $\pi=(v_0,v_1,\dots,v_T)$ with $v_t\in\mathcal{V}$ and $(v_t,v_{t+1})\in\mathcal{E}$.
Let $v_0=b$ denote a designated base/initial cell (and optionally $v_T=b$ for return-to-base missions).

The base node $b$ is positioned outside the tessellated AOI and connected to every cell on the outer ring of $\mathcal{V}$ whose straight-line segment $\overline{b\,\mathbf{s}_i}$ does not intersect any obstacle. To represent the end of the mission, a duplicate terminal node $b'$ is introduced, which in our current closed-loop experiments shares the exact same adjacency set. Architecturally, explicitly decoupling the departure and arrival nodes enables the framework to natively support open-path missions (e.g., launching from a mother ship and recovering at a distinct coastal base) by simply assigning distinct spatial coordinates and adjacency masks to $b'$, without requiring any modifications to the underlying neural policy. Finally, this line-of-sight constraint ensures that both departure and return transits safely avoid overflying exclusion zones.

\subsection{Combinatorial Optimization View: Relation to TSP/VRP}
\label{subsec:tsp_relation}

When coverage is enforced by visiting a set of discrete locations, CPP becomes closely related to routing problems such as TSP, vehicle routing problems (VRP), and their generalizations \citep{bahnemann2021revisiting, Raza2022}. In the simplest complete-coverage case, one seeks a minimum-cost route that visits all required cells, which mirrors a Hamiltonian-path structure on the AOI graph.

However, the grid-based maritime CPP addressed here differs from the classical TSP in three operationally significant ways:
(i)~\emph{sparse feasibility}: movements are restricted to local hexagonal adjacency rather than a fully connected graph, and irregular obstacle configurations can render certain node orderings infeasible;
(ii)~\emph{kinematic costs}: heading changes incur 
maneuvering penalties that break the symmetry of Euclidean 
edge weights, coupling the cost of visiting a node to the 
direction of approach; and
(iii)~\emph{dead-end risk}: unlike the classical TSP where 
every permutation of nodes constitutes a valid tour, the 
sparse and irregular topology of hex grids with obstacles 
means that certain visitation sequences lead to 
unrecoverable states from which complete coverage is 
impossible, making future feasibility non-trivial to assess without lookahead mechanisms.

Note that, by design, the hexagonal cell size is matched to 
the sensor footprint radius~$r_s$ 
(Section~\ref{subsec:hex_tessellation}), so that complete 
area coverage requires visiting every cell 
in~$\mathcal{V}$. This deliberately reduces the problem to a 
constrained Hamiltonian-path formulation on~$G$, avoiding the 
additional complexity of partial-coverage models where 
sensing range exceeds cell size. Extensions to non-uniform 
coverage priorities and multi-vehicle coordination, while 
operationally relevant, are beyond the scope of this work 
and are discussed in Section~\ref{sec:conclusion}.

\subsection{Markov Decision Process Formulation}
\label{subsec:mdp}

While the CPP problem can be naturally formulated as a MILP targeting the Hamiltonian path, the underlying routing constraints are NP-hard and inherit the combinatorial explosion of the TSP/CVRP family, so exact solution cost grows steeply with the number of nodes \citep{Cho2021, Zuo2020, zhao2024joint}. While a budgeted CP-SAT solver remains feasible at the scale considered here $(|\mathcal{V}|\in[28,46])$, it requires seconds to minutes per instance and proves optimality on only a minority within an operational budget (Section~\ref{subsec:exact_comparison}); for larger grids it rapidly becomes prohibitive. In maritime scenarios requiring rapid replanning, relying on per-instance exact optimization is therefore operationally unfeasible.

We therefore cast coverage path planning on the AOI graph $G=(\mathcal{V},\mathcal{E})$ as a finite-horizon MDP solved by DRL.

\paragraph{State.}
Each node $i\in\mathcal{V}$ carries geometric features (normalized centroid coordinates) and a nonnegative priority weight $w_i$ (\emph{hexscore}). The state at step~$t$ comprises: (i)~the current node $v_t$, (ii)~the set of already visited nodes $\mathcal{S}_t\subseteq\mathcal{V}$, represented as a binary mask, and (iii)~the static instance data $(G,\{w_i\})$. This state representation allows the MDP formulation to natively support both uniform coverage path planning ($w_i = 0$ for all $i$) and priority-weighted coverage ($w_i \ge 0$ drawn from a spatial distribution). While the neural architecture fully integrates these priority maps, the experiments in this paper isolate the geometric routing challenge by focusing exclusively on the uniform coverage case; heterogeneous priority fields are deferred to future work.

\paragraph{Actions and feasibility masking.}
The action $a_t$ selects the next node $v_{t+1}$ among the unvisited feasible neighbors of $v_t$:
\begin{equation}
a_t \in \mathcal{A}(s_t) \;=\; \left\{ j \in \mathcal{N}(v_t) \;:\; j\notin\mathcal{S}_t \right\},
\end{equation}
where $\mathcal{N}(v_t)$ denotes the graph neighborhood of $v_t$. Invalid actions are prevented by applying an additive mask to the policy logits before the softmax activation. This technique has been shown to preserve valid policy gradients while strictly outperforming penalty-based constraints in discrete action spaces \citep{Huang2020}. Crucially, our mask enforces two strict topological rules: (1) it assigns $-\infty$ to non-neighbors and already-visited nodes to guarantee dynamically feasible, self-avoiding paths, and (2) it explicitly masks the terminal base node until 100\% of the target cells have been visited. This terminal-masking prevents the agent from exploiting early-return behaviors to minimize kinematic costs, forcing it to navigate the entire region before it can extract the episodic completion reward.

\paragraph{Terminal conditions.}
An episode terminates in one of two ways: (i)~\emph{successful completion}, when the agent has visited all target cells and returned to the base node, receiving a large positive reward; or (ii)~\emph{dead-end failure}, when no unvisited neighbor is reachable, in which case a penalty is applied. This latter condition is common in irregular hex grids where narrow passages and obstacles create geometric choke points. We address it with a Breadth-First Search (BFS) look-ahead mechanism described in Section~\ref{subsec:deadend}.

\paragraph{Reward function.}
The total return of a trajectory $\pi$ of length $T$ is:
\begin{equation}
R(\pi) \;=\; \sum_{t=0}^{T-1} \Big( r_{step}^{(t)} + r_{hex}^{(t)} - c_{dist}^{(t)} - c_{turn}^{(t)} \Big) \;+\; R_{episodic},
\label{eq:return}
\end{equation}
where the dense components are evaluated at each transition $(v_t, v_{t+1})$, and $R_{episodic}$ is a terminal modifier. The reward components, calibrated to prevent degenerate behaviors such as premature termination or redundant looping, are summarized in Table~\ref{tab:rewards}.

\begin{table}[!htb]
\small
\centering
\caption{Dense and episodic components of the reward function.}
\label{tab:rewards}
\begin{threeparttable}
\begin{tabularx}{\linewidth}{@{} l c l Y @{}}
\toprule
\textbf{Component} & \textbf{Value} & \textbf{Type} & \textbf{Operational justification} \\
\midrule
$r_{step}$ & $+2.0$ & Dense & Incentive per newly visited hexagonal cell. \\
$r_{hex}$ & $+0.5 \cdot w_{v_t}$ & Dense & Time-decayed priority bonus for early visitation of high-value zones (inactive in uniform-coverage experiments)\textsuperscript{$\dagger$}. \\
$c_{dist}$ & $-1.0 \cdot \bar{d}$ & Dense & Penalty proportional to normalized inter-cell distance. \\
$c_{turn}$ & $-0.25 \cdot f(\theta)$ & Dense & Kinematic penalty for sharp heading changes, promoting smooth trajectories. \\
$r_{complete}$ & $+100.0$ & Episodic & Sparse reward for full coverage and return to base. \\
$r_{death}$ & $-40.0$ & Episodic & Penalty for falling into an unrecoverable geometric dead-end. \\
\bottomrule
\end{tabularx}
\begin{tablenotes}
\item[\textsuperscript{$\dagger$}] The hexscore channel $w_i$ is architecturally supported but set to zero in all reported experiments (Section~\ref{subsec:dataset}).
\end{tablenotes}
\end{threeparttable}
\end{table}

Here, the normalized distance $\bar{d}_t = \|\mathbf{s}_{v_t} - \mathbf{s}_{v_{t+1}}\|_2 \cdot \sqrt{|\mathcal{V}|}$ scales the inter-cell Euclidean distance by a density factor that normalizes the penalty across instances of varying size, ensuring that larger grids do not trivially dominate the cost structure. The turn penalty $f(\theta_t)$ is a composite function of the heading change angle $\theta_t\in[0,\pi]$ between consecutive movement vectors. To reflect the physical reality of maritime vehicles, where initiating a course change involves a fixed mechanical overhead (e.g., rudder actuation) plus an angle-dependent hydrodynamic resistance, the penalty combines a discrete activation cost and a quadratic magnitude cost:
\begin{equation}
f(\theta_t) = \begin{cases}
2\left[ \left(\frac{\theta_t}{\pi}\right)^{\!2} 
+ c_{base} \right] & \theta_t > 0,\\
0 & \theta_t = 0,
\end{cases}
\label{eq:turn_penalty}
\end{equation}
where $c_{base}$ is a constant maneuvering penalty. In our implementation, we empirically calibrate $c_{base} = 1/12$. This specific value ensures that for the minimum required course correction on a hexagonal grid ($60^\circ$), the fixed mechanical overhead ($1/12 \approx 0.083$) and the quadratic dynamic resistance ($(1/3)^2 \approx 0.111$) are of comparable magnitude, appropriately reflecting the high inertia and rudder-shift costs of maritime platforms. For larger maneuvers, the quadratic term naturally dominates. Because the minimum non-zero heading change is $60^\circ$, every maneuver robustly activates this full penalty structure. Straight-line motion ($\theta_t = 0$) and the first step of each episode incur no turn cost. Consequently, the fixed term heavily penalizes the total \emph{number} of turns, while the quadratic term strictly limits sharp reversals, explicitly discouraging zigzagging and promoting dynamically smooth trajectories.

The objective is to learn a stochastic policy $\pi_\theta(a_t\mid s_t)$ that maximizes the expected return $\max_\theta\; \mathbb{E}_{\pi_\theta}[R(\pi)]$ over a distribution of irregular AOI geometries.

\section{Proposed Method}
\label{sec:method}

This section describes the three components of our approach: (i)~a Transformer-based pointer policy that constructs coverage tours autoregressively (Section~\ref{subsec:architecture}), (ii)~a critic-free GRPO training scheme that avoids learning a value function (Section~\ref{subsec:grpo}), and (iii)~an early dead-end detection mechanism that accelerates convergence on irregular grids (Section~\ref{subsec:deadend}). We close with a description of the training procedure, including data generation and augmentation (Section~\ref{subsec:training}).

\subsection{Transformer-based pointer policy}
\label{subsec:architecture}

\begin{figure*}[!htb] 
    \centering
    \includegraphics[width=\textwidth]{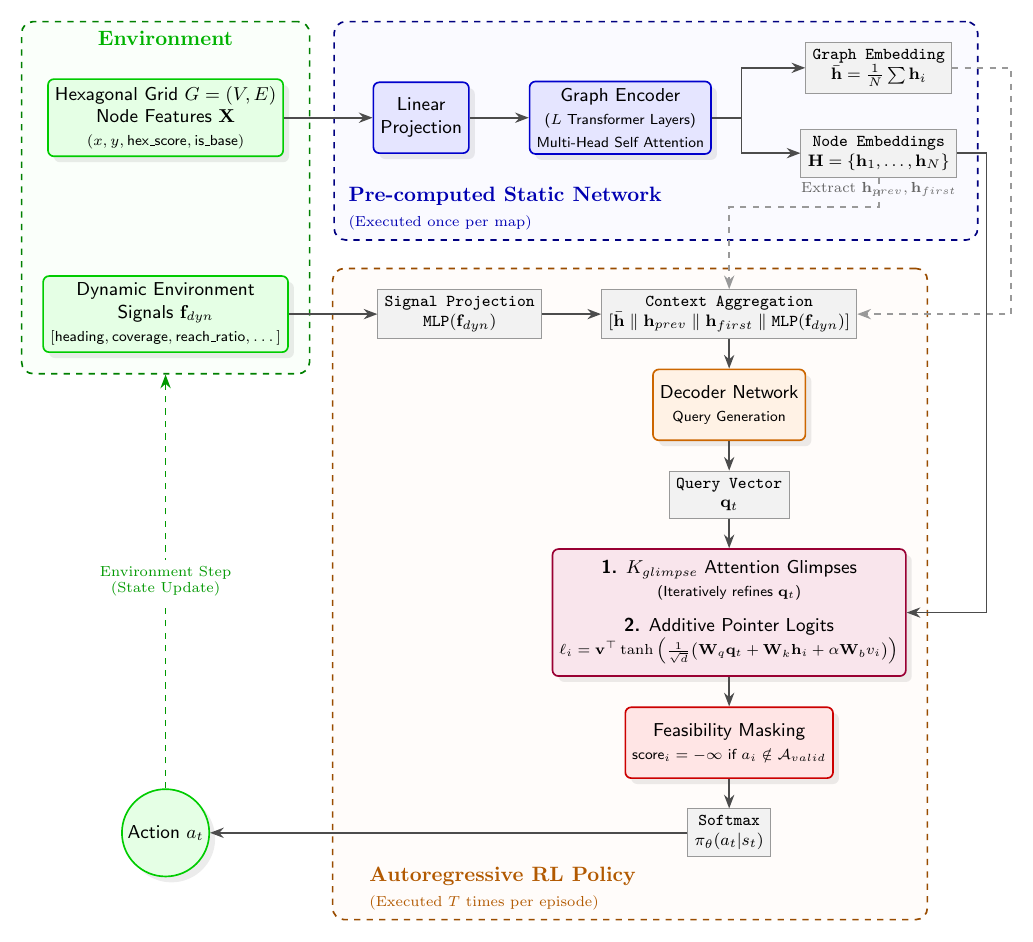}
    \caption{Proposed Transformer-based pointer policy for coverage path planning. The Graph Encoder pre-computes static node embeddings, while the Decoder dynamically aggregates the agent's spatial context and environmental signals to generate a query. The Pointer Network computes attention scores over valid nodes, strictly constrained by a feasibility mask to guarantee valid routing.}
    \label{fig:architecture}
\end{figure*}

Our policy follows the attention-based neural combinatorial optimization paradigm \citep{Kool2018, berto2025rl4co}: an encoder produces contextual node embeddings for the AOI graph, and an autoregressive decoder \emph{points} to the next node to visit. The full architecture is depicted in Fig.~\ref{fig:architecture}.

\paragraph{Node features and embedding.}
Each node $i\in\mathcal{V}$ is represented by a feature vector $\mathbf{x}_i=[x_i, y_i, w_i, m_i]$. To ensure the policy generalizes robustly across AOIs of varying absolute physical dimensions, the spatial coordinates $(x_i, y_i)$ are centered at the base node and normalized by the maximum radial extent of the graph within the instance. $w_i$ is the hexscore priority, and $m_i$ is a binary indicator identifying the base node. A learnable linear projection maps $\mathbf{x}_i$ to an initial embedding $\mathbf{h}_i^{(0)}\in\mathbb{R}^{d}$.

\paragraph{Graph-aware encoder.}
A multi-layer Transformer encoder exchanges information across nodes. To respect the sparse structure of the hex graph, attention is restricted to $k$-hop neighborhoods using the adjacency matrix as a mask, yielding graph-structured self-attention. After $L$ layers, we obtain contextual embeddings $\mathbf{h}_i=\mathbf{h}_i^{(L)}$ and a global graph embedding $\bar{\mathbf{h}} = \frac{1}{|\mathcal{V}|}\sum_{i}\mathbf{h}_i$.

\paragraph{Autoregressive decoder.}
At step $t$, the decoder constructs a context-aware query from four sources: (i)~the embedding of the current node $v_t$, (ii)~the embedding of the first node $v_0$, (iii)~the global graph embedding $\bar{\mathbf{h}}$, and (iv)~a set of environmental signals encoding coverage progress, heading direction, frontier statistics, and reachability status. These signals are projected to the model dimension and concatenated before passing through a context network (two-layer MLP with residual connection) that produces the decoder output $\mathbf{q}_t\in\mathbb{R}^{d}$.

\paragraph{Pointer mechanism with masking.}
The decoder output attends to the key projections of all node embeddings through $K_{glimpse}$ attention glimpses \citep{Kool2018}, iteratively refining the query vector $\mathbf{q}_t$. The final logits are computed via additive (Bahdanau-style) attention with a $\tanh$ activation and a learnable projection vector $\mathbf{v}$:
\begin{equation}
\ell_{t}(j) \;=\; \mathbf{v}^\top \tanh\left( \frac{1}{\sqrt{d}} \big(\mathbf{W}_q\mathbf{q}_t + \mathbf{W}_k\mathbf{h}_j + \alpha\, \mathbf{W}_b\, v_j\big) \right),
\end{equation}
where $v_j\in\{0,1\}$ encodes the visitation status of node $j$, $\alpha$ is a learnable gating scalar mapping the binary status into the continuous embedding space, and $\mathbf{W}_q, \mathbf{W}_k, \mathbf{W}_b$ are learnable weight matrices. An additive action mask $\mathcal{M}_t(j)\in\{-\infty,0\}$ enforces feasibility:
\begin{equation}
\pi_\theta(a_t=j\mid s_t)=\text{softmax}\Big(\text{clip}\big(\ell_{t}(j) + \mathcal{M}_t(j),\; C\big)\Big)_j,
\end{equation}
where $C>0$ is a $\tanh$ clipping constant that bounds logit magnitudes. The mask assigns $-\infty$ to non-neighbors and already-visited nodes, ensuring that all sampled trajectories are feasible by construction. During training, actions are sampled from the categorical distribution. For evaluation during the training phase (validation), greedy decoding is used exclusively to strictly monitor policy convergence. At test time, however, the trained policy supports multiple decoding strategies of varying computational cost to trade off latency for path quality, as detailed in Section~\ref{subsec:inference_strategies}.

While some attention-based routing models employ scaled dot-product mechanisms for the final pointer logits \citep{Kool2018}, we adopt an additive (Bahdanau-style) attention head following the original Pointer Network formulation \citep{Vinyals2015, Bello2016}. The additive attention evaluates the compatibility between the decoder's query and the candidate keys through a non-linear projection ($\tanh$), acting effectively as a single-hidden-layer Multi-Layer Perceptron. This non-linearity provides greater expressive capacity to capture complex geometric relationships between the agent's current state and the surrounding irregular grid. Furthermore, the intrinsic saturation of the inner $\tanh$ function naturally bounds the pre-logit activations. Combined with the outer clipping parameter $C$, this architectural choice significantly enhances numerical stability during policy gradient updates, preventing premature entropy collapse and maintaining healthy exploration in the early stages of training.

\paragraph{Complexity.}
With neighbor-restricted attention, encoder complexity scales as $\mathcal{O}(|\mathcal{V}|\cdot d_{max} \cdot d^2)$ per layer, where $d_{max}$ is the maximum node degree (at most~6 for hexagonal grids). Decoding is $\mathcal{O}(|\mathcal{V}|\cdot d^2)$ per step. This is advantageous for large AOIs compared to full-attention Transformers that scale quadratically in $|\mathcal{V}|$.

\subsection{Critic-free Group-Relative Policy Optimization}
\label{subsec:grpo}

Standard actor-critic methods such as PPO \citep{Schulman2017} require learning a value function $V_\phi(s)$ to estimate advantages. In combinatorial routing, the value function must generalize across highly diverse graph topologies with sparse terminal rewards, which often leads to high bias and training instability. Shared-instance baseline mechanisms, such as POMO \citep{kwon2020pomo}, mitigate this by evaluating multiple parallel trajectories constructed from diverse starting nodes, using their average return as an instance-specific baseline.

Because our maritime CPP formulation models missions deploying from a fixed base node, we cannot exploit starting-node symmetries. Instead, we adapt this shared-baseline principle through GRPO~\citep{shao2024deepseekmath}. For each training instance $n$, we sample a group of $G$ trajectories $\{\pi_{n}^{(g)}\}_{g=1}^{G}$. To guarantee constructive diversity across the group despite the single fixed starting node, we rely on the policy's stochastic categorical sampling, which is heavily promoted during early training via a high initial temperature annealing schedule ($T_{init}=1.5 \to 1.0$). We then compute the relative advantages by standardizing the episodic returns (Eq.~\ref{eq:return}) within the group:
\begin{equation}
A_{n}^{(g)} \;=\;
\frac{R_{n}^{(g)} - \mu_n}{\sigma_n + \epsilon},
\label{eq:grpo_adv}
\end{equation}
where
$\mu_n=\frac{1}{G}\sum_{g}R_{n}^{(g)}$,\;
$\sigma_n=\sqrt{\frac{1}{G}\sum_{g}(R_{n}^{(g)}-\mu_n)^2}$,
and $\epsilon$ is a small constant for numerical stability. This formulation evaluates whether a trajectory outperformed its peers on the exact same map, effectively bypassing the generalization bottleneck of a global critic network. The advantage $A_{n}^{(g)}$ is therefore an \emph{outcome-level} quantity: it is computed once per rollout from its total episodic return and shared across all decision steps of that rollout, consistent with the one-step MDP treatment of constructive routing \citep{berto2025rl4co}, rather than a per-step return-to-go.

The policy is then optimized via a clipped surrogate objective applied at the \emph{per-step} level. Let $r_{t}(\theta)={\pi_\theta(a_t\mid s_t)}/{\pi_{\theta_{\text{old}}}(a_t\mid s_t)}$ be the importance-sampling ratio for action $a_t$. The GRPO loss is:
\begin{multline}
\mathcal{L}_{\text{GRPO}}(\theta) = -\frac{1}{|\mathcal{T}|}\sum_{(n,g,t)\in\mathcal{T}} \min \Bigg( r_{t}(\theta) A_n^{(g)}, \\
\text{clip}\big(r_{t}(\theta), 1-\varepsilon, 1+\varepsilon\big) A_n^{(g)} \Bigg) - \beta \bar{\mathcal{H}},
\label{eq:grpo_loss}
\end{multline}

where $\mathcal{T}$ denotes the set of valid (non-padding) tokens across all trajectories, $\varepsilon$ is the clipping parameter, $\bar{\mathcal{H}}$ is the mean per-step policy entropy, and $\beta$ is the entropy coefficient. The per-step formulation, as opposed to per-trajectory clipping, ensures that the trust region constraint is enforced at each decision point, which is critical for long-horizon routing where per-trajectory ratios can grow exponentially with sequence length.

\paragraph{Multi-epoch reuse.}
For each batch of $B$ instances with $G$ rollouts each, we perform $K$ inner optimization epochs over shuffled minibatches, re-evaluating log-probabilities and entropy under the current policy parameters at each step. This is analogous to the inner-loop structure of PPO, and provides substantial sample efficiency compared to single-update REINFORCE methods.

\subsection{Early dead-end detection via BFS}
\label{subsec:deadend}

On irregularly shaped hexagonal grids, narrow passages, peninsulas, and obstacle configurations frequently create situations where the agent, having visited certain nodes, can no longer reach all remaining targets, a \emph{geometric dead-end}. Without early detection, the agent continues making decisions along a doomed trajectory, receiving misleading intermediate rewards before eventually triggering the terminal death penalty. This contaminates credit assignment: the policy cannot distinguish between the critical misstep that created the dead-end and the subsequent (irrelevant) actions.

We integrate a Breadth-First Search (BFS) reachability check into the environment. At each step~$t$, after the agent moves to node $v_t$, the BFS computes the set of unvisited nodes reachable from $v_t$ through unvisited (or transit-eligible) intermediate nodes, and verifies whether the terminal base node remains accessible. If either condition fails, i.e., some target nodes or the base are unreachable, the episode is terminated immediately with the death penalty $r_{death}$.

This mechanism provides two benefits:
\begin{enumerate}
    \item \textbf{Sharper credit assignment.} The trajectory is truncated at the step where the dead-end becomes inevitable, so the episodic return defining the outcome-level advantage (Eq.~\ref{eq:grpo_adv}) reflects the failure promptly instead of being diluted by subsequent uninformative steps. Since credit is shared uniformly across the rollout, this early truncation concentrates the negative signal on the short prefix containing the causal misstep.
    \item \textbf{Reduced wasted computation.} Doomed trajectories are truncated early, freeing rollout budget for informative trajectories. Empirically, enabling BFS detection substantially reduces the fraction of non-informative rollouts during early training, allowing the policy gradient to receive meaningful credit-assignment signal from the first epoch.
\end{enumerate}

The BFS has worst-case complexity $\mathcal{O}(|\mathcal{V}|+|\mathcal{E}|)$ per step. For hex grids with $|\mathcal{V}|\le 46$ and $|\mathcal{E}|\le 6|\mathcal{V}|$, this overhead is negligible compared to the Transformer forward pass.

\subsection{Training procedure}
\label{subsec:training}

\paragraph{Dataset generation.}
Training and evaluation use a synthetic dataset 
of 10{,}000 irregular hexagonal AOI instances 
($|\mathcal{V}|\in[28,46]$) with stochastic 
cell-level obstacle patterns; the generation 
procedure and morphological families are 
described in Section~\ref{subsec:dataset}.

\paragraph{Geometric augmentation.}
During training, each batch undergoes stochastic geometric augmentation with probability $p_{aug}=0.9$: random rotations and reflections are applied to the AOI graph. This exposes the policy to geometric invariances and significantly improves generalization to unseen AOI shapes without increasing the dataset size.

\paragraph{Temperature annealing.}
To balance exploration and exploitation, sampling temperature is linearly annealed from $T_{init}=1.5$ to $T_{final}=1.0$ over the first 10 epochs. Higher initial temperature encourages diverse trajectory exploration during early training, while convergence to unit temperature ensures that the final policy is evaluated under standard softmax probabilities.

\paragraph{Hyperparameters.}
The full training configuration is summarized in Table~\ref{tab:hyperparams}. Key choices include $G=16$ rollouts per instance (balancing baseline quality against computational cost), $K=4$ inner PPO-style epochs per batch, a learning rate of $3\times 10^{-5}$ with linear annealing, and an entropy coefficient of $\beta=0.02$.

\begin{table}[!htb]
\small
\centering
\caption{Training hyperparameters.}
\label{tab:hyperparams}
\begin{tabularx}{\linewidth}{@{} l Y c @{}}
\toprule
\textbf{Parameter} & \textbf{Description} & \textbf{Value} \\
\midrule
$d$ & Model dimension & 128 \\
$L$ & Encoder layers & 3 \\
$n_h$ & Attention heads & 8 \\
$K_{glimpse}$ & Decoder glimpses & 2 \\
$G$ & Rollouts per instance & 16 \\
$K$ & Inner optimization epochs & 4 \\
$\varepsilon$ & PPO clip parameter & 0.2 \\
$\beta$ & Entropy coefficient & 0.02 \\
lr & Learning rate & $3 \times 10^{-5}$ \\
-- & LR schedule & Linear annealing \\
$B$ & Batch size (instances) & 32 \\
$B_{mb}$ & Minibatch size (trajectories) & 8 \\
-- & Optimizer & Adam \\
$\|\nabla\|$ & Max gradient norm & 0.5\\
-- & Max epochs (early stop at 30) & 300 \\
-- & Augmentation probability & 0.9 \\
$T_{init}/T_{final}$ & Temperature annealing & 1.5 / 1.0 \\
-- & Early dead-end detection & Enabled \\
\bottomrule
\end{tabularx}
\end{table}

\paragraph{Influence of key hyperparameters.}
The temperature annealing schedule $(T_{\mathrm{init}} = 1.5 \!\to\! 1.0)$ is the most influential setting for feasibility: a higher initial temperature promotes constructive diversity within each GRPO group and prevents premature entropy collapse, at the cost of slower early convergence; annealing to unit temperature stabilizes the final policy. The group size $G = 16$ trades gradient-variance reduction against memory and per-step cost, and smaller groups were observed to yield noisier advantage estimates. The entropy coefficient $\beta = 0.02$ and the linearly annealed learning rate $(3\times10^{-5})$ jointly control the exploration-to-exploitation transition; larger $\beta$ delayed convergence while smaller values induced early determinism and lower validation success. Finally, training beyond epoch~30 improved training success while degrading validation performance (Fig.~\ref{fig:training_curves}), which motivated the validation-based checkpoint selection.

\section{Experimental Setup}
\label{sec:experiments}

\subsection{Dataset description}
\label{subsec:dataset}

Training instances are generated synthetically to represent a range of maritime AOI morphologies. Each instance is constructed in three stages: (i)~an irregular polygon is sampled to define the outer boundary of the AOI, (ii)~a hexagonal tessellation is applied with cell circumradius $r_h = r_s$ matching the sensor footprint (Section~\ref{subsec:hex_tessellation}), and (iii)~a subset of interior cells is randomly removed to simulate islands, shoals, or other navigational exclusion zones. This cell-level obstacle generation is mathematically equivalent to inscribing small polygonal holes within the AOI and discarding any hexagon whose centroid falls inside them, but operates directly on the graph topology without requiring explicit geometric intersection tests.

The outer polygons are drawn from three morphological families representing common maritime scenarios: (i)~compact convex regions 
(open-water patrol zones), (ii)~elongated or concave polygons (coastal strips, fjords, channel approaches), and (iii)~irregular shapes 
with narrow passages created by the interaction of boundary concavities and internal cell removals. The combination of polygon shape and stochastic cell removal produces a wide spectrum of graph topologies, including bottlenecks, peninsulas, and disconnected-looking corridors that are characteristic of real maritime environments with islands and reefs.

Since the number of visitable cells $|\mathcal{V}|$ depends on the ratio between the AOI area and the cell area (Section~\ref{subsec:hex_tessellation}), the geometric bounds of our dataset were rigorously calibrated to reflect real-world maritime domain parameters. We generate operational areas ranging from $1{,}600$ to $3{,}600$ square nautical miles ($\text{NM}^2$). Crucially, by constraining the total area rather than bounding box dimensions, the generator naturally accommodates highly diverse morphologies under the same operational footprint, spanning from compact open-water Search and Rescue (SAR) sectors to elongated coastal patrol corridors. The cell circumradius is sampled between $5.0$ and $7.0$~NM, explicitly matching the effective detection horizon of an X-band marine radar on a medium-sized Autonomous Surface Vehicle (ASV) or the EO/IR footprint of a tactical UAV. Additionally, the base node is stochastically placed at a standoff distance of $100$ to $250$~NM, simulating the realistic offshore transit from a coastal naval base or a mother ship.

By driving the tessellation strictly through these doctrinal operational capabilities, the resulting spatial graphs consistently emerge with $|\mathcal{V}|\in[28, 46]$ valid target cells. This demonstrates that instances of this topological scale are not merely mathematical abstractions, but rather the exact combinatorial resolution required to plan persistent surveillance missions for modern maritime assets.

A total of 10{,}000 instances are generated and partitioned into 8{,}000 training, 1{,}000 validation, and 1{,}000 test instances. Polygon vertices, cell removal patterns, and base-node locations are sampled randomly subject to connectivity constraints ensuring that the resulting graph remains connected.

\paragraph{Hamiltonian-path feasibility audit.}
Since our formulation requires visiting every cell exactly once (Section~\ref{subsec:tsp_relation}), it is essential that every instance in the dataset admits at least one feasible Hamiltonian path. We verify this exhaustively using a depth-first search with strict backtracking over the full graph. All 10{,}000 instances pass this audit. Consequently, the exhaustive DFS serves as a \emph{feasibility reference} confirming a 100\% theoretical solve rate. However, because DFS merely finds the first valid topological sequence without optimizing kinematic costs or distance, its path quality is generally poor. Therefore, it is utilized strictly as a ground-truth upper bound for feasibility, not as a target benchmark for path optimization.
Because this reference only certifies feasibility and does not optimize distance or turns, its path-quality metrics are expected to exceed those of the distance-optimizing MILP and of the learned policy; they are reported for completeness only.

In the experiments reported here, all instances use uniform coverage priority ($w_i=0$ for all $i\in\mathcal{V}$), reducing the objective to 
minimum-cost complete coverage. The architecture natively supports non-uniform priorities via the hexscore input channel; experiments with 
heterogeneous priority fields are deferred to future work (Section~\ref{sec:conclusion}).

\subsection{Baseline methods}
\label{subsec:baselines}

To evaluate the learned policy against a broad spectrum of classical CPP strategies, we implement 
13~heuristic baselines spanning six algorithmic 
families. All baselines operate on the same hexagonal grid graph under identical adjacency constraints. However, a crucial methodological distinction exists regarding visitation constraints. While our RL policy is strictly constrained to single-visit paths via its action masking (terminating an episode if a dead-end is reached), classical heuristic baselines may traverse previously visited nodes (via backtracking, 
flyback transit, or overlapping sweeps) as a 
natural consequence of their construction logic, 
thereby achieving complete area coverage at the 
cost of redundant motion. This difference highlights a fundamental operational trade-off: traditional heuristics guarantee 100\% coverage completion at the cost of inefficient overlapping paths, whereas our learned policy prioritizes maximum kinematic and distance efficiency by attempting to solve the stricter Hamiltonian path variant. The families and their members are summarized in Table~\ref{tab:baselines}.

\begin{table*}[!htb]
\small
\centering
\caption{Baseline heuristic methods grouped by 
algorithmic family. All methods operate on the same 
hex-graph representation under identical 
constraints.}
\label{tab:baselines}
\begin{tabularx}{\textwidth}{@{} l l Y @{}}
\toprule
\textbf{Family} & \textbf{Method} & 
\textbf{Description} \\
\midrule

\multirow{3}{*}{\parbox{2.8cm}{\raggedright 
Linear sweep\\(Boustrophedon)}}
& \texttt{sweep\_boustrophedon} 
& Classic lawnmower pattern with alternating row 
directions, minimizing inter-row transit. \\
& \texttt{sweep\_row\_oneway} 
& Unidirectional row sweep; all rows traversed in 
the same direction with flyback transit. \\
& \texttt{sweep\_segment\_snake} 
& Obstacle-aware boustrophedon that decomposes the 
AOI into convex segments and applies snake 
patterns within each. \\
\midrule

\multirow{2}{*}{\parbox{2.8cm}{\raggedright 
Interleaved sweep\\(Skip-row)}}
& \texttt{sweep\_row\_interleave} 
& Skips rows to allow wider turning arcs; rows are 
visited in an interleaved order 
(e.g., 1, 7, 2, 8, \ldots). \\
& \texttt{sweep\_segment\_interleave} 
& Combines segment decomposition with interleaved 
row ordering for kinematically constrained 
vehicles. \\
\midrule

\multirow{3}{*}{\parbox{2.8cm}{\raggedright 
Contour / Spiral\\(Boundary rings)}}
& \texttt{boundary\_spiral\_inward} 
& Traces concentric rings from the AOI perimeter 
inward toward the centroid. \\
& \texttt{boundary\_spiral\_outward} 
& Starts at the centroid and spirals outward to the 
perimeter. \\
& \texttt{sweep\_boundary\_peel} 
& Iteratively erodes the outermost layer of 
unvisited cells, adapting dynamically if the 
remaining area splits. \\
\midrule

\multirow{2}{*}{\parbox{2.8cm}{\raggedright 
Spanning-tree\\coverage (STC)}}
& \texttt{stc\_tree\_coverage} 
& Builds a minimum spanning tree over grouped 
cells; the agent circumnavigates the tree edges. \\
& \texttt{stc\_like} 
& Adapted STC for hex grids with mandatory 
start/end nodes, using BFS to connect dead 
branches. \\
\midrule

\multirow{2}{*}{\parbox{2.8cm}{\raggedright 
Graph-based\\local search}}
& \texttt{warnsdorff} 
& Adapts the Knight's Tour heuristic: always moves 
to the unvisited neighbor with the fewest 
remaining neighbors, clearing difficult corners 
first. \\
& \texttt{dfs\_backtrack} 
& Depth-first greedy traversal with shortest-path 
backtracking when a dead-end is reached. \\
\midrule

{\parbox{2.8cm}{\raggedright Space-filling\\curve}}
& \texttt{morton\_zorder} 
& Orders cell centroids by their Morton code 
(bit-interleaved coordinates), producing a 
Z-order curve that preserves spatial locality. \\

\bottomrule
\end{tabularx}
\end{table*}

The six families represent fundamentally different algorithmic paradigms in CPP: geometric decomposition (sweep families), morphology-following (contour/spiral), topological coverage guarantees (STC), local graph heuristics, and space-filling 
curves. This diversity ensures that performance comparisons are not biased toward any single planning philosophy.

\paragraph{Exact and metaheuristic optimizers.} Beyond the heuristic baselines above, we additionally include a budgeted exact solver (CP-SAT MILP) and a feasibility-preserving memetic genetic algorithm (GA) as \emph{per-instance} optimization references, rather than amortized policies. Unlike the constructive heuristics, these solvers re-optimize each instance from scratch under a fixed time budget; their formulation, budget, and comparative results (including paired statistical tests) are presented in Section~\ref{subsec:exact_comparison}.

\subsection{Inference strategies}
\label{subsec:inference_strategies}

Unlike classical exact solvers that compute a single solution, generative routing models allow multiple decoding strategies at inference time. We evaluate the trained policy on the test set using three modes that trade computational cost for solution quality:
\begin{enumerate}
    \item \textbf{Greedy (\texttt{RL-Greedy}):} At each step, the node with the highest logit is selected deterministically. This provides single-pass, lowest-latency inference ($\mathcal{O}(|\mathcal{V}|)$ steps).
    \item \textbf{Best-of-$K$ sampling (\texttt{RL-BoK}):} $K$ independent trajectories are sampled in parallel from the stochastic policy. We select the trajectory that successfully completes the Hamiltonian path with the highest episodic return (i.e., minimum kinematic and distance cost). If no trajectory achieves strict complete coverage (i.e., all $K$ rollouts reach a geometric dead-end before visiting every cell), we employ a robust fallback logic: the model outputs the partial trajectory that maximizes the total covered area before failure, breaking ties by minimum distance. Thanks to GPU batching, generating $K=16$ parallel rollouts is computationally efficient and only marginally slower than greedy decoding.
    \item \textbf{Best-of-$K$ with local search (\texttt{RL-BoK+2opt}):} The best sampled trajectory is refined via a 2-opt local search that iteratively reverses sub-segments of the tour to untangle crossings, accepting improvements in total path cost. Crucially, because the environment is a sparse graph and not a fully connected Euclidean space, feasibility (strict adjacency and single-visit constraints) is explicitly verified after each proposed reversal. This demonstrates the seamless integration of neural construction heuristics with classical refinement.
\end{enumerate}

The \texttt{RL-BoK} and \texttt{RL-BoK+2opt} variants are then applied only at test time to the selected trained model and should therefore be interpreted as inference-time enhancements rather than separately trained policies.
\subsection{Evaluation metrics}
\label{subsec:metrics}

To capture the trade-off between guaranteed coverage, path efficiency, and computational latency, we evaluate all methods along five route-quality and feasibility metrics, with planning latency reported separately:

\begin{enumerate}
    \item \textbf{Hamiltonian Success Rate (HSR, \%)}: the fraction of test instances for which a method produces a valid start-to-terminal path that visits every required node exactly once. For the proposed RL policy, this is the primary success criterion. For classical baselines, this metric indicates whether their output happens to satisfy the stricter Hamiltonian requirement, even though they were not designed for it.

    \item \textbf{Coverage Completion Rate (CCR, \%)}: the fraction of test instances for which all required nodes are covered and the terminal node is reached, regardless of revisits. This reflects the native operating regime of classical coverage heuristics.

    \item \textbf{Node Revisits}: the average number of previously visited nodes traversed during the generated route. This measures the overlap cost incurred by revisit-allowed strategies. By construction, the proposed RL policy yields zero revisits on instances counted as Hamiltonian successes.

    \item \textbf{Normalized Path Distance}: the total Euclidean route length normalized by the characteristic inter-cell spacing. To isolate route quality from outright feasibility, this metric is reported conditionally on the pairwise common solved subset between the compared methods.

    \item \textbf{Number of Turns}: the number of non-zero heading changes along the route. As with normalized distance, this metric is reported on the pairwise common solved subset in order to compare route quality independently of failure cases.

    \item \textbf{Planning latency}: the median wall-clock time per instance, measured with batch size~1 and explicit device synchronization. This metric is reported separately in Section~\ref{subsec:efficiency}.
\end{enumerate}

For the main route-quality comparison, we use the subset
\[
\begin{aligned}
\mathcal{I}_{\cap}^{(m)} = 
\{\, i \in \mathcal{D}_{test} \;|\;& 
\texttt{RL-BoK16+2opt solves } i \text{ under HSR} \\
&\text{and baseline } m \text{ completes } i \,\},
\end{aligned}
\]
and compute distance and turn metrics only on $\mathcal{I}_{\cap}^{(m)}$.
This conditional analysis should be interpreted as a comparison of route
quality on matched instances, not as a substitute for global feasibility metrics.

\subsection{Implementation details}
\label{subsec:implementation}

All experiments were conducted under Windows Subsystem for Linux~2 (WSL2) running Ubuntu~24.04.4~LTS on a workstation equipped with an AMD Ryzen~9~5900HX CPU, 31~GiB of system RAM, and an NVIDIA GeForce RTX~3070 Laptop GPU (8~GB VRAM). The software stack utilized Python~3.12.2, PyTorch~2.4.0, and CUDA~12.4.

The neural policy comprises approximately 2.05~million trainable parameters. Training was configured for up to 300~epochs with early stopping (patience of 4~epochs on validation success rate); the best checkpoint was selected at epoch~30 (val.\ SR\,=\,95.5\%), after which validation performance began to decrease while training performance continued to rise. The total wall-clock time for the 34~epochs executed was approximately 440~GPU-hours (seed~42).

Model selection was based exclusively on validation performance under greedy decoding. Final test-set evaluation was performed on the selected checkpoint using the three inference modes described in Section~\ref{subsec:inference_strategies}.

To assess the feasibility of real-time onboard deployment, all reported inference latencies reflect single-instance end-to-end processing (batch size~1) including data transfer, environment construction, and neural forward pass with explicit device synchronization, reporting the median over the 1{,}000~test instances. Classical heuristic baselines 
were executed sequentially on the CPU; neural inference used GPU acceleration. Detailed latency 
results are reported in Section~\ref{subsec:efficiency}.

\section{Results and Discussion}
\label{sec:results}
 
All metrics reported in this section are computed exclusively on
the 1{,}000 unseen test instances, which were held out during
both training and hyperparameter selection.

\subsection{Coverage performance and feasibility}
\label{subsec:coverage}
 
Table~\ref{tab:coverage} reports the Hamiltonian Success Rate (HSR), Coverage Completion Rate (CCR), and mean node revisits for every evaluated method.  Two structural patterns emerge immediately.
 
First, all 12~revisit-allowed heuristics achieve 100\% CCR by design: when a geometric dead-end is encountered, they backtrack or fly back through previously visited cells,
trading path efficiency for coverage guarantees.  However, \emph{none} of these heuristics produces a Hamiltonian (zero-revisit) path on any test instance with the sole exception of \texttt{DFS\_Backtrack}, which achieves HSR\,=\,33.8\%.  This confirms that on irregular hex grids with obstacles, achieving strict single-visit coverage is a qualitatively harder objective than relaxed coverage.
 
Second, \texttt{Warnsdorff}, the only heuristic explicitly designed for Hamiltonian-like traversal, achieves HSR\,=\,CCR\,=\,46.0\%.  Because it forbids revisits, every
failure is simultaneously a coverage failure. Among the greedy construction heuristics evaluated here, \texttt{Warnsdorff} is the strongest, although it still fails on more than half of the instances.

\begin{figure}[!t]
\centering
\includegraphics[width=\linewidth]{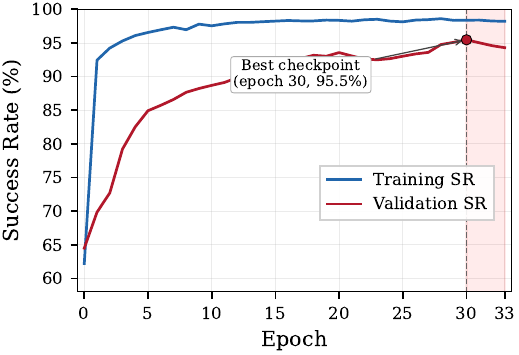}
\caption{Training and validation success rate (greedy decoding) across epochs. The dashed line marks the selected checkpoint (epoch~30, val.\ SR\,=\,95.5\%). The shaded region indicates overfitting, where validation performance declines while training performance remains stable.}
\label{fig:training_curves}
\end{figure}

Fig.~\ref{fig:training_curves} shows the evolution of training and validation success rates across epochs. The policy reaches near-saturation training performance (${>}98\%$) within the first 12~epochs, while validation performance converges more gradually. Beyond epoch~30, validation SR begins to decline while training SR
continues to increase marginally, confirming standard overfitting behavior and validating the early-stopping criterion.
 
Against this landscape, the proposed RL policy delivers a substantial improvement. Under greedy decoding alone, \texttt{RL-Greedy} achieves HSR\,=\,95.7\%, more than double the best heuristic. Stochastic Best-of-16 sampling (\texttt{RL-BoK16}) raises this value to 98.6\%, and the addition of adjacency-aware 2-opt refinement (\texttt{RL-BoK16+2opt}) reaches 99.1\%, 0.9 percentage points below the exhaustive feasibility reference (\texttt{Exact\_DFS}, HSR\,=\,100\%). All three RL variants achieve zero revisits on every solved instance, confirming that the action-masking mechanism (Section~\ref{subsec:mdp}) enforces strict Hamiltonian feasibility by construction.
 
\begin{figure*}[!htb]
\centering
\begin{minipage}[c]{0.5\textwidth}

  \begin{subfigure}[c]{\linewidth}
    \centering
    \includegraphics[width=0.98\linewidth]{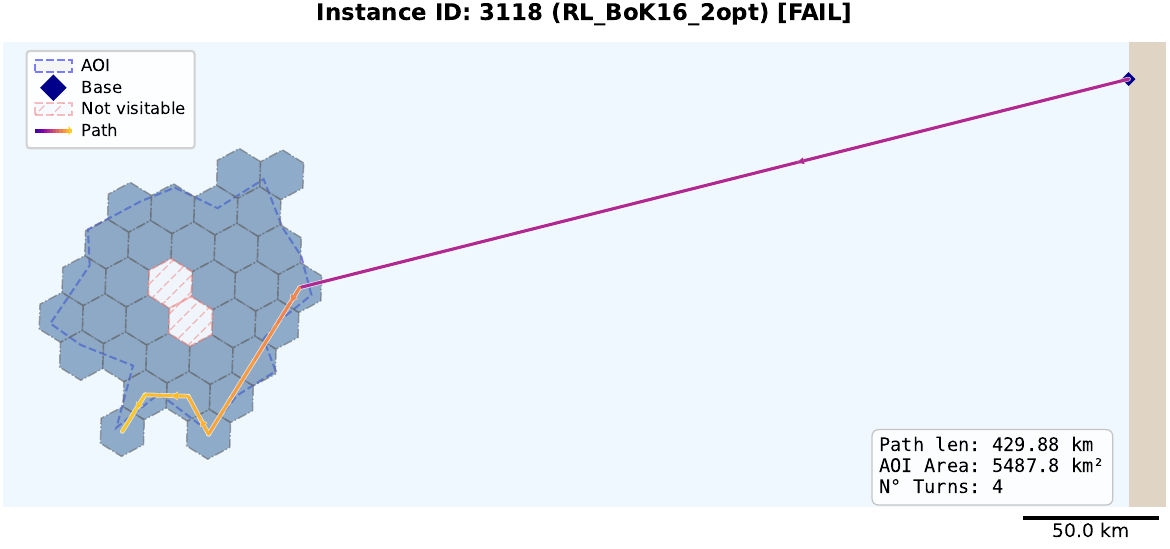}
    \caption{\scriptsize \textbf{Corridor trap:} a narrow passage is exhausted before all regions remain reachable.}
    \label{fig:fail_corridor}
  \end{subfigure}
  \begin{subfigure}[c]{\linewidth}
    \centering
    \includegraphics[width=0.98\linewidth]{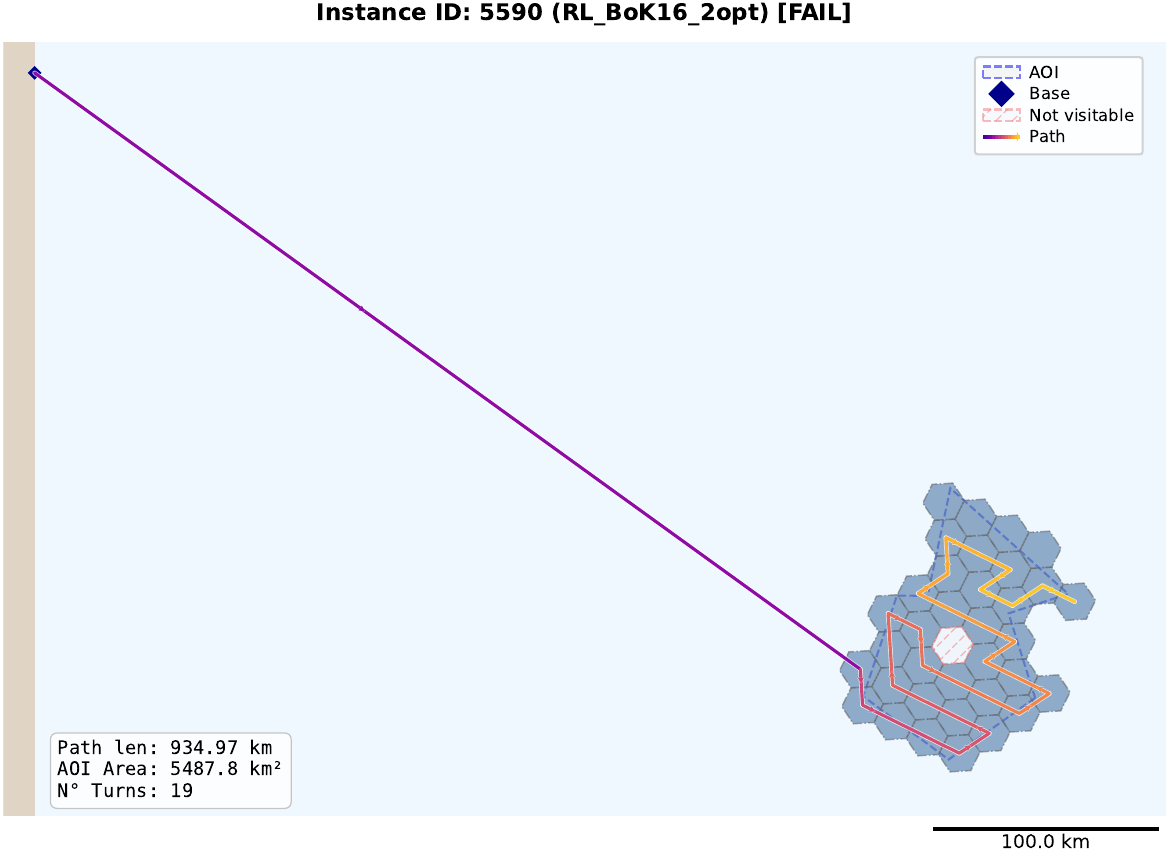}
    \caption{\scriptsize \textbf{Graph bisection:} crossing a bottleneck disconnects unvisited cells.}
    \label{fig:fail_bisection}
  \end{subfigure}
\end{minipage}
\begin{minipage}[c]{0.48\textwidth}
  \begin{subfigure}[c]{\linewidth}
    \centering
    \includegraphics[width=0.9\textwidth]{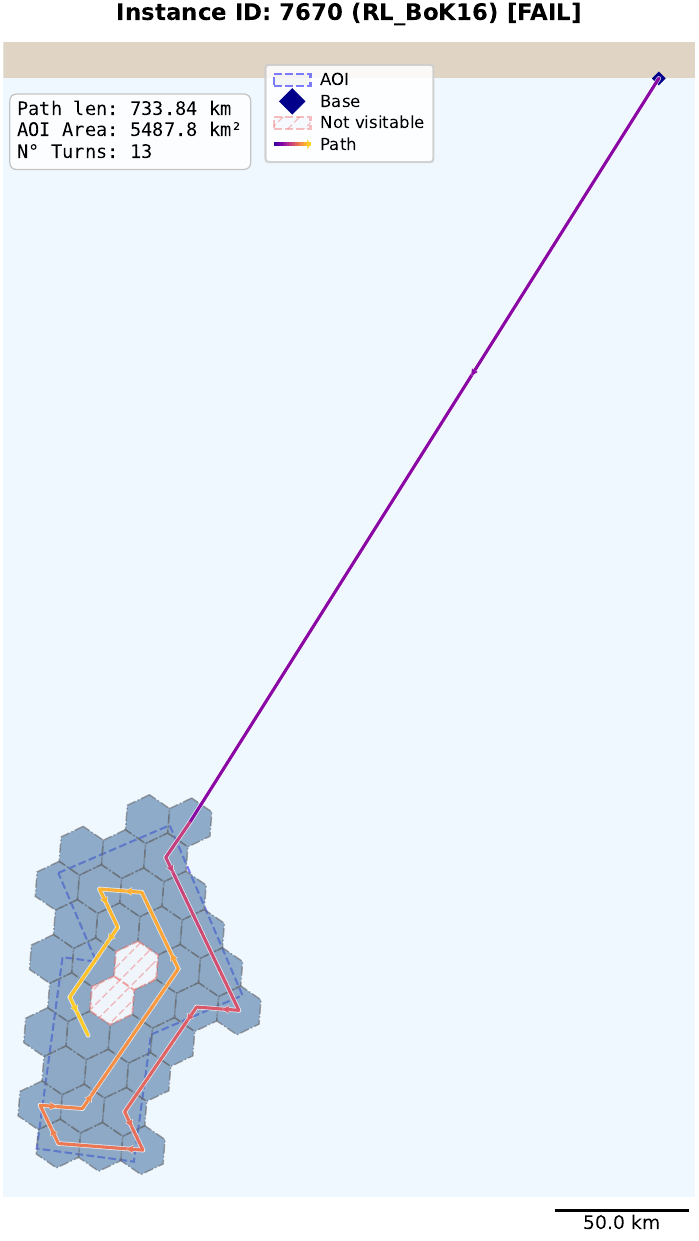}
    \caption{\scriptsize \textbf{Self-occlusion:} wrapping around an obstacle blocks access without revisits.}
    \label{fig:fail_wedge}
  \end{subfigure}
\end{minipage}

\caption{Residual failure modes of the \texttt{RL-BoK16+2opt} inference mode on the nine unsolved test instances (0.9\%). All failures arise from severe bottlenecked topologies in which a sampled rollout commits to a branch that leaves remaining cells unreachable under the strict no-revisit constraint. Best-of-16 sampling reduces these events, but adjacency-aware 2-opt cannot recover infeasible partial constructions because it only refines feasible tours. These cases motivate explicit connectivity lookahead or bounded revisitation in future operational variants.}
\label{fig:failure_modes}
\end{figure*}

The progression from 95.7\% to 99.1\% across the three inference modes illustrates the value of stochastic sampling and local refinement at test time: sampling explores alternative branching decisions in topologically challenging instances, and 2-opt untangles residual crossings without breaking adjacency constraints.  Manual inspection of the nine instances not solved by \texttt{RL-BoK16+2opt} reveals a consistent topological pattern. All failures involve internal obstacles or narrow passages that partition the AOI into sub-regions connected by one or two-cell bottlenecks. In these cases, the policy successfully covers one sub-region but exhausts the cells forming the connecting passage, leaving the remaining sub-region unreachable under the strict no-revisit constraint. Fig.~\ref{fig:failure_modes} visualizes the three recurrent failure modes: corridor traps, graph bisections, and self-occlusion around internal obstacles. These failures are not corrected by the 2-opt refinement because 2-opt only improves already feasible sampled tours; it cannot restore Hamiltonian feasibility when all sampled rollouts have already reached a geometric dead-end. This sequencing failure indicates that the remaining errors are caused by globally non-obvious entry--exit decisions through bottlenecks, a long-horizon planning challenge that the local attention mechanism and 16 stochastic samples are sometimes insufficient to resolve. Increasing the sampling budget $K$, adding explicit graph-connectivity lookahead to the decoder, or allowing a bounded node-revisitation budget are potential mitigations for future work.

\begin{table*}[!t]
\small\centering
\caption{Coverage performance on 1{,}000 unseen test instances.
  HSR: Hamiltonian Success Rate (strict single-visit).
  CCR: Coverage Completion Rate (revisits permitted).
  Methods are grouped by type and sorted by HSR within each group.}
\label{tab:coverage}
\begin{tabular}{@{} l l r r r@{\;\,$\pm$\;\,}l @{}}
\toprule
\textbf{Type} & \textbf{Method} & \textbf{HSR (\%)} & \textbf{CCR (\%)}
  & \multicolumn{2}{c}{\textbf{Revisits\,$\mu\pm\sigma$}} \\
\midrule
Feasibility & Exact\_DFS               & 100.0 & 100.0 & 0.0 & 0.0 \\
\midrule
\multirow{3}{*}{RL (ours)}
& RL-BoK16+2opt                   &  99.1 &  99.1 & 0.0 & 0.0 \\
& RL-BoK16                        &  98.6 &  98.6 & 0.0 & 0.0 \\
& RL-Greedy                       &  95.7 &  95.7 & 0.0 & 0.0 \\
\midrule
\multirow{13}{*}{Heuristic}
& Warnsdorff                       &  46.0 &  46.0 &  0.0 &  0.0 \\
& DFS\_Backtrack                   &  33.8 & 100.0 &  3.2 &  6.0 \\
& Sweep\_Boustrophedon             &   0.0 & 100.0 &  6.4 &  2.0 \\
& Sweep\_Segment\_Snake            &   0.0 & 100.0 &  6.5 &  2.1 \\
& Sweep\_Row\_OneWay               &   0.0 & 100.0 &  8.5 &  3.4 \\
& Boundary\_Spiral\_outward        &   0.0 & 100.0 & 12.3 & 12.6 \\
& Boundary\_Spiral\_inward         &   0.0 & 100.0 & 12.4 & 12.7 \\
& Sweep\_Segment\_Interleave       &   0.0 & 100.0 & 16.0 &  4.1 \\
& Sweep\_Boundary\_Peel            &   0.0 & 100.0 & 16.2 & 12.5 \\
& Morton\_Zorder                   &   0.0 & 100.0 & 18.5 &  5.4 \\
& Sweep\_Row\_Interleave           &   0.0 & 100.0 & 18.4 &  4.2 \\
& STC\_Tree\_Coverage              &   0.0 & 100.0 & 35.8 &  3.5 \\
& STC\_like                        &   0.0 & 100.0 & 35.8 &  3.5 \\
\midrule
\multirow{2}{*}{\shortstack[l]{Exact /\\Evolutionary}}
& MILP\_CPSAT (300\,s) & 100.0 & 100.0 & 0.0 & 0.0 \\
& GA (300\,s)          &  97.9 &  97.9 & 0.0 & 0.0 \\
\bottomrule
\end{tabular}
\end{table*} 

\subsection{Ablation of method components}
\label{subsec:ablation}
To isolate the contribution of the two non-standard components of our training scheme, BFS early dead-end detection (Section~\ref{subsec:deadend}) and geometric augmentation (Section~\ref{subsec:training}), we retrain the policy
under three configurations for a controlled budget of 25~epochs, holding all other hyperparameters fixed (Table~\ref{tab:hyperparams}) and evaluating on the
same 1{,}000 test instances under \texttt{RL-BoK16+2opt} decoding. These runs are independent of the production model reported in Tables~\ref{tab:coverage}--\ref{tab:quality} and are intended only to measure the \emph{relative} effect of each component.

\begin{table}[!t]
\small\centering
\caption{Component ablation (controlled 25-epoch runs, \texttt{RL-BoK16+2opt}
decoding on the 1{,}000 test instances). Each row removes a single component
from the full configuration.}
\label{tab:ablation}
\begin{tabular}{@{} l r r @{}}
\toprule
\textbf{Configuration} & \textbf{HSR (\%)} & \textbf{$\Delta$ HSR} \\
\midrule
Full (BFS + augmentation)        & 98.1 & --- \\
\;\;$-$ BFS dead-end detection    & 24.8 & $-73.3$ \\
\;\;$-$ geometric augmentation    & 97.2 & $-0.9$ \\
\bottomrule
\end{tabular}
\end{table}

Removing BFS dead-end detection degrades test HSR drastically, from 98.1\% to 24.8\%, a level comparable to the strongest revisit-free heuristic (\texttt{DFS\_Backtrack}, 33.8\%). Without the reachability check, the policy receives delayed and poorly attributed penalties: the negative signal arrives many steps after the decision that rendered coverage infeasible, contaminating
credit assignment and preventing the policy from reliably learning to avoid geometric dead-ends. This confirms that early dead-end detection is a prerequisite for effective learning on irregular hex grids, not an incremental optimization. Removing geometric augmentation leaves test HSR almost unchanged
(97.2\% vs.\ 98.1\%), a modest but consistent $0.9$-point reduction attributable to its role as a regularizer that exposes the policy to rotational and reflective invariances absent from the finite training set.

\subsection{Path quality analysis}
\label{subsec:quality}
 
To compare route quality independently of feasibility
differences, distance and turn metrics are computed on the
\emph{pairwise common solved subset}
$\mathcal{I}_{\cap}^{(m)}$ between each method~$m$ and the
reference solver \texttt{RL-BoK16+2opt}, which solved
991~of~1{,}000 test instances.
Table~\ref{tab:quality} summarizes these results.
 
\paragraph{Distance}
\texttt{RL-BoK16+2opt} achieves the lowest mean normalized distance among the heuristic and feasibility-reference baselines ($3.100 \pm 0.314$) across all 991~common instances, outperforming every heuristic and the feasibility-reference DFS. The best heuristic, \texttt{Warnsdorff} ($3.325 \pm 0.38$), is 7.4\% longer, but is only comparable on 454~instances due to its low solve rate.  Among full-coverage heuristics, the best performer is \texttt{DFS\_Backtrack} ($3.340 \pm 0.449$), 7.7\% longer than the RL policy.  The most commonly deployed survey pattern, \texttt{Boustrophedon} ($3.564 \pm 0.429$), incurs 15.0\% more distance.  At the extreme, the spanning-tree methods (\texttt{STC}) produce paths 49\% longer than the RL solution.
 
The 2-opt refinement provides a modest but consistent distance improvement over raw sampling: \texttt{RL-BoK16+2opt} is 0.3\% shorter than \texttt{RL-BoK16} ($3.110 \pm 0.315$). Greedy decoding and Best-of-16 produce identical mean distances, confirming that stochastic sampling primarily improves feasibility rather than path geometry.
 
\paragraph{Turns}
The RL policy produces dramatically smoother paths. \texttt{RL-BoK16} achieves the fewest turns among the heuristic baselines ($32.3 \pm 4.6$), 24.1\% fewer than the best heuristic \texttt{Warnsdorff} ($42.5 \pm 5.6$) and 37.5\% fewer than \texttt{Boustrophedon} ($51.6 \pm 6.6$).  This reduction is a direct consequence of the continuous turn penalty $f(\theta)$ in the reward function (Equation~\ref{eq:turn_penalty}), which the policy internalizes to produce kinematically smooth trajectories.  The 2-opt refinement increases turns slightly ($34.1 \pm 5.3$) relative to raw BoK because segment reversals can introduce heading changes while reducing distance, a classical distance and turns trade-off in 2-opt optimization on constrained graphs.
 
\paragraph{Revisits}
All three RL variants achieve exactly zero revisits by construction (action masking forbids revisits).  Among heuristics, the cost of guaranteed coverage ranges from
3.2~revisits (DFS\_Backtrack) to 35.8 (STC), which translates to 8--95\% redundant motion.  This redundancy represents wasted fuel and endurance in maritime operations,
a cost entirely avoided by the RL policy.
 
\begin{table*}[!tb]
\small\centering
\caption{Path quality on the pairwise common solved subset
  with \texttt{RL-BoK16+2opt} (991 of 1{,}000 test instances solved).
  $n$ denotes the number of instances solved by \emph{both} the listed
  method and the RL reference. Methods sorted by normalized distance
  within each group.}
\label{tab:quality}
\begin{tabular}{@{} l l r r@{\;\,$\pm$\;\,}l r@{\;\,$\pm$\;\,}l r r @{}}
\toprule
\textbf{Type} & \textbf{Method} & $n$
  & \multicolumn{2}{c}{\textbf{Dist.\,$\mu\pm\sigma$}}
  & \multicolumn{2}{c}{\textbf{Turns\,$\mu\pm\sigma$}}
  & \textbf{Steps} & \textbf{Revisits} \\
\midrule
\multirow{3}{*}{RL}
& RL-BoK16+2opt & 991 & 3.100 & 0.314 & 34.1 & 5.3 & 37.8 & 0.0 \\
& RL-BoK16      & 986 & 3.110 & 0.315 & 32.3 & 4.6 & 37.8 & 0.0 \\
& RL-Greedy     & 957 & 3.111 & 0.314 & 32.9 & 4.9 & 37.8 & 0.0 \\
\midrule
Feasibility
& Exact\_DFS    & 991 & 3.308 & 0.373 & 43.5 & 6.7 & 37.8 & 0.0 \\
\midrule
\multirow{13}{*}{Heuristic}
& Warnsdorff                &  454 & 3.325 & 0.379 & 42.5 & 5.6 & 37.2 & 0.0 \\
& DFS\_Backtrack            &  991 & 3.340 & 0.447 & 49.4 & 12.4 & 41.0 & 3.2 \\
& Sweep\_Boustrophedon      &  991 & 3.564 & 0.427 & 51.6 & 6.6 & 44.2 & 6.4 \\
& Sweep\_Segment\_Snake     &  991 & 3.567 & 0.428 & 51.7 & 6.5 & 44.3 & 6.5 \\
& Sweep\_Row\_OneWay        &  991 & 3.672 & 0.483 & 54.4 & 6.9 & 46.3 & 8.5 \\
& Boundary\_Spiral\_inward  &  991 & 3.864 & 0.771 & 51.3 & 19.1 & 50.3 & 12.5 \\
& Sweep\_Boundary\_Peel     &  991 & 3.865 & 0.730 & 60.0 & 17.8 & 54.0 & 16.2 \\
& Boundary\_Spiral\_outward &  991 & 3.866 & 0.764 & 51.1 & 18.2 & 50.1 & 12.3 \\
& Sweep\_Segment\_Interl.   &  991 & 3.980 & 0.564 & 60.9 & 8.0 & 53.8 & 16.0 \\
& Sweep\_Row\_Interleave    &  991 & 4.058 & 0.552 & 54.1 & 7.4 & 56.2 & 18.4 \\
& Morton\_Zorder            &  991 & 4.135 & 0.618 & 72.5 & 9.7 & 56.3 & 18.5 \\
& STC\_Tree\_Coverage       &  991 & 4.622 & 0.720 & 75.9 & 12.8 & 73.6 & 35.8 \\
& STC\_like                 &  991 & 4.622 & 0.720 & 78.9 & 12.0 & 73.6 & 35.8 \\
\midrule
\multirow{2}{*}{\shortstack[l]{Exact /\\Evol.}}
& MILP\_CPSAT & 991 & 3.079 & 0.311 & 29.9 & 4.1 & 37.8 & 0.0 \\
& GA          & 970 & 3.100 & 0.317 & 32.3 & 5.3 & 37.8 & 0.0 \\
\bottomrule
\end{tabular}
\end{table*}

\subsection{Comparison with exact and evolutionary baselines}
\label{subsec:exact_comparison}
To position the learned policy against non-heuristic optimizers, we compare it with a budgeted exact solver (CP-SAT MILP) and a memetic genetic algorithm (GA), both granted a 300\,s per-instance budget and optimizing the same distance-plus-turn objective as the reward (Eq.~\ref{eq:return}).
Tables~\ref{tab:coverage} and~\ref{tab:quality} report their results; paired comparisons use the common solved subset with \texttt{RL-BoK16+2opt}.

The MILP solves all 1{,}000 instances feasibly but proves optimality on only 35.5\% within the budget, at 229\,s per instance on average; the GA reaches 97.9\% feasibility at 179\,s. The learned policy attains normalized distance within $0.6\%$ of the MILP solutions (median 3.073 vs.\ 3.055; Cliff's
$\delta=0.04$) and statistically indistinguishable from the GA ($p=0.22$, Wilcoxon), while producing an inference decision in $\sim$32\,ms---roughly four orders of magnitude faster. We emphasize that the proposed method does \emph{not} outperform the exact solver in path length; rather, it matches near-optimal quality at inference latencies compatible with onboard replanning, whereas the exact and evolutionary solvers do not. The MILP yields the fewest turns (29.9), because it directly optimizes the full route objective over the complete instance, rather than constructing the tour autoregressively through local policy decisions, and the 2-opt refinement trades a small increase in heading changes for reduced distance (Section~\ref{subsec:quality}).

It is important to distinguish the \emph{nature} of these costs. The 300~s budget is a per-instance, online cost incurred by the MILP and GA \emph{every time} a new area is planned. In contrast, the DRL training cost ($\approx$440~GPU-hours, incurred once, offline, on shore) produces a policy that is subsequently reused across an unlimited number of unseen instances at $\sim$32~ms each, with no re-optimization. These are costs of different nature: a one-time amortized investment versus a recurring per-instance expense. For a deployed autonomous surface vehicle, the operationally relevant quantity is the per-instance planning latency during the mission, where the learned policy is roughly four orders of magnitude faster than the budgeted solvers while retaining near-optimal quality. The training budget is therefore not directly comparable to the per-instance optimization budget on a single time axis; both are reported in Table~\ref{tab:latency} for completeness.
To make the amortization concrete, the one-time training cost ($\approx$440~GPU-hours, i.e., $\approx$1.58$\times$10$^{6}$~s) is recovered, in cumulative per-instance solver time alone, after approximately 6{,}900 instances for the MILP and 8{,}850 for the GA; beyond this point the learned policy is strictly cheaper in total compute, while every additional instance is planned in $\sim$32~ms. This break-even is readily exceeded over the operational lifetime of a fleet, for which the policy is trained only once. We stress, however, that total compute is not the primary consideration: the decisive operational advantage is the per-instance latency ($\sim$32~ms versus 229~s), which enables the real-time onboard replanning that per-instance exact and metaheuristic solvers cannot provide at any budget.

\subsection{Computational efficiency}
\label{subsec:efficiency}

Table~\ref{tab:latency} summarizes the computational cost of all methods, separating one-time training from per-instance planning (inference for the learned policy, optimization for the exact and metaheuristic solvers), measured as described in Section~\ref{subsec:implementation}.

Among the heuristics, simple graph-based methods (\texttt{Warnsdorff}, \texttt{STC}) execute sequentially on the CPU in under 1\,ms, while sweep-family methods that require geometric axis-selection subproblems take 27--31\,ms. In contrast, the RL policy leverages GPU acceleration. Under greedy decoding, the neural forward pass requires 30.6\,ms per instance. While this is comparable to the slower CPU heuristics, the neural policy is solving a strictly harder, tightly constrained Hamiltonian path problem rather than a relaxed coverage sweep. Crucially, Best-of-16 sampling adds only 0.6\,ms over the greedy baseline (31.2\,ms total) because all 16 rollouts are processed simultaneously in a single batched GPU operation. The subsequent 2-opt refinement adds a further 1.0\,ms (32.2\,ms total), as its $\mathcal{O}(n^2)$ inner loop over short paths ($|\mathcal{V}|\leq 46$) is highly optimized via Numba JIT compilation.

Although comparing CPU-bound heuristics with GPU-accelerated neural policies involves different hardware paradigms, it accurately reflects modern operational realities. Modern autonomous surface vehicles are increasingly equipped with edge AI accelerators (e.g., NVIDIA Jetson modules) designed specifically to host such neural workloads. At approximately 32\,ms per instance, all three RL inference modes operate orders of magnitude faster than the typical decision and replanning cycles of maritime autonomous systems, which range from seconds to minutes, fully validating the feasibility of real-time onboard deployment.
 
\begin{table}[!t]
\small\centering
\caption{Computational cost of all methods, separating one-time (offline) cost
from per-instance (online) planning cost. RL inference is measured on a laptop
GPU; classical heuristics, the DFS feasibility audit, and the exact/metaheuristic
solvers on CPU. The MILP and GA are granted a 300\,s per-instance budget (mean
solving time reported). Training is a one-time cost, amortized across all
instances.}
\label{tab:latency}
\begin{tabular}{@{} l r r @{}}
\toprule
\textbf{Method} & \textbf{One-time cost} & \textbf{Per-instance cost} \\
\midrule
Warnsdorff                & --- & 0.4\,ms \\
STC\_Tree\_Coverage       & --- & 0.4\,ms \\
STC\_like                 & --- & 0.4\,ms \\
DFS\_Backtrack            & --- & 0.6\,ms \\
Morton\_Zorder            & --- & 0.7\,ms \\
Sweep\_Boundary\_Peel     & --- & 0.7\,ms \\
Boundary\_Spiral\_inward  & --- & 1.1\,ms \\
Boundary\_Spiral\_outward & --- & 1.2\,ms \\
Sweep\_Segment\_Snake     & --- & 26.7\,ms \\
Sweep\_Boustrophedon      & --- & 27.1\,ms \\
Sweep\_Row\_OneWay        & --- & 29.2\,ms \\
Sweep\_Row\_Interleave    & --- & 29.8\,ms \\
Sweep\_Segment\_Interl.   & --- & 31.5\,ms \\
\midrule
Exact\_DFS (feasibility)  & --- & 1.76\,s (dataset audit) \\
MILP\_CPSAT               & --- & 229\,s (300\,s budget) \\
GA                        & --- & 179\,s (300\,s budget) \\
\midrule
RL-Greedy (GPU)           & \multirow{3}{*}{$\approx$440\,GPU-h} & 30.6\,ms \\
RL-BoK16 (GPU)            &                                     & 31.2\,ms \\
RL-BoK16+2opt (GPU)       &                                     & 32.2\,ms \\
\bottomrule
\end{tabular}
\end{table}

\subsection{Qualitative path visualization}
\label{subsec:visualization}
 
\begin{figure*}[!t]
\centering
\newcommand{\subfigwidth}{0.19\textwidth}

\begin{subfigure}[t]{\subfigwidth}
\centering
\includegraphics[width=\linewidth]{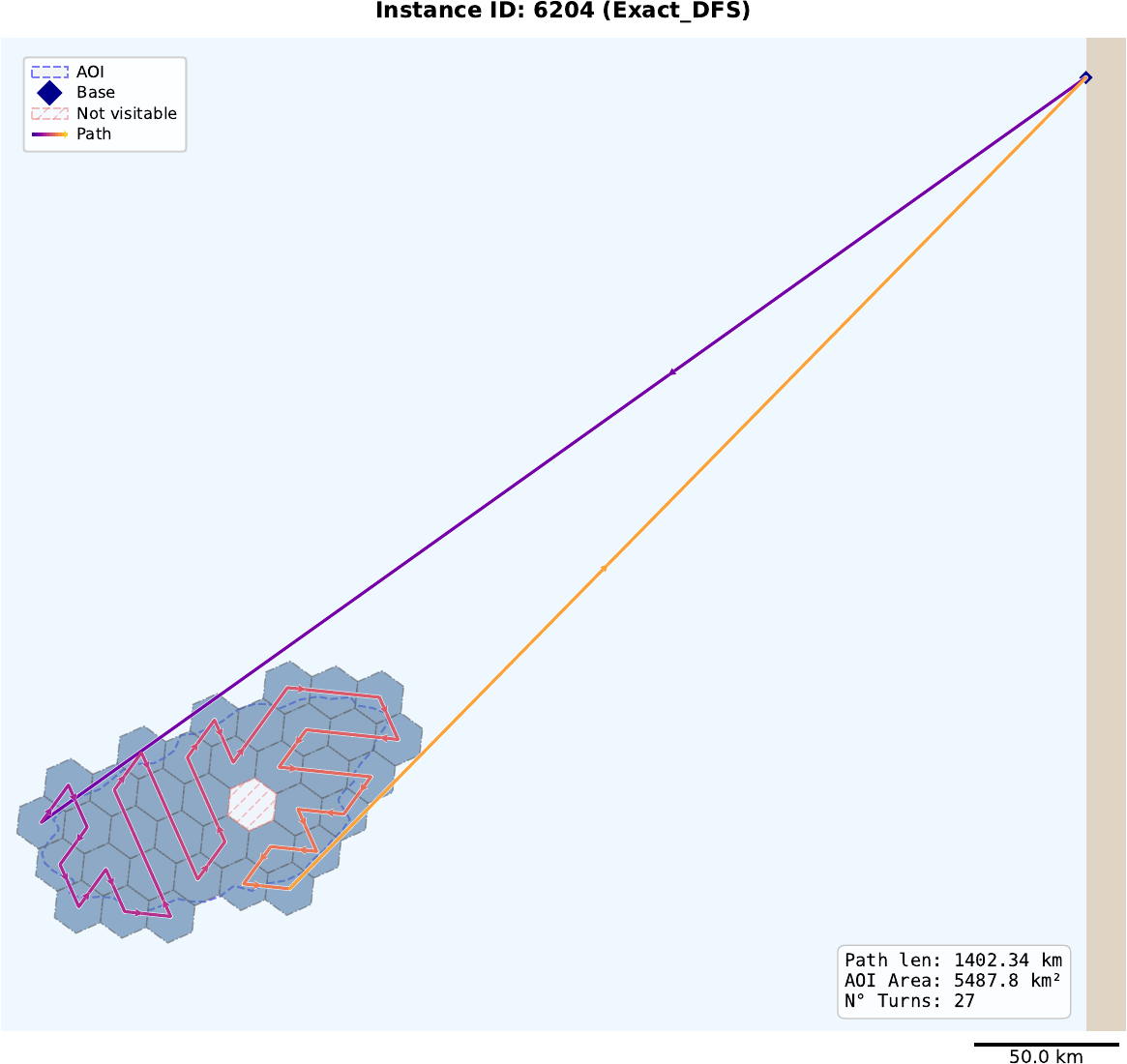}
\caption{\scriptsize Exact DFS\textsuperscript{H}}
\label{fig:qual_A_dfs}
\end{subfigure}\hfill
\begin{subfigure}[t]{\subfigwidth}
\centering
\includegraphics[width=\linewidth]{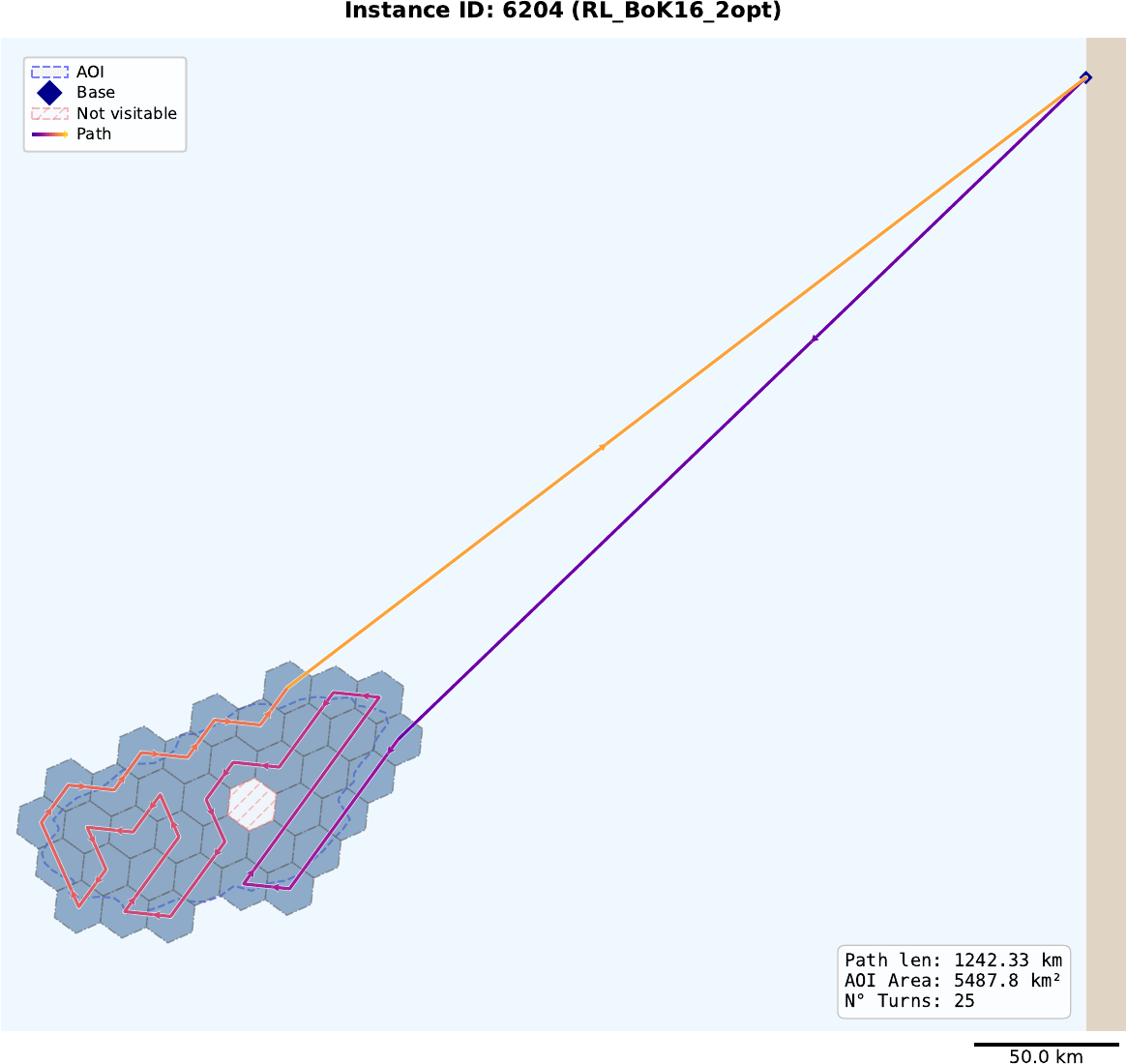}
\caption{\scriptsize RL-BoK16+2opt}
\label{fig:qual_A_rl}
\end{subfigure}\hfill
\begin{subfigure}[t]{\subfigwidth}
\centering
\includegraphics[width=\linewidth]{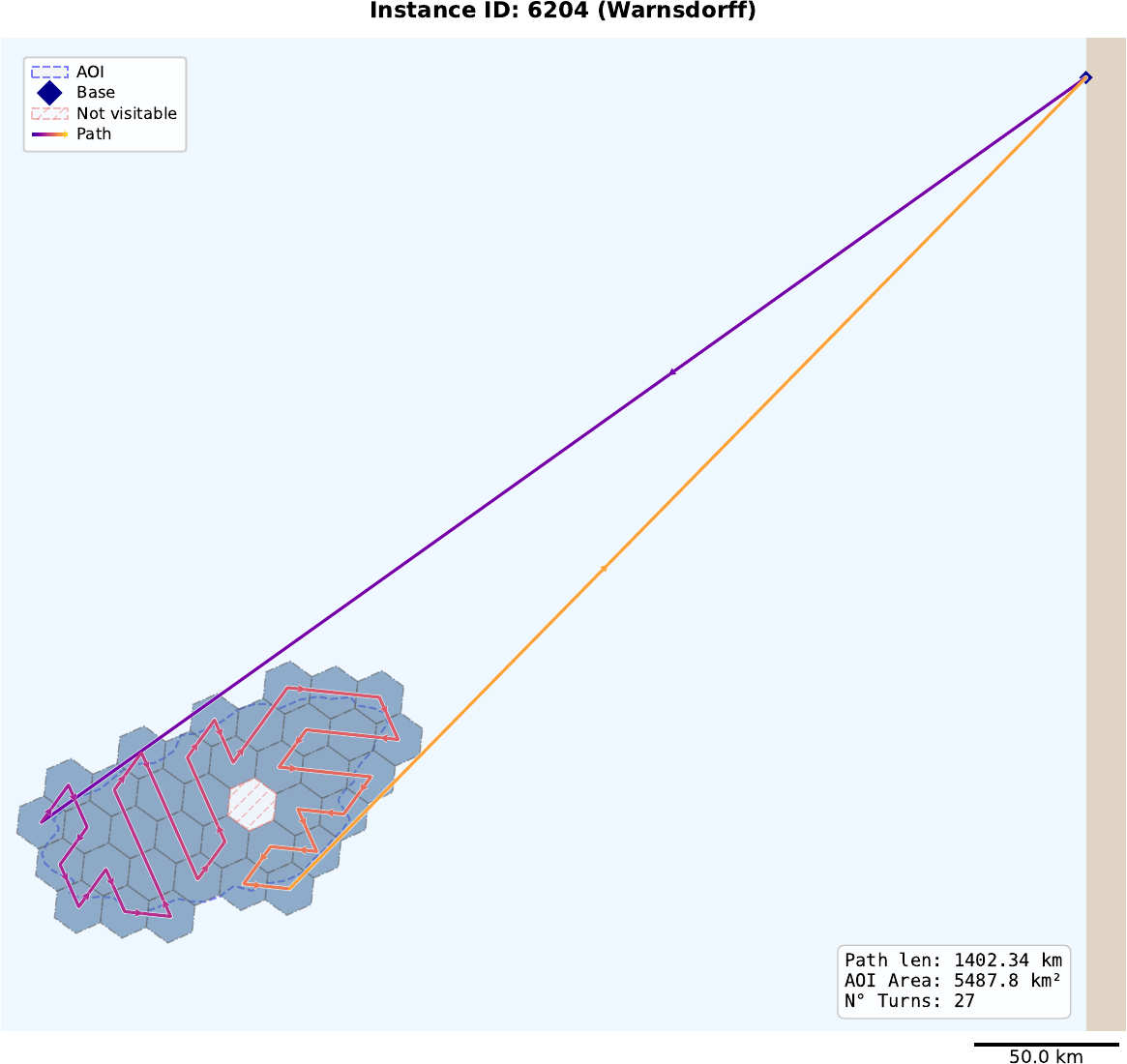}
\caption{\scriptsize Warnsdorff}
\label{fig:qual_A_warn}
\end{subfigure}\hfill
\begin{subfigure}[t]{\subfigwidth}
\centering
\includegraphics[width=\linewidth]{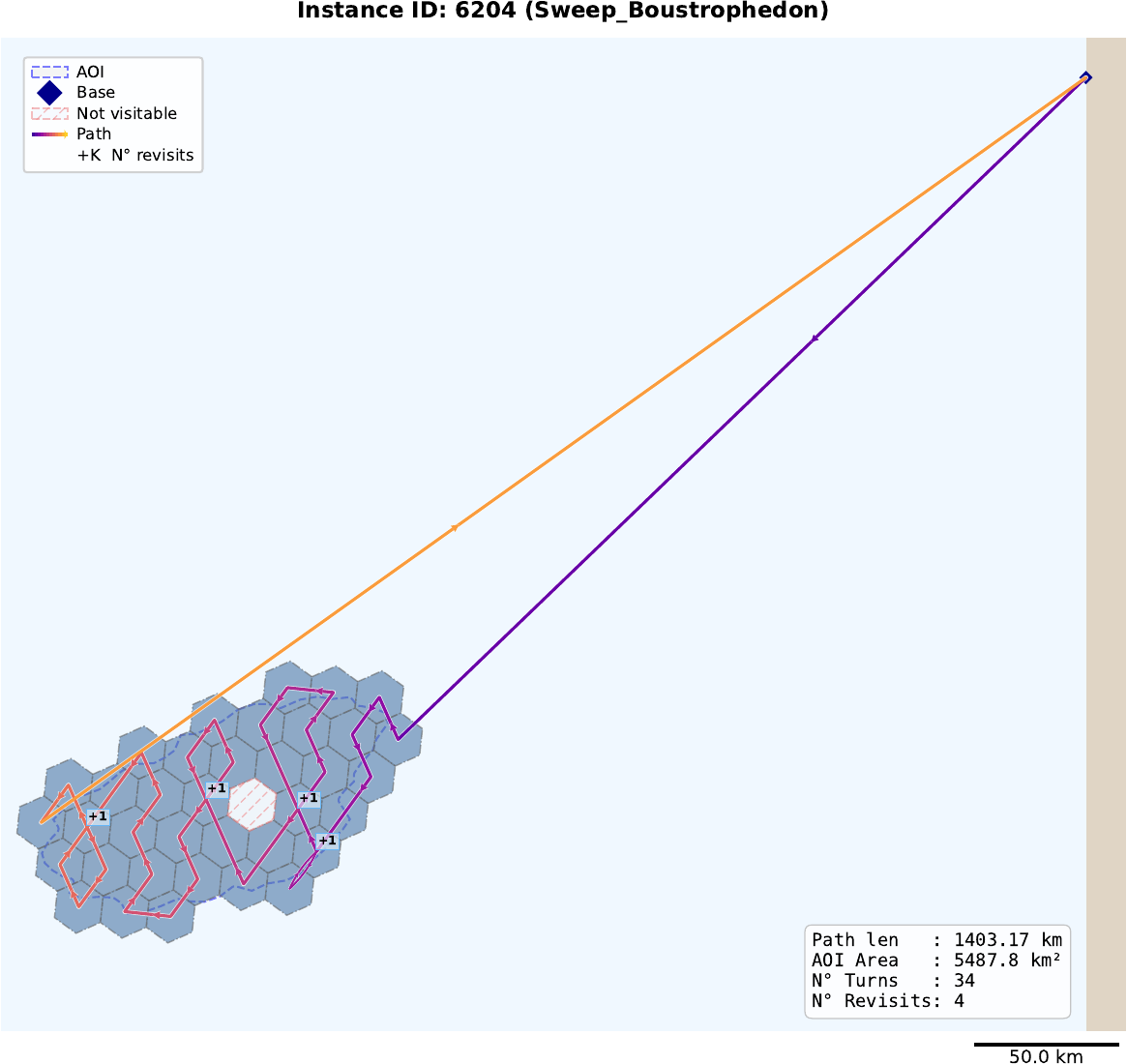}
\caption{\scriptsize Boustrophedon}
\label{fig:qual_A_bous}
\end{subfigure}\hfill
\begin{subfigure}[t]{\subfigwidth}
\centering
\includegraphics[width=\linewidth]{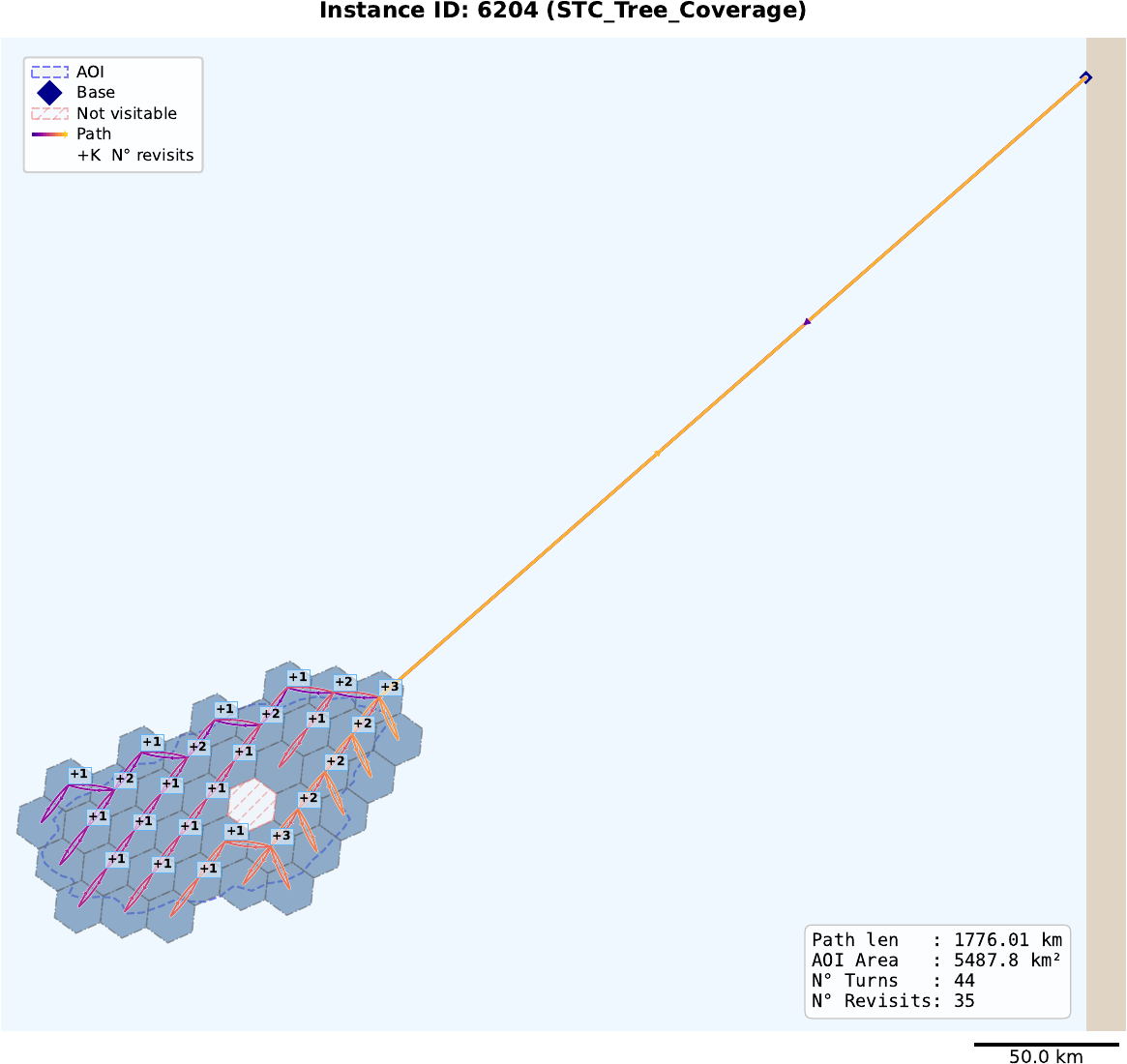}
\caption{\scriptsize STC}
\label{fig:qual_A_stc}
\end{subfigure}

\vspace{0.15cm}

\begin{subfigure}[t]{\subfigwidth}
\centering
\includegraphics[width=\linewidth]{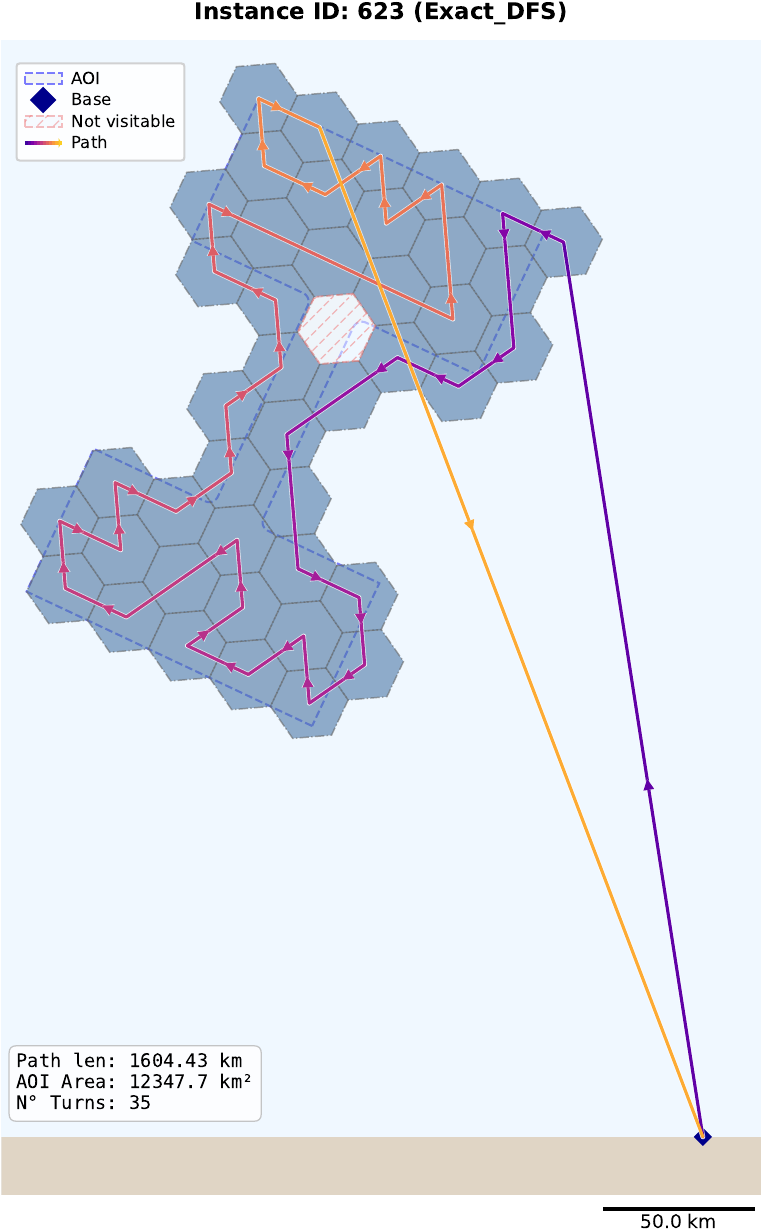}
\caption{\scriptsize Exact DFS\textsuperscript{H}}
\label{fig:qual_B_dfs}
\end{subfigure}\hfill
\begin{subfigure}[t]{\subfigwidth}
\centering
\includegraphics[width=\linewidth]{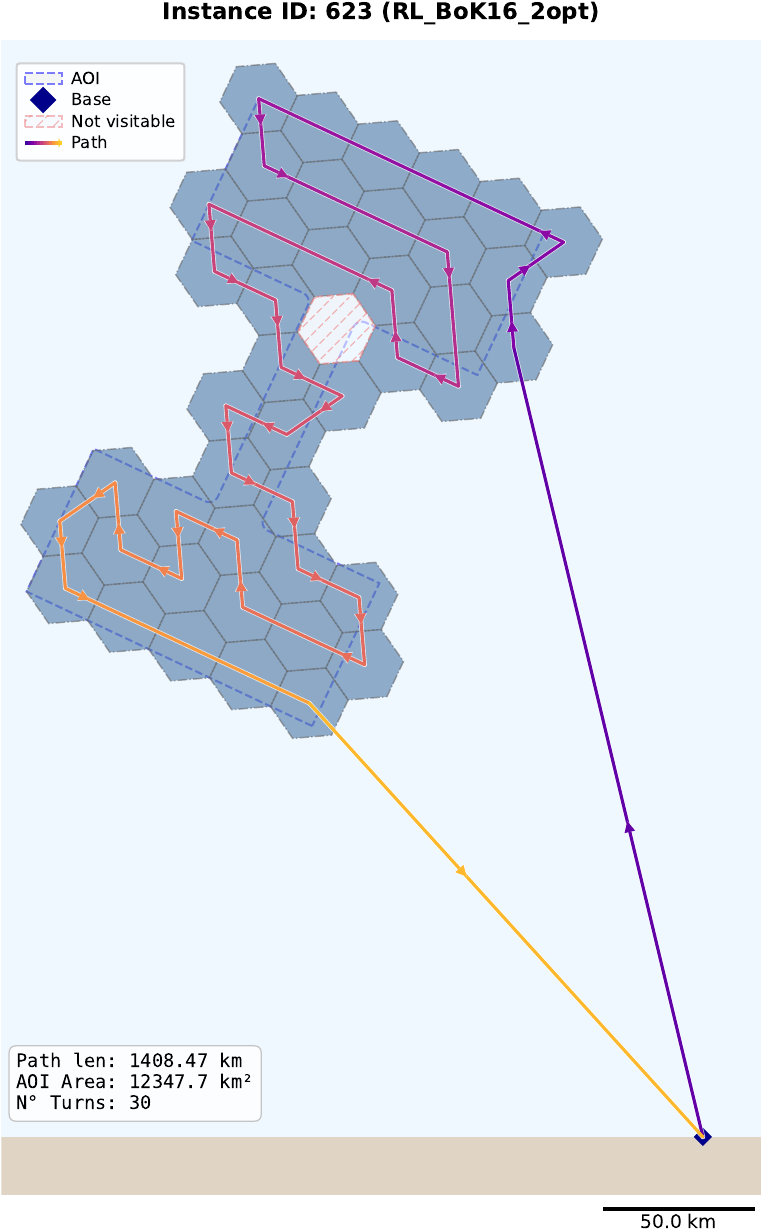}
\caption{\scriptsize RL-BoK16+2opt}
\label{fig:qual_B_rl}
\end{subfigure}\hfill
\begin{subfigure}[t]{\subfigwidth}
\centering
\includegraphics[width=\linewidth]{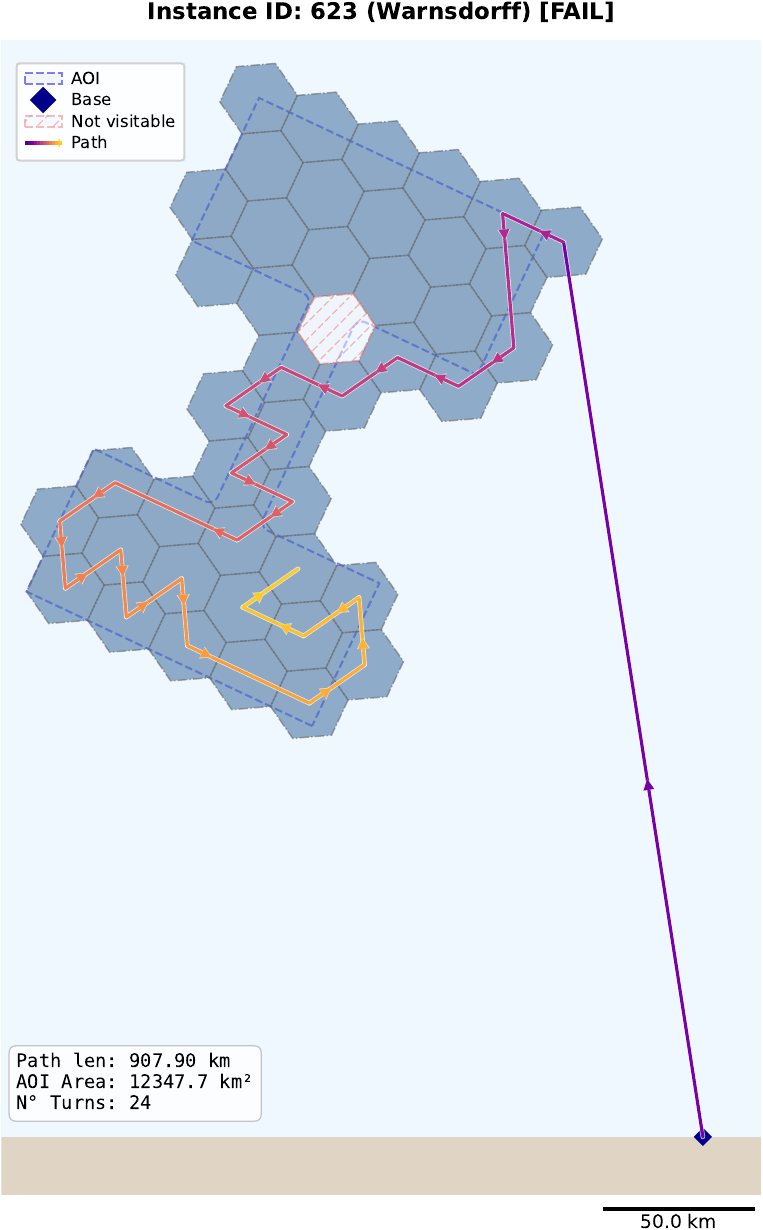}
\caption{\scriptsize Warnsdorff}
\label{fig:qual_B_warn}
\end{subfigure}\hfill
\begin{subfigure}[t]{\subfigwidth}
\centering
\includegraphics[width=\linewidth]{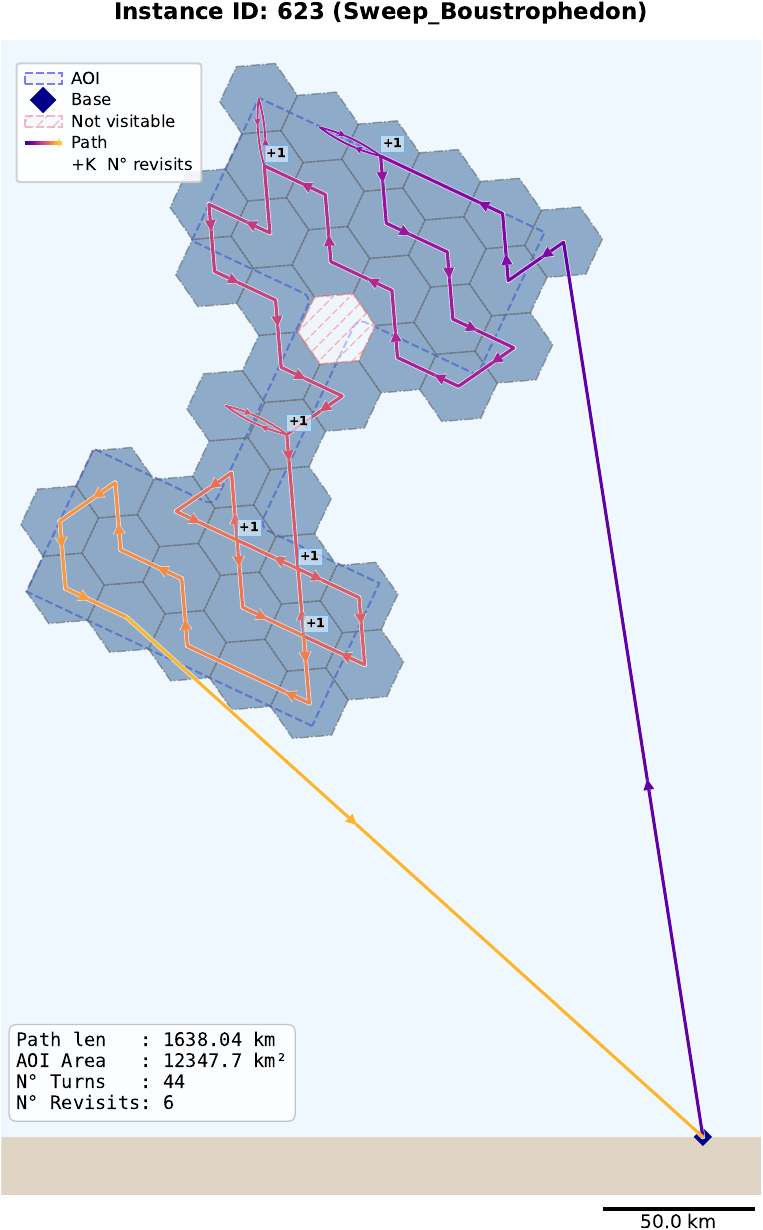}
\caption{\scriptsize Boustrophedon}
\label{fig:qual_B_bous}
\end{subfigure}\hfill
\begin{subfigure}[t]{\subfigwidth}
\centering
\includegraphics[width=\linewidth]{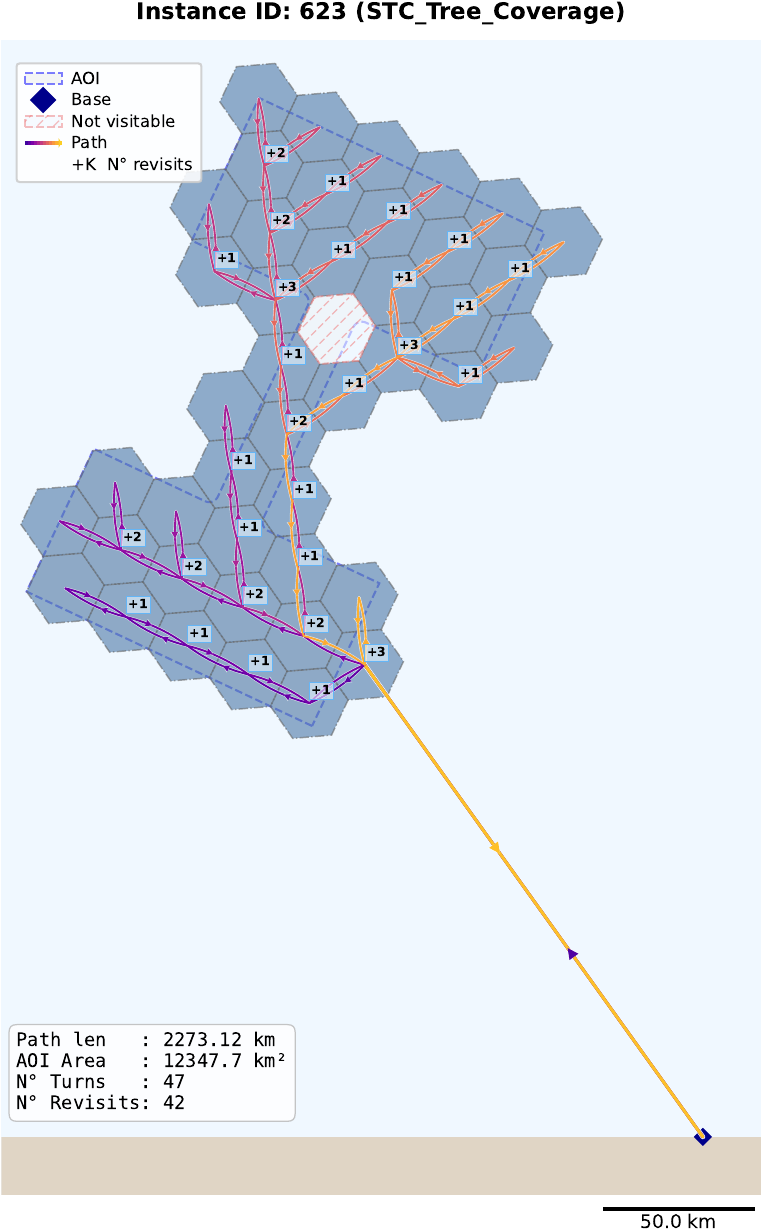}
\caption{\scriptsize STC}
\label{fig:qual_B_stc}
\end{subfigure}

\vspace{0.15cm}

\begin{subfigure}[t]{\subfigwidth}
\centering
\includegraphics[width=\linewidth]{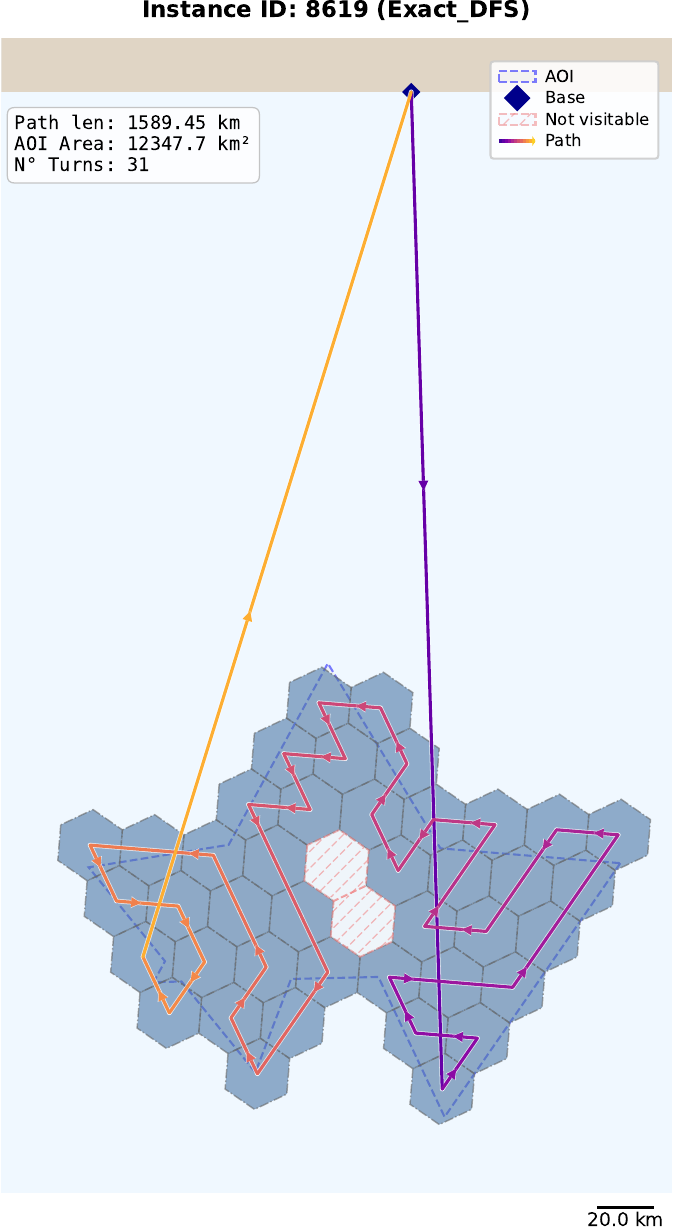}
\caption{\scriptsize Exact DFS\textsuperscript{H}}
\label{fig:qual_C_dfs}
\end{subfigure}\hfill
\begin{subfigure}[t]{\subfigwidth}
\centering
\includegraphics[width=\linewidth]{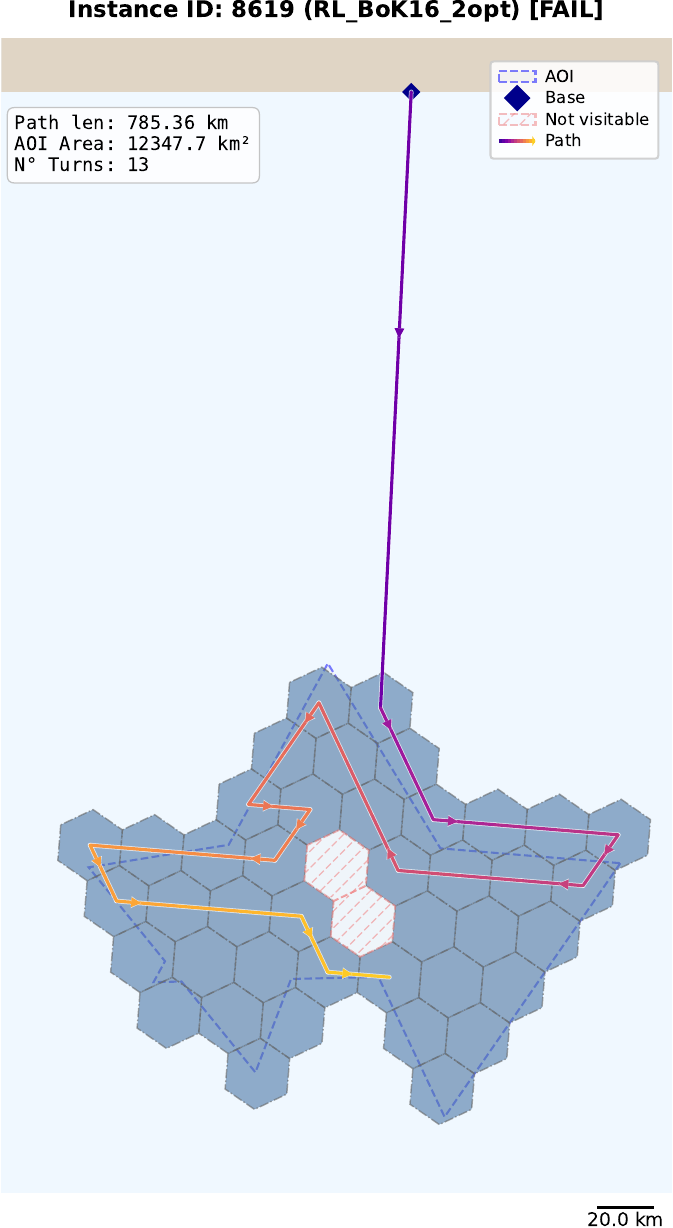}
\caption{\scriptsize RL-BoK16+2opt}
\label{fig:qual_C_rl}
\end{subfigure}\hfill
\begin{subfigure}[t]{\subfigwidth}
\centering
\includegraphics[width=\linewidth]{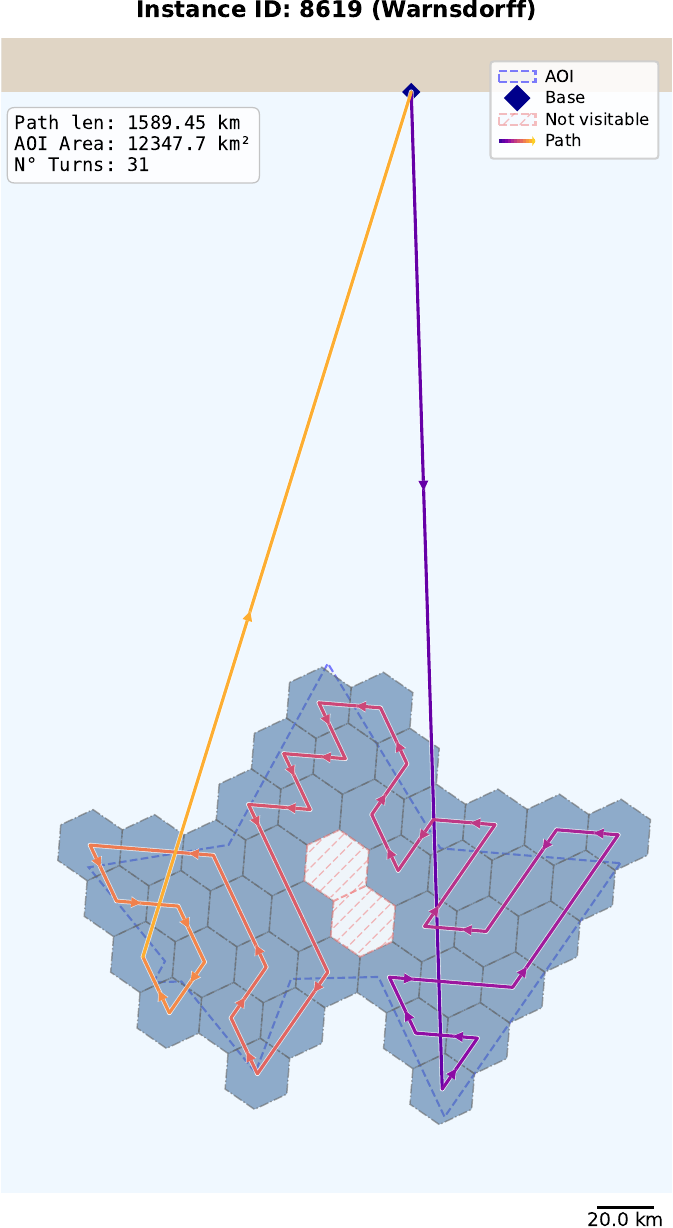}
\caption{\scriptsize Warnsdorff}
\label{fig:qual_C_warn}
\end{subfigure}\hfill
\begin{subfigure}[t]{\subfigwidth}
\centering
\includegraphics[width=\linewidth]{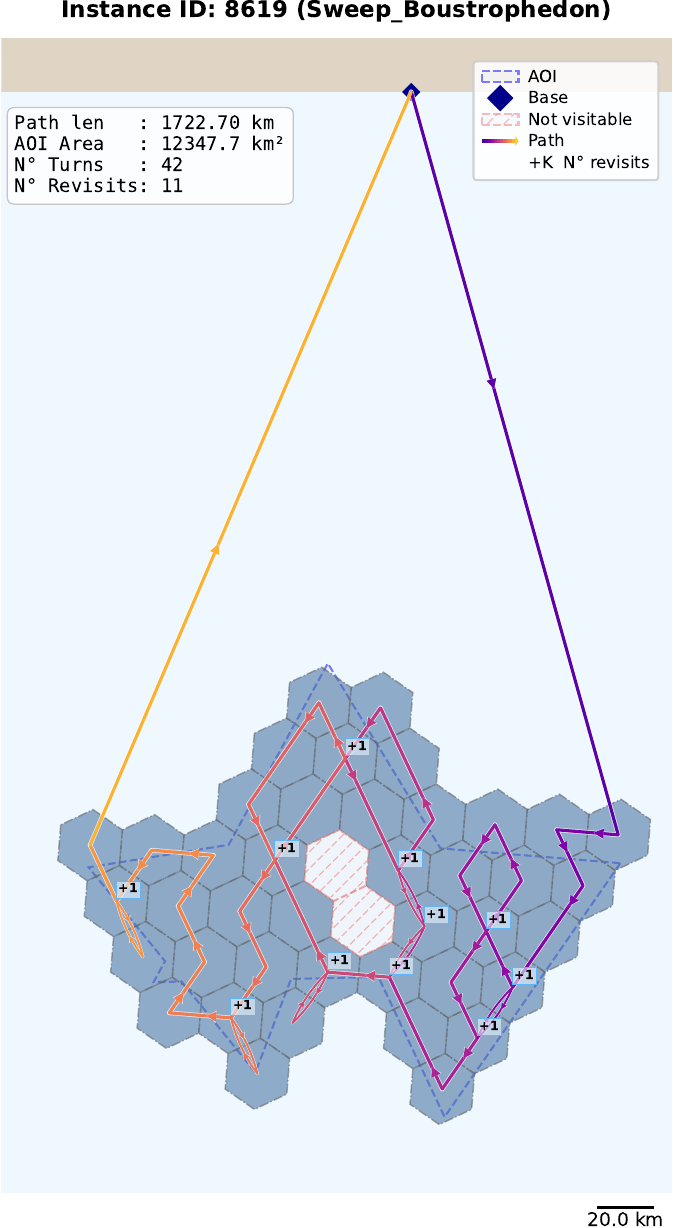}
\caption{\scriptsize Boustrophedon}
\label{fig:qual_C_bous}
\end{subfigure}\hfill
\begin{subfigure}[t]{\subfigwidth}
\centering
\includegraphics[width=\linewidth]{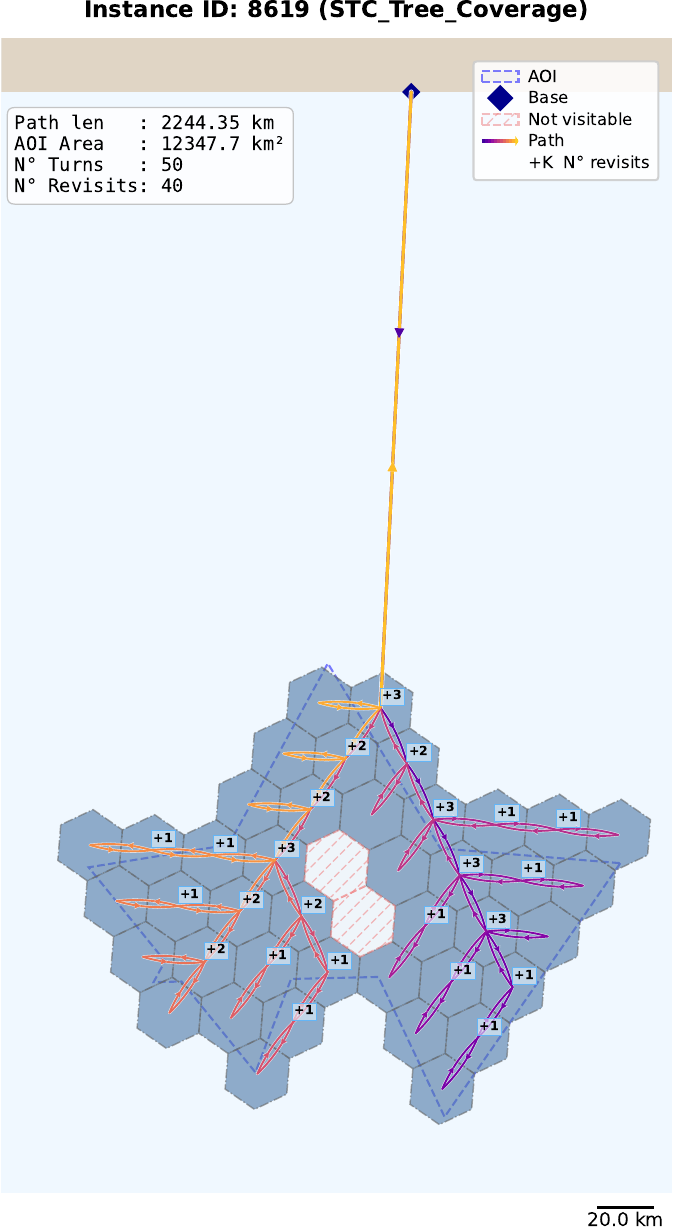}
\caption{\scriptsize STC}
\label{fig:qual_C_stc}
\end{subfigure}

\caption{Comparative path visualizations across three representative maritime topologies. The leftmost column displays the Exact DFS, denoted by \textsuperscript{H} for Hamiltonian, serving as a topological feasibility reference. Row~1 shows a small open-water configuration where all methods succeed, although \texttt{STC} introduces dense overlaps. Row~2 presents an H-shaped coastal corridor; here, the RL policy achieves a smooth trajectory, whereas greedy local heuristics such as \texttt{Warnsdorff} fail to escape the resulting geometric dead-end. Row~3 illustrates a generic irregular shape, where the RL policy identifies a valid single-visit sequence that avoids the costly kinematic reversals typical of \texttt{Boustrophedon} sweeps.}
\label{fig:qualitative}
\end{figure*}

Fig.~\ref{fig:qualitative} presents a comparative visualization of selected methods on representative test instances.  The instances were chosen to illustrate three characteristic scenarios: (a)~a compact AOI with a central obstacle, (b)~an elongated corridor with narrow passages, and (c)~an irregular shape with multiple peninsulas.
 
Across all scenarios, the RL policy produces visually smoother trajectories with fewer sharp heading changes, consistent with the quantitative turn reduction reported in Section~\ref{subsec:quality}.  The \texttt{Boustrophedon} pattern, while achieving complete coverage, exhibits the characteristic zigzag structure with frequent 180$^\circ$ reversals.  \texttt{Warnsdorff} produces relatively smooth paths when it succeeds, but fails entirely on instances (b) and (c) where narrow passages create topological traps.
The \texttt{STC} methods produce the most redundant paths, with extensive backtracking visible as dense, overlapping segments. Overall, these qualitative visualizations consistently reflect the quantitative trends observed across the entire 1{,}000-instance test set, confirming the RL policy's superior ability to adapt to complex maritime topologies without relying on redundant motion.
 
\subsection{Study limitations}
\label{subsec:limitations}

The present study should be interpreted with three scope limitations in mind. First, the proposed RL policy is optimized for the strict no-revisit Hamiltonian formulation, whereas most classical baselines are naturally designed for revisit-allowed complete coverage. For this reason, Hamiltonian feasibility and relaxed coverage completion are reported separately, and route-quality metrics are evaluated conditionally on common solved subsets.

Second, the experiments are conducted on a large synthetic dataset of irregular maritime AOIs. Although the generator is calibrated to operationally meaningful area scales, obstacle densities, and offshore base distances, transfer to real nautical charts with shoreline artifacts, bathymetric constraints, and sensor uncertainty remains to be validated.

Third, the training results reported here correspond to a single full training run due to the substantial computational cost of the end-to-end DRL pipeline. Consequently, the reported uncertainty reflects test-set variability rather than run-to-run optimization variability. Extending the study to multiple independent training seeds is a natural next step for future work.

Fourth, the learned policy does not outperform exact optimization in path length: within the operational time budget, the CP-SAT MILP attains marginally shorter tours and fewer turns. The contribution of the proposed method is therefore not solution optimality but near-MILP quality at inference latencies compatible with real-time onboard replanning, a regime the exact and metaheuristic solvers do not reach. Closing the residual quality gap to the exact optimum, for instance through learned 2-opt-style refinement aligned with the turn-aware objective, remains open.
\section{Conclusion and Future Work}
\label{sec:conclusion}

This paper introduced a Deep Reinforcement Learning framework for Maritime Coverage Path Planning on irregular hexagonal grids. By framing the problem as a neural combinatorial optimization task and training a Transformer-based pointer policy via a critic-free Group-Relative Policy Optimization (GRPO) scheme, we bypassed the limitations of classical geometric decomposition methods. The incorporation of an early dead-end detection mechanism via BFS significantly sharpened credit assignment during training.

Extensive benchmarking over 1{,}000 unseen irregular test areas demonstrates that the learned policy, particularly under Best-of-16 sampling with adjacency-aware 2-opt refinement (\texttt{RL-BoK16+2opt}), achieves a 99.1\% Hamiltonian success rate, more than doubling the best classical heuristic (46.0\%) while producing paths 7\% shorter and with up to 24.1\% fewer heading changes than the closest heuristic baseline, all with zero node revisits. Inference latency of approximately 32~ms per instance on a laptop GPU confirms the viability of real-time onboard deployment on modern autonomous maritime platforms.

Beyond the heuristic baselines, we benchmarked the policy against a budgeted CP-SAT MILP and a memetic GA. The learned policy attains normalized distance within 0.6\% of the MILP optimum and statistically indistinguishable from the GA, while planning each instance roughly four orders of magnitude faster (32~ms versus 229~s and 179~s). A component ablation further confirmed that BFS dead-end detection is essential for feasibility (98.1\% to 24.8\% HSR when removed) and that geometric augmentation contributes a smaller generalization gain. Together, these results position the method not as a replacement for exact solvers in offline settings, but as the practical choice when near-optimal coverage must be planned within real-time operational constraints.

Several extensions of this work merit investigation. First, \emph{non-uniform coverage priorities} can be incorporated by activating the \texttt{hexscore} field already embedded in the node features, enabling time-critical surveillance where high-value zones (e.g., distress areas informed by drift models) are visited preferentially. Second, \emph{multi-platform coordination} can be addressed by partitioning the AOI graph and assigning sub-tours to heterogeneous assets, extending the single-agent formulation to multi-agent fleet-level planning. Third, \emph{node revisitation} under a time budget, architecturally supported but not evaluated in the present study, could improve coverage robustness on highly constrained topologies where strict single-visit Hamiltonian paths are fragile. Fourth, \emph{scaling to larger AOIs} (e.g., hyper-resolution maritime grids with 100+ cells) will involve exploring architectural variants, such as alternative attention mechanisms, sparse graph representations, or region-level hierarchical decomposition, to further accelerate training and inference. Finally, \emph{validation on real maritime charts} (e.g., Chilean archipelago coastlines) and \emph{deployment on autonomous surface vehicles} under realistic oceanographic conditions and sensor models represent the natural applied extension of this framework toward operational Maritime Domain Awareness.

\section*{CRediT authorship contribution statement}
\textbf{Carlos S. Sepúlveda:} Conceptualization, Methodology, Software, Formal analysis, Investigation, Writing – original draft, Writing – review \& editing.
\textbf{Gonzalo A. Ruz:} Supervision, Validation, Writing – review \& editing.
\section*{Declaration of competing interest}
The authors declare that they have no known competing financial interests or personal relationships that could have appeared to influence the work reported in this paper.

\section*{Data availability}
The synthetic benchmark instances, trained model checkpoint, evaluation scripts, and baseline implementations required to reproduce the results reported in this study are available from the corresponding author upon reasonable request.

\section*{Acknowledgments}
This work is also supported by the Chilean Navy through the Directorate of Programs, Research and Development (Armada de Chile), which provided the necessary time and authorization for the development of this research.
The authors thank ANID FONDECYT 1230315, ANID-MILENIO-NCN2024\_103, ANID-MILENIO-NCN2024\_047, and Centro de Modelamiento Matemático (CMM) FB210005, BASAL funds for centers of excellence from ANID-Chile, and the ANID Doctorado Nacional Scholarship, grant number 21210465.

\appendix

\section{Taxonomy of Surveillance and Routing Problems}
\label{app:taxonomy}
Table~\ref{tab:taxonomy} summarizes the main families of coverage, patrolling, and routing problems underpinning our approach.
\captionsetup[table]{skip=4pt}
\begin{table*}[htbp]
  \small
  \centering
  \setlength{\tabcolsep}{3pt}%
  \rowcolors{2}{rowgray}{white}%
  \caption{Taxonomy of surveillance-related coverage, patrolling, and routing problems underpinning our CPP formulation on hexagonal grids.}
  \label{tab:taxonomy}
  \begin{tabularx}{\textwidth}{%
      p{0.19\textwidth} 
      Y                 
      Y                 
  }
    \toprule
    \textbf{Family} &
    \textbf{Typical objective and common formulation} &
    \textbf{Relation to this work and representative references} \\
    \midrule

    \textit{Region coverage (CPP)} &
    Achieve complete coverage of a known area of interest (AOI) while minimising path length, time, or energy. Classical CPP uses cellular decompositions (boustrophedon, trapezoidal, convex) and sweep patterns (e.g., lawnmower), extended to multi-robot and UAV settings. &
    Our hexagonal AOI discretisation is a discrete CPP formulation in which each hex cell represents a region of the maritime AOI. We follow CPP surveys in robotics and UAVs \citep{choset2001coverage,Galceran2013,cabreira2019,Fevgas2022} and connect to hex-based CPP and SAR planning in maritime contexts \citep{Azpurua2018,Cho2021,Kadioglu2019}, and to recent USV-based coastal and offshore survey coverage and cooperative multi-vehicle formulations \citep{zhao2024optimal,zhao2024joint,shen2025multiple,ma2026fault}. \\

    \textit{Sweep and barrier coverage} &
    Ensure that every point of a region or a virtual barrier is periodically visited by mobile sensors, subject to revisit intervals or detection constraints. Models often rely on TSP/graph tours, Eulerian paths, or approximation algorithms for large-scale sweep coverage and barrier placement. &
    When maritime AOIs degenerate into corridors (routes, shorelines, chokepoints), our hex-grid CPP reduces to sweep or barrier-type problems over 1D/2D structures. Sweep-coverage and barrier-coverage results in WSNs \citep{Li2011,Gorain2014,Benahmed2019,Nguyen2018,Li2019,Kong2016} motivate modeling “barriers’’ or critical routes via specialized boundary constraints and periodic revisit scheduling. \\

    \textit{Patrolling and persistent surveillance} &
    Maintain high-quality surveillance over long horizons by minimizing maximal idle time, maximizing information gain, or enforcing revisit-frequency constraints under vehicle range and kinematic limitations. Formulations often use infinite-horizon control or MILP-based routing with information metrics. &
    Weighted coverage objective can be interpreted as a single-visit proxy for persistent-surveillance value on hex-grids. Patrolling and persistent-surveillance works with UAVs and surface vehicles \citep{Nigam2009,Zuo2020,Bandarupalli2021,Savkin2019,Yanez2020,Luis2021b} motivate the use of information-based metrics that could inform future extensions with non-uniform priorities. \\

    \textit{Routing for surveillance and mission planning} &
    Plan routes for one or multiple platforms that visit regions or tasks while minimizing mission cost (time, fuel) and respecting range, time windows, and motion constraints. Typical formulations are TSP/VRP variants, orienteering, and Dubins-type routing, often solved via MILP or heuristics. &
    Our approach inherits the routing structure of maritime surveillance mission-planning models: hex cells act as “customers’’ to be visited, and tours must satisfy sensor and endurance constraints. Classical radar/SAR routing and UAV mission-planning work \citep{Panton1999,John2001,Grob2006,Karasakal2016,Coutinho2018,Otto2018,Cho2021,Cho2021b} serve as baselines to benchmark our learned policies. \\

    \textit{Spatial crowdsourcing and task assignment} &
    Assign spatial sensing or monitoring tasks to agents (humans, vehicles, sensors) to maximise utility or coverage subject to capacity, location, and temporal constraints. Models are often matching or assignment problems with uncertainty in agent availability and task locations. &
    In multi-platform maritime surveillance, our hex-based CPP can be extended with task-assignment layers that allocate subsets of hex cells or sub-tours to heterogeneous assets. Spatial crowdsourcing and mobile-sensing literature \citep{Wu2019,Bhatti2021,Tong2020,ZhouZhen2019,Chen2020} informs potential extensions where different vehicles or sensor types share the same hex-grid representation. \\

    \textit{Learning-based combinatorial optimisation (NCO / RL4CO)} &
    Learn parametric policies that output near-optimal solutions for routing, scheduling, or CPP problems, enabling reuse across instances without re-solving from scratch. Approaches include pointer networks and attention-based models trained with RL or supervised signals on graph-structured inputs. &
    Our transformer-like pointer policy over hex-graphs belongs to this family. We adapt ideas from attention-based routing and RL4CO \citep{Vinyals2015,Bello2016,Nazari2018,Kool2018,Xin2021,Li2021a,berto2025rl4co,darvariu2024graph} and combine them with advanced training strategies for long-horizon coverage on hex-grids, bridging maritime CPP with neural combinatorial optimisation. \\
    \bottomrule
  \end{tabularx}
\end{table*}

\bibliographystyle{elsarticle-harv}
\bibliography{references}

\begin{thebibliography}{83}
\expandafter\ifx\csname natexlab\endcsname\relax\def\natexlab#1{#1}\fi
\providecommand{\url}[1]{\texttt{#1}}
\providecommand{\href}[2]{#2}
\providecommand{\path}[1]{#1}
\providecommand{\DOIprefix}{doi:}
\providecommand{\ArXivprefix}{arXiv:}
\providecommand{\URLprefix}{URL: }
\providecommand{\Pubmedprefix}{pmid:}
\providecommand{\doi}[1]{\href{http://dx.doi.org/#1}{\path{#1}}}
\providecommand{\Pubmed}[1]{\href{pmid:#1}{\path{#1}}}
\providecommand{\bibinfo}[2]{#2}
\ifx\xfnm\relax \def\xfnm[#1]{\unskip,\space#1}\fi
\bibitem[{Ai et~al.(2021)Ai, Jia, Xu, Xu, Wen, Li and Zhang}]{Ai2021}
\bibinfo{author}{Ai, B.}, \bibinfo{author}{Jia, M.}, \bibinfo{author}{Xu, H.}, \bibinfo{author}{Xu, J.}, \bibinfo{author}{Wen, Z.}, \bibinfo{author}{Li, B.}, \bibinfo{author}{Zhang, D.}, \bibinfo{year}{2021}.
\newblock \bibinfo{title}{Coverage path planning for maritime search and rescue using reinforcement learning}.
\newblock \bibinfo{journal}{Ocean Engineering} \bibinfo{volume}{241}.
\newblock \DOIprefix\doi{10.1016/j.oceaneng.2021.110098}.
\bibitem[{Alpdemir(2022)}]{Alpdemir2022}
\bibinfo{author}{Alpdemir, M.N.}, \bibinfo{year}{2022}.
\newblock \bibinfo{title}{Tactical uav path optimization under radar threat using deep reinforcement learning}.
\newblock \bibinfo{journal}{Neural Computing and Applications} \bibinfo{volume}{34}, \bibinfo{pages}{5649--5664}.
\bibitem[{Azad et~al.(2017)Azad, Islam and Chakraborty}]{Azad2017}
\bibinfo{author}{Azad, A.S.}, \bibinfo{author}{Islam, M.M.}, \bibinfo{author}{Chakraborty, S.}, \bibinfo{year}{2017}.
\newblock \bibinfo{title}{A heuristic initialized stochastic memetic algorithm for mdpvrp with interdependent depot operations}.
\newblock \bibinfo{journal}{IEEE Transactions on Cybernetics} \bibinfo{volume}{47}, \bibinfo{pages}{4302--4315}.
\newblock \DOIprefix\doi{10.1109/TCYB.2016.2607220}.
\bibitem[{Azpúrua et~al.(2018)Azpúrua, Freitas, Macharet and Campos}]{Azpurua2018}
\bibinfo{author}{Azpúrua, H.}, \bibinfo{author}{Freitas, G.M.}, \bibinfo{author}{Macharet, D.G.}, \bibinfo{author}{Campos, M.F.}, \bibinfo{year}{2018}.
\newblock \bibinfo{title}{Multi-robot coverage path planning using hexagonal segmentation for geophysical surveys}.
\newblock \bibinfo{journal}{Robotica} \bibinfo{volume}{36}, \bibinfo{pages}{1144--1166}.
\newblock \DOIprefix\doi{10.1017/S0263574718000292}.
\bibitem[{B{\"a}hnemann et~al.(2021)B{\"a}hnemann, Lawrance, Chung, Pantic, Siegwart and Nieto}]{bahnemann2021revisiting}
\bibinfo{author}{B{\"a}hnemann, R.}, \bibinfo{author}{Lawrance, N.}, \bibinfo{author}{Chung, J.J.}, \bibinfo{author}{Pantic, M.}, \bibinfo{author}{Siegwart, R.}, \bibinfo{author}{Nieto, J.}, \bibinfo{year}{2021}.
\newblock \bibinfo{title}{Revisiting boustrophedon coverage path planning as a generalized traveling salesman problem}, in: \bibinfo{booktitle}{Field and Service Robotics: Results of the 12th International Conference}, \bibinfo{organization}{Springer}. pp. \bibinfo{pages}{277--290}.
\bibitem[{Bandarupalli et~al.(2021)Bandarupalli, Swarup, Weston and Chaterji}]{Bandarupalli2021}
\bibinfo{author}{Bandarupalli, A.}, \bibinfo{author}{Swarup, D.}, \bibinfo{author}{Weston, N.}, \bibinfo{author}{Chaterji, S.}, \bibinfo{year}{2021}.
\newblock \bibinfo{title}{Persistent airborne surveillance using semi-autonomous drone swarms}, in: \bibinfo{booktitle}{Proceedings of the 7th Workshop on Micro Aerial Vehicle Networks, Systems, and Applications}, pp. \bibinfo{pages}{19--24}.
\newblock \DOIprefix\doi{10.1145/3469259.3470487}.
\bibitem[{Bello et~al.(2016)Bello, Pham, Le, Norouzi and Bengio}]{Bello2016}
\bibinfo{author}{Bello, I.}, \bibinfo{author}{Pham, H.}, \bibinfo{author}{Le, Q.V.}, \bibinfo{author}{Norouzi, M.}, \bibinfo{author}{Bengio, S.}, \bibinfo{year}{2016}.
\newblock \bibinfo{title}{Neural combinatorial optimization with reinforcement learning}.
\newblock \bibinfo{journal}{5th International Conference on Learning Representations, ICLR 2017 - Workshop Track Proceedings} .
\bibitem[{Benahmed and Benahmed(2019)}]{Benahmed2019}
\bibinfo{author}{Benahmed, T.}, \bibinfo{author}{Benahmed, K.}, \bibinfo{year}{2019}.
\newblock \bibinfo{title}{Optimal barrier coverage for critical area surveillance using wireless sensor networks}.
\newblock \bibinfo{journal}{International Journal of Communication Systems} \bibinfo{volume}{32}.
\newblock \DOIprefix\doi{10.1002/dac.3955}.
\bibitem[{Berto et~al.(2025)Berto, Hua, Park, Luttmann, Ma, Bu, Wang, Ye, Kim, Choi et~al.}]{berto2025rl4co}
\bibinfo{author}{Berto, F.}, \bibinfo{author}{Hua, C.}, \bibinfo{author}{Park, J.}, \bibinfo{author}{Luttmann, L.}, \bibinfo{author}{Ma, Y.}, \bibinfo{author}{Bu, F.}, \bibinfo{author}{Wang, J.}, \bibinfo{author}{Ye, H.}, \bibinfo{author}{Kim, M.}, \bibinfo{author}{Choi, S.}, et~al., \bibinfo{year}{2025}.
\newblock \bibinfo{title}{Rl4co: an extensive reinforcement learning for combinatorial optimization benchmark}, in: \bibinfo{booktitle}{Proceedings of the 31st ACM SIGKDD Conference on Knowledge Discovery and Data Mining V. 2}, pp. \bibinfo{pages}{5278--5289}.
\bibitem[{Bhatti et~al.(2021)Bhatti, Fan, Wang, Gao, Wu and Chen}]{Bhatti2021}
\bibinfo{author}{Bhatti, S.S.}, \bibinfo{author}{Fan, J.}, \bibinfo{author}{Wang, K.}, \bibinfo{author}{Gao, X.}, \bibinfo{author}{Wu, F.}, \bibinfo{author}{Chen, G.}, \bibinfo{year}{2021}.
\newblock \bibinfo{title}{An approximation algorithm for bounded task assignment problem in spatial crowdsourcing}.
\newblock \bibinfo{journal}{IEEE Transactions on Mobile Computing} \bibinfo{volume}{20}, \bibinfo{pages}{2536--2549}.
\newblock \DOIprefix\doi{10.1109/TMC.2020.2984380}.
\bibitem[{Biniaz et~al.(2017)Biniaz, Liu, Maheshwari and Smid}]{Biniaz2017}
\bibinfo{author}{Biniaz, A.}, \bibinfo{author}{Liu, P.}, \bibinfo{author}{Maheshwari, A.}, \bibinfo{author}{Smid, M.}, \bibinfo{year}{2017}.
\newblock \bibinfo{title}{Approximation algorithms for the unit disk cover problem in 2d and 3d}.
\newblock \bibinfo{journal}{Computational Geometry} \bibinfo{volume}{60}, \bibinfo{pages}{8--18}.
\newblock \DOIprefix\doi{10.1016/J.COMGEO.2016.04.002}.
\bibitem[{Bischof et~al.(2018)Bischof, Fontugne and Bustamante}]{bischof2018untangling}
\bibinfo{author}{Bischof, Z.S.}, \bibinfo{author}{Fontugne, R.}, \bibinfo{author}{Bustamante, F.E.}, \bibinfo{year}{2018}.
\newblock \bibinfo{title}{Untangling the world-wide mesh of undersea cables}, in: \bibinfo{booktitle}{Proceedings of the 17th ACM workshop on hot topics in networks}, pp. \bibinfo{pages}{78--84}.
\bibitem[{Boots et~al.(1999)Boots, Okabe and Sugihara}]{Boots1999}
\bibinfo{author}{Boots, B.}, \bibinfo{author}{Okabe, A.}, \bibinfo{author}{Sugihara, K.}, \bibinfo{year}{1999}.
\newblock \bibinfo{title}{Spatial tessellations}, in: \bibinfo{booktitle}{Geographical Information Systems}, pp. \bibinfo{pages}{503--526}.
\bibitem[{Boraz(2009)}]{Boraz2009}
\bibinfo{author}{Boraz, S.C.}, \bibinfo{year}{2009}.
\newblock \bibinfo{title}{Maritime domain awareness: Myths and realities}.
\newblock \bibinfo{journal}{Naval War College Review} \bibinfo{volume}{62}, \bibinfo{pages}{137--146}.
\newblock \DOIprefix\doi{10.2307/26397039}.
\bibitem[{Bueger et~al.(2024)Bueger, Edmunds and Stockbruegger}]{bueger2024securing}
\bibinfo{author}{Bueger, C.}, \bibinfo{author}{Edmunds, T.P.}, \bibinfo{author}{Stockbruegger, J.}, \bibinfo{year}{2024}.
\newblock \bibinfo{title}{Securing the Seas: A Comprehensive Assessment of Global Maritime Security}.
\newblock \bibinfo{type}{Technical Report}. United Nations.
\bibitem[{Cabreira et~al.(2019)Cabreira, Brisolara and Paulo}]{cabreira2019}
\bibinfo{author}{Cabreira, T.M.}, \bibinfo{author}{Brisolara, L.B.}, \bibinfo{author}{Paulo, R.F.}, \bibinfo{year}{2019}.
\newblock \bibinfo{title}{Survey on coverage path planning with unmanned aerial vehicles}.
\newblock \bibinfo{journal}{Drones} \bibinfo{volume}{3}, \bibinfo{pages}{1--38}.
\newblock \DOIprefix\doi{10.3390/drones3010004}.
\bibitem[{Chen et~al.(2015)Chen, Chang and Chen}]{Chen2015}
\bibinfo{author}{Chen, C.C.}, \bibinfo{author}{Chang, C.Y.}, \bibinfo{author}{Chen, P.Y.}, \bibinfo{year}{2015}.
\newblock \bibinfo{title}{Linear time approximation algorithms for the relay node placement problem in wireless sensor networks with hexagon tessellation}.
\newblock \bibinfo{journal}{Journal of Sensors} \bibinfo{volume}{2015}.
\newblock \DOIprefix\doi{10.1155/2015/565983}.
\bibitem[{Chen et~al.(2020)Chen, Wang, Ota, Dong, Zhao and Liu}]{Chen2020}
\bibinfo{author}{Chen, M.}, \bibinfo{author}{Wang, T.}, \bibinfo{author}{Ota, K.}, \bibinfo{author}{Dong, M.}, \bibinfo{author}{Zhao, M.}, \bibinfo{author}{Liu, A.}, \bibinfo{year}{2020}.
\newblock \bibinfo{title}{Intelligent resource allocation management for vehicles network: An a3c learning approach}.
\newblock \bibinfo{journal}{Computer Communications} \bibinfo{volume}{151}, \bibinfo{pages}{485--494}.
\newblock \DOIprefix\doi{10.1016/J.COMCOM.2019.12.054}.
\bibitem[{Cho et~al.(2021a)Cho, Park, Lee, Shim and Kim}]{Cho2021b}
\bibinfo{author}{Cho, S.W.}, \bibinfo{author}{Park, H.J.}, \bibinfo{author}{Lee, H.}, \bibinfo{author}{Shim, D.H.}, \bibinfo{author}{Kim, S.Y.}, \bibinfo{year}{2021}a.
\newblock \bibinfo{title}{Coverage path planning for multiple unmanned aerial vehicles in maritime search and rescue operations}.
\newblock \bibinfo{journal}{Computers and Industrial Engineering} \bibinfo{volume}{161}.
\newblock \DOIprefix\doi{10.1016/j.cie.2021.107612}.
\bibitem[{Cho et~al.(2021b)Cho, Park, Park and Kim}]{Cho2021}
\bibinfo{author}{Cho, S.W.}, \bibinfo{author}{Park, J.H.}, \bibinfo{author}{Park, H.J.}, \bibinfo{author}{Kim, S.}, \bibinfo{year}{2021}b.
\newblock \bibinfo{title}{Multi-uav coverage path planning based on hexagonal grid decomposition in maritime search and rescue}.
\newblock \bibinfo{journal}{Mathematics} \bibinfo{volume}{10}, \bibinfo{pages}{83}.
\bibitem[{Choi et~al.(2019)Choi, Choi, Briceno and Mavris}]{Choi2019}
\bibinfo{author}{Choi, Y.}, \bibinfo{author}{Choi, Y.}, \bibinfo{author}{Briceno, S.}, \bibinfo{author}{Mavris, D.N.}, \bibinfo{year}{2019}.
\newblock \bibinfo{title}{Multi-uas path-planning for a large-scale disjoint disaster management}, in: \bibinfo{booktitle}{2019 International Conference on Unmanned Aircraft Systems (ICUAS)}, pp. \bibinfo{pages}{799--807}.
\bibitem[{Choi et~al.(2020)Choi, Choi, Briceno and Mavris}]{Choi2020}
\bibinfo{author}{Choi, Y.}, \bibinfo{author}{Choi, Y.}, \bibinfo{author}{Briceno, S.}, \bibinfo{author}{Mavris, D.N.}, \bibinfo{year}{2020}.
\newblock \bibinfo{title}{Energy-constrained multi-uav coverage path planning for an aerial imagery mission using column generation}.
\newblock \bibinfo{journal}{Journal of Intelligent \& Robotic Systems} \bibinfo{volume}{97}, \bibinfo{pages}{125--139}.
\bibitem[{Choset(2001)}]{choset2001coverage}
\bibinfo{author}{Choset, H.}, \bibinfo{year}{2001}.
\newblock \bibinfo{title}{Coverage for robotics--a survey of recent results}.
\newblock \bibinfo{journal}{Annals of mathematics and artificial intelligence} \bibinfo{volume}{31}, \bibinfo{pages}{113--126}.
\bibitem[{{Consejo de Ministros para el desarrollo de una Política Oceánica Nacional}(2023)}]{programaocea}
\bibinfo{author}{{Consejo de Ministros para el desarrollo de una Política Oceánica Nacional}}, \bibinfo{year}{2023}.
\newblock \bibinfo{title}{Programa Oceánico Nacional: Plan Oceánico Sostenible Chile 2023}.
\newblock \bibinfo{type}{Technical Report}. Ministerio de Relaciones Exteriores. \bibinfo{address}{Valparaíso}.
\bibitem[{Coutinho et~al.(2018)Coutinho, Battarra and Fliege}]{Coutinho2018}
\bibinfo{author}{Coutinho, W.P.}, \bibinfo{author}{Battarra, M.}, \bibinfo{author}{Fliege, J.}, \bibinfo{year}{2018}.
\newblock \bibinfo{title}{The unmanned aerial vehicle routing and trajectory optimisation problem, a taxonomic review}.
\newblock \bibinfo{journal}{Computers \& Industrial Engineering} \bibinfo{volume}{120}, \bibinfo{pages}{116--128}.
\newblock \DOIprefix\doi{10.1016/J.CIE.2018.04.037}.
\bibitem[{Darvariu et~al.(2024)Darvariu, Hailes and Musolesi}]{darvariu2024graph}
\bibinfo{author}{Darvariu, V.A.}, \bibinfo{author}{Hailes, S.}, \bibinfo{author}{Musolesi, M.}, \bibinfo{year}{2024}.
\newblock \bibinfo{title}{Graph reinforcement learning for combinatorial optimization: A survey and unifying perspective}.
\newblock \bibinfo{journal}{arXiv preprint arXiv:2404.06492} .
\bibitem[{Dogancay et~al.(2021)Dogancay, Tu and Ibal}]{Dogancay2021}
\bibinfo{author}{Dogancay, K.}, \bibinfo{author}{Tu, Z.}, \bibinfo{author}{Ibal, G.}, \bibinfo{year}{2021}.
\newblock \bibinfo{title}{Research into vessel behaviour pattern recognition in the maritime domain: Past, present and future}.
\newblock \bibinfo{journal}{Digital Signal Processing} \bibinfo{volume}{119}, \bibinfo{pages}{103191}.
\newblock \DOIprefix\doi{10.1016/J.DSP.2021.103191}.
\bibitem[{Fevgas et~al.(2022)Fevgas, Lagkas, Argyriou and Sarigiannidis}]{Fevgas2022}
\bibinfo{author}{Fevgas, G.}, \bibinfo{author}{Lagkas, T.}, \bibinfo{author}{Argyriou, V.}, \bibinfo{author}{Sarigiannidis, P.}, \bibinfo{year}{2022}.
\newblock \bibinfo{title}{Coverage path planning methods focusing on energy efficient and cooperative strategies for unmanned aerial vehicles}.
\newblock \bibinfo{journal}{Sensors} \bibinfo{volume}{22}.
\newblock \DOIprefix\doi{10.3390/s22031235}.
\bibitem[{Galceran and Carreras(2013)}]{Galceran2013}
\bibinfo{author}{Galceran, E.}, \bibinfo{author}{Carreras, M.}, \bibinfo{year}{2013}.
\newblock \bibinfo{title}{A survey on coverage path planning for robotics}.
\newblock \bibinfo{journal}{Robotics and Autonomous systems} \bibinfo{volume}{61}, \bibinfo{pages}{1258--1276}.
\bibitem[{Gao et~al.(2022)Gao, Kang, Zhang, Liu and Zhao}]{Gao2022}
\bibinfo{author}{Gao, M.}, \bibinfo{author}{Kang, Z.}, \bibinfo{author}{Zhang, A.}, \bibinfo{author}{Liu, J.}, \bibinfo{author}{Zhao, F.}, \bibinfo{year}{2022}.
\newblock \bibinfo{title}{Mass autonomous navigation system based on ais big data with dueling deep q networks prioritized replay reinforcement learning}.
\newblock \bibinfo{journal}{Ocean Engineering} \bibinfo{volume}{249}.
\newblock \DOIprefix\doi{10.1016/J.OCEANENG.2022.110834}.
\bibitem[{Gorain and Mandal(2014)}]{Gorain2014}
\bibinfo{author}{Gorain, B.}, \bibinfo{author}{Mandal, P.S.}, \bibinfo{year}{2014}.
\newblock \bibinfo{title}{Approximation algorithms for sweep coverage in wireless sensor networks}.
\newblock \bibinfo{journal}{Journal of parallel and Distributed Computing} \bibinfo{volume}{74}, \bibinfo{pages}{2699--2707}.
\bibitem[{Grob(2006)}]{Grob2006}
\bibinfo{author}{Grob, M.J.}, \bibinfo{year}{2006}.
\newblock \bibinfo{title}{Routing of platforms in a maritime surface surveillance operation}.
\newblock \bibinfo{journal}{European Journal of Operational Research} \bibinfo{volume}{170}, \bibinfo{pages}{613--628}.
\newblock \DOIprefix\doi{10.1016/j.ejor.2004.02.029}.
\bibitem[{Huang and Onta{\~n}{\'o}n(2022)}]{Huang2020}
\bibinfo{author}{Huang, S.}, \bibinfo{author}{Onta{\~n}{\'o}n, S.}, \bibinfo{year}{2022}.
\newblock \bibinfo{title}{A closer look at invalid action masking in policy gradient algorithms}, in: \bibinfo{booktitle}{The International FLAIRS Conference Proceedings}, p. \bibinfo{pages}{n/a}.
\newblock \DOIprefix\doi{10.32473/flairs.v35i.130584}.
\bibitem[{Islam et~al.(2007)Islam, Meijer, Rodr{\'\i}guez, Rappaport and Xiao}]{islam2007hamilton}
\bibinfo{author}{Islam, K.}, \bibinfo{author}{Meijer, H.}, \bibinfo{author}{Rodr{\'\i}guez, Y.N.}, \bibinfo{author}{Rappaport, D.}, \bibinfo{author}{Xiao, H.}, \bibinfo{year}{2007}.
\newblock \bibinfo{title}{Hamilton circuits in hexagonal grid graphs.}, in: \bibinfo{booktitle}{CCCG}, pp. \bibinfo{pages}{85--88}.
\bibitem[{John et~al.(2001)John, Panton and White}]{John2001}
\bibinfo{author}{John, M.}, \bibinfo{author}{Panton, D.}, \bibinfo{author}{White, K.}, \bibinfo{year}{2001}.
\newblock \bibinfo{title}{Mission planning for regional surveillance}.
\newblock \bibinfo{journal}{Annals of Operations Research} \bibinfo{volume}{108}, \bibinfo{pages}{157--173}.
\bibitem[{Kadioglu et~al.(2019)Kadioglu, Urtis and Papanikolopoulos}]{Kadioglu2019}
\bibinfo{author}{Kadioglu, E.}, \bibinfo{author}{Urtis, C.}, \bibinfo{author}{Papanikolopoulos, N.}, \bibinfo{year}{2019}.
\newblock \bibinfo{title}{Uav coverage using hexagonal tessellation}, in: \bibinfo{booktitle}{27th Mediterranean Conference on Control and Automation, MED 2019 - Proceedings}, \bibinfo{publisher}{Institute of Electrical and Electronics Engineers Inc.}. pp. \bibinfo{pages}{37--42}.
\newblock \DOIprefix\doi{10.1109/MED.2019.8798564}.
\bibitem[{Kaelbling et~al.(1996)Kaelbling, Littman and Moore}]{Kaelbling1996}
\bibinfo{author}{Kaelbling, L.P.}, \bibinfo{author}{Littman, M.L.}, \bibinfo{author}{Moore, A.W.}, \bibinfo{year}{1996}.
\newblock \bibinfo{title}{Reinforcement learning: A survey}.
\newblock \bibinfo{journal}{Journal of Artificial Intelligence Research} \bibinfo{volume}{4}, \bibinfo{pages}{237--285}.
\newblock \DOIprefix\doi{10.1613/JAIR.301}.
\bibitem[{Karapetyan et~al.(2019)Karapetyan, Braude, Moulton, Burstein, White, O’Kane and Rekleitis}]{Karapetyan2019}
\bibinfo{author}{Karapetyan, N.}, \bibinfo{author}{Braude, A.}, \bibinfo{author}{Moulton, J.}, \bibinfo{author}{Burstein, J.A.}, \bibinfo{author}{White, S.}, \bibinfo{author}{O’Kane, J.M.}, \bibinfo{author}{Rekleitis, I.}, \bibinfo{year}{2019}.
\newblock \bibinfo{title}{Riverine coverage with an autonomous surface vehicle over known environments}, in: \bibinfo{booktitle}{2019 IEEE/RSJ International Conference on Intelligent Robots and Systems (IROS)}, \bibinfo{publisher}{IEEE}. pp. \bibinfo{pages}{3098--3104}.
\bibitem[{Karasakal(2016)}]{Karasakal2016}
\bibinfo{author}{Karasakal, O.}, \bibinfo{year}{2016}.
\newblock \bibinfo{title}{Minisum and maximin aerial surveillance over disjoint rectangles}.
\newblock \bibinfo{journal}{Top} \bibinfo{volume}{24}, \bibinfo{pages}{705--724}.
\newblock \DOIprefix\doi{10.1007/s11750-016-0416-1}.
\bibitem[{Kong et~al.(2016)Kong, Lin, Xie, Qiao, Jin, Zeng, Ren and Liu}]{Kong2016}
\bibinfo{author}{Kong, L.}, \bibinfo{author}{Lin, S.}, \bibinfo{author}{Xie, W.}, \bibinfo{author}{Qiao, X.}, \bibinfo{author}{Jin, X.}, \bibinfo{author}{Zeng, P.}, \bibinfo{author}{Ren, W.}, \bibinfo{author}{Liu, X.Y.}, \bibinfo{year}{2016}.
\newblock \bibinfo{title}{Adaptive barrier coverage using software defined sensor networks}.
\newblock \bibinfo{journal}{IEEE Sensors Journal} \bibinfo{volume}{16}, \bibinfo{pages}{7364--7372}.
\newblock \DOIprefix\doi{10.1109/JSEN.2016.2566808}.
\bibitem[{Kool et~al.(2018)Kool, Hoof and Welling}]{Kool2018}
\bibinfo{author}{Kool, W.}, \bibinfo{author}{Hoof, H.V.}, \bibinfo{author}{Welling, M.}, \bibinfo{year}{2018}.
\newblock \bibinfo{title}{Attention, learn to solve routing problems!}
\newblock \bibinfo{journal}{arXiv preprint arXiv:1803.08475} .
\bibitem[{Kumar and Kumar(2023)}]{Kumar2023}
\bibinfo{author}{Kumar, K.}, \bibinfo{author}{Kumar, N.}, \bibinfo{year}{2023}.
\newblock \bibinfo{title}{Region coverage-aware path planning for unmanned aerial vehicles: A systematic review}.
\newblock \bibinfo{journal}{Physical Communication} \bibinfo{volume}{59}, \bibinfo{pages}{102073}.
\newblock \DOIprefix\doi{10.1016/J.PHYCOM.2023.102073}.
\bibitem[{Kwon et~al.(2020)Kwon, Choo, Kim, Yoon, Gwon and Min}]{kwon2020pomo}
\bibinfo{author}{Kwon, Y.D.}, \bibinfo{author}{Choo, J.}, \bibinfo{author}{Kim, B.}, \bibinfo{author}{Yoon, I.}, \bibinfo{author}{Gwon, Y.}, \bibinfo{author}{Min, S.}, \bibinfo{year}{2020}.
\newblock \bibinfo{title}{Pomo: Policy optimization with multiple optima for reinforcement learning}.
\newblock \bibinfo{journal}{Advances in Neural Information Processing Systems} \bibinfo{volume}{33}, \bibinfo{pages}{21188--21198}.
\bibitem[{Li et~al.(2021a)Li, Ma, Gao, Cao, Lim, Song and Zhang}]{Li2021a}
\bibinfo{author}{Li, J.}, \bibinfo{author}{Ma, Y.}, \bibinfo{author}{Gao, R.}, \bibinfo{author}{Cao, Z.}, \bibinfo{author}{Lim, A.}, \bibinfo{author}{Song, W.}, \bibinfo{author}{Zhang, J.}, \bibinfo{year}{2021}a.
\newblock \bibinfo{title}{Deep reinforcement learning for solving the heterogeneous capacitated vehicle routing problem}.
\newblock \bibinfo{journal}{IEEE Transactions on Cybernetics} \bibinfo{volume}{54}, \bibinfo{pages}{13572--13585}.
\newblock \DOIprefix\doi{10.1109/TCYB.2021.3111082}.
\bibitem[{Li et~al.(2021b)Li, Xin, Cao, Lim, Song and Zhang}]{Li2021b}
\bibinfo{author}{Li, J.}, \bibinfo{author}{Xin, L.}, \bibinfo{author}{Cao, Z.}, \bibinfo{author}{Lim, A.}, \bibinfo{author}{Song, W.}, \bibinfo{author}{Zhang, J.}, \bibinfo{year}{2021}b.
\newblock \bibinfo{title}{Heterogeneous attentions for solving pickup and delivery problem via deep reinforcement learning}.
\newblock \bibinfo{journal}{IEEE TRANSACTIONS ON INTELLIGENT TRANSPORTATION SYSTEMS} .
\bibitem[{Li et~al.(2020)Li, Xiong, She and Wu}]{Li2020}
\bibinfo{author}{Li, J.}, \bibinfo{author}{Xiong, Y.}, \bibinfo{author}{She, J.}, \bibinfo{author}{Wu, M.}, \bibinfo{year}{2020}.
\newblock \bibinfo{title}{A path planning method for sweep coverage with multiple uavs}.
\newblock \bibinfo{journal}{IEEE Internet of Things Journal} \bibinfo{volume}{7}, \bibinfo{pages}{8967--8978}.
\newblock \DOIprefix\doi{10.1109/JIOT.2020.2999083}.
\bibitem[{Li and Chen(2022)}]{Li2022}
\bibinfo{author}{Li, L.}, \bibinfo{author}{Chen, H.}, \bibinfo{year}{2022}.
\newblock \bibinfo{title}{Uav enhanced target-barrier coverage algorithm for wireless sensor networks based on reinforcement learning}.
\newblock \bibinfo{journal}{Sensors} \bibinfo{volume}{22}, \bibinfo{pages}{6381}.
\newblock \DOIprefix\doi{10.3390/S22176381}.
\bibitem[{Li et~al.(2011)Li, Cheng, Liu, Liu, Li, Liao and Lab}]{Li2011}
\bibinfo{author}{Li, M.}, \bibinfo{author}{Cheng, W.}, \bibinfo{author}{Liu, K.}, \bibinfo{author}{Liu, Y.}, \bibinfo{author}{Li, X.}, \bibinfo{author}{Liao, X.}, \bibinfo{author}{Lab, W.J.}, \bibinfo{year}{2011}.
\newblock \bibinfo{title}{Sweep coverage with mobile sensors}.
\newblock \bibinfo{journal}{IEEE Transactions on Mobile Computing} \bibinfo{volume}{10}, \bibinfo{pages}{1534--1545}.
\bibitem[{Li et~al.(2019)Li, Shen, Huang and Guo}]{Li2019}
\bibinfo{author}{Li, S.}, \bibinfo{author}{Shen, H.}, \bibinfo{author}{Huang, Q.}, \bibinfo{author}{Guo, L.}, \bibinfo{year}{2019}.
\newblock \bibinfo{title}{Optimizing the sensor movement for barrier coverage in a sink-based deployed mobile sensor network}.
\newblock \bibinfo{journal}{IEEE Access} \bibinfo{volume}{7}, \bibinfo{pages}{156301--156314}.
\newblock \DOIprefix\doi{10.1109/ACCESS.2019.2949025}.
\bibitem[{Li and Savkin(2021)}]{LiSavkin2021}
\bibinfo{author}{Li, X.}, \bibinfo{author}{Savkin, A.V.}, \bibinfo{year}{2021}.
\newblock \bibinfo{title}{Networked unmanned aerial vehicles for surveillance and monitoring: A survey}.
\newblock \bibinfo{journal}{Future Internet} \bibinfo{volume}{13}.
\newblock \DOIprefix\doi{10.3390/FI13070174}.
\bibitem[{Liu and Liu(2021)}]{liu2021agent}
\bibinfo{author}{Liu, Z.}, \bibinfo{author}{Liu, Y.}, \bibinfo{year}{2021}.
\newblock \bibinfo{title}{Agent-based simulation of multi-uav search-track for dynamic targets in sweep coverage}, in: \bibinfo{booktitle}{Journal of Physics: Conference Series}, \bibinfo{organization}{IOP Publishing}. p. \bibinfo{pages}{012033}.
\bibitem[{Luis et~al.(2021)Luis, Reina and Marin}]{Luis2021b}
\bibinfo{author}{Luis, S.Y.}, \bibinfo{author}{Reina, D.G.}, \bibinfo{author}{Marin, S.L.}, \bibinfo{year}{2021}.
\newblock \bibinfo{title}{A multiagent deep reinforcement learning approach for path planning in autonomous surface vehicles: The ypacaraí lake patrolling case}.
\newblock \bibinfo{journal}{IEEE Access} \bibinfo{volume}{9}, \bibinfo{pages}{17084--17099}.
\newblock \DOIprefix\doi{10.1109/ACCESS.2021.3053348}.
\bibitem[{Luis et~al.(2020)Luis, Reina and Marín}]{Yanez2020}
\bibinfo{author}{Luis, S.Y.}, \bibinfo{author}{Reina, D.G.}, \bibinfo{author}{Marín, S.L.T.}, \bibinfo{year}{2020}.
\newblock \bibinfo{title}{A deep reinforcement learning approach for the patrolling problem of water resources through autonomous surface vehicles: The ypacarai lake case}.
\newblock \bibinfo{journal}{IEEE Access} \bibinfo{volume}{8}, \bibinfo{pages}{204076--204093}.
\newblock \DOIprefix\doi{10.1109/ACCESS.2020.3036938}.
\bibitem[{Ma et~al.(2026)Ma, Peng, Gu, Sun, Yang and Wang}]{ma2026fault}
\bibinfo{author}{Ma, G.}, \bibinfo{author}{Peng, Z.}, \bibinfo{author}{Gu, N.}, \bibinfo{author}{Sun, G.}, \bibinfo{author}{Yang, M.}, \bibinfo{author}{Wang, D.}, \bibinfo{year}{2026}.
\newblock \bibinfo{title}{Fault-tolerant coverage path planning for multiple unmanned surface vehicles via cooperative game and deterministic finite state machines}.
\newblock \bibinfo{journal}{Ocean Engineering} \bibinfo{volume}{345}, \bibinfo{pages}{123579}.
\newblock \DOIprefix\doi{10.1016/j.oceaneng.2025.123579}.
\bibitem[{Mier et~al.(2023)Mier, Valente and Bruin}]{Mier2023}
\bibinfo{author}{Mier, G.}, \bibinfo{author}{Valente, J.}, \bibinfo{author}{Bruin, S.D.}, \bibinfo{year}{2023}.
\newblock \bibinfo{title}{Fields2cover: An open-source coverage path planning library for unmanned agricultural vehicles}.
\newblock \bibinfo{journal}{IEEE Robotics and Automation Letters} \bibinfo{volume}{8}, \bibinfo{pages}{2166--2172}.
\bibitem[{Nazari et~al.(2018)Nazari, Oroojlooy, Takáč and Snyder}]{Nazari2018}
\bibinfo{author}{Nazari, M.}, \bibinfo{author}{Oroojlooy, A.}, \bibinfo{author}{Takáč, M.}, \bibinfo{author}{Snyder, L.V.}, \bibinfo{year}{2018}.
\newblock \bibinfo{title}{Reinforcement learning for solving the vehicle routing problem}.
\newblock \bibinfo{journal}{Advances in neural information processing systems} \bibinfo{volume}{31}.
\bibitem[{Nguyen and So-In(2018)}]{Nguyen2018}
\bibinfo{author}{Nguyen, T.G.}, \bibinfo{author}{So-In, C.}, \bibinfo{year}{2018}.
\newblock \bibinfo{title}{Distributed deployment algorithm for barrier coverage in mobile sensor networks}.
\newblock \bibinfo{journal}{IEEE Access} \bibinfo{volume}{6}, \bibinfo{pages}{21042--21052}.
\newblock \DOIprefix\doi{10.1109/ACCESS.2018.2822263}.
\bibitem[{Nielsen et~al.(2019)Nielsen, Sung and Nielsen}]{Nielsen2019}
\bibinfo{author}{Nielsen, L.D.}, \bibinfo{author}{Sung, I.}, \bibinfo{author}{Nielsen, P.}, \bibinfo{year}{2019}.
\newblock \bibinfo{title}{Convex decomposition for a coverage path planning for autonomous vehicles: Interior extension of edges}.
\newblock \bibinfo{journal}{Sensors} \bibinfo{volume}{19}.
\newblock \DOIprefix\doi{10.3390/s19194165}.
\bibitem[{Nigam et~al.(2009)Nigam, Bieniawski, Kroo and Vian}]{Nigam2009}
\bibinfo{author}{Nigam, N.}, \bibinfo{author}{Bieniawski, S.}, \bibinfo{author}{Kroo, I.}, \bibinfo{author}{Vian, J.}, \bibinfo{year}{2009}.
\newblock \bibinfo{title}{Control of multiple uavs for persistent surveillance: Algorithm description and hardware demonstration}.
\newblock \bibinfo{journal}{IEEE Transactions on Control Systems Technology} \bibinfo{volume}{20}, \bibinfo{pages}{1236--1251}.
\bibitem[{Otto et~al.(2018)Otto, Agatz, Campbell, Golden and Pesch}]{Otto2018}
\bibinfo{author}{Otto, A.}, \bibinfo{author}{Agatz, N.}, \bibinfo{author}{Campbell, J.}, \bibinfo{author}{Golden, B.}, \bibinfo{author}{Pesch, E.}, \bibinfo{year}{2018}.
\newblock \bibinfo{title}{Optimization approaches for civil applications of unmanned aerial vehicles (uavs) or aerial drones: A survey}.
\newblock \bibinfo{journal}{Networks} \bibinfo{volume}{72}, \bibinfo{pages}{411--458}.
\newblock \DOIprefix\doi{10.1002/NET.21818}.
\bibitem[{Panton and Elbers(1999)}]{Panton1999}
\bibinfo{author}{Panton, D.M.}, \bibinfo{author}{Elbers, A.W.}, \bibinfo{year}{1999}.
\newblock \bibinfo{title}{Mission planning for synthetic aperture radar surveillance}.
\newblock \bibinfo{journal}{Interfaces} \bibinfo{volume}{29}, \bibinfo{pages}{73--88}.
\newblock \DOIprefix\doi{10.1287/inte.29.2.73}.
\bibitem[{Raza et~al.(2022)Raza, Sajid and Singh}]{Raza2022}
\bibinfo{author}{Raza, S.M.}, \bibinfo{author}{Sajid, M.}, \bibinfo{author}{Singh, J.}, \bibinfo{year}{2022}.
\newblock \bibinfo{title}{Vehicle routing problem using reinforcement learning: Recent advancements}, in: \bibinfo{booktitle}{Lecture Notes in Electrical Engineering}, \bibinfo{publisher}{Springer Science and Business Media Deutschland GmbH}. pp. \bibinfo{pages}{269--280}.
\newblock \DOIprefix\doi{10.1007/978-981-19-0840-8_20}.
\bibitem[{Savkin and Huang(2019)}]{Savkin2019}
\bibinfo{author}{Savkin, A.V.}, \bibinfo{author}{Huang, H.}, \bibinfo{year}{2019}.
\newblock \bibinfo{title}{Proactive deployment of aerial drones for coverage over very uneven terrains: A version of the 3d art gallery problem}.
\newblock \bibinfo{journal}{Sensors} \bibinfo{volume}{19}, \bibinfo{pages}{1438}.
\newblock \DOIprefix\doi{10.3390/S19061438}.
\bibitem[{Schulman et~al.(2017)Schulman, Wolski, Dhariwal, Radford and Klimov}]{Schulman2017}
\bibinfo{author}{Schulman, J.}, \bibinfo{author}{Wolski, F.}, \bibinfo{author}{Dhariwal, P.}, \bibinfo{author}{Radford, A.}, \bibinfo{author}{Klimov, O.}, \bibinfo{year}{2017}.
\newblock \bibinfo{title}{Proximal policy optimization algorithms}.
\newblock \bibinfo{journal}{arXiv preprint arXiv:1707.06347} .
\bibitem[{Shao et~al.(2024)Shao, Wang, Zhu, Xu, Song, Bi, Zhang, Zhang, Li et~al.}]{shao2024deepseekmath}
\bibinfo{author}{Shao, Z.}, \bibinfo{author}{Wang, P.}, \bibinfo{author}{Zhu, Q.}, \bibinfo{author}{Xu, R.}, \bibinfo{author}{Song, J.}, \bibinfo{author}{Bi, X.}, \bibinfo{author}{Zhang, H.}, \bibinfo{author}{Zhang, M.}, \bibinfo{author}{Li, Y.}, et~al., \bibinfo{year}{2024}.
\newblock \bibinfo{title}{Deepseekmath: Pushing the limits of mathematical reasoning in open language models}.
\newblock \bibinfo{journal}{arXiv preprint arXiv:2402.03300} .
\bibitem[{Shen et~al.(2025)Shen, Zhu, Bai, Deng, Xue, Cao, Mu and Qin}]{shen2025multiple}
\bibinfo{author}{Shen, J.}, \bibinfo{author}{Zhu, Z.}, \bibinfo{author}{Bai, G.}, \bibinfo{author}{Deng, Z.}, \bibinfo{author}{Xue, Y.}, \bibinfo{author}{Cao, X.}, \bibinfo{author}{Mu, X.}, \bibinfo{author}{Qin, H.}, \bibinfo{year}{2025}.
\newblock \bibinfo{title}{Multiple unmanned sailboats cooperative coverage: Task allocation and path planning in island environments}.
\newblock \bibinfo{journal}{Ocean Engineering} \bibinfo{volume}{339}, \bibinfo{pages}{122172}.
\newblock \DOIprefix\doi{10.1016/j.oceaneng.2025.122172}.
\bibitem[{Siew et~al.(2022)Siew, Jang, Roberts, Linares and Fletcher}]{siew2022cislunar}
\bibinfo{author}{Siew, P.M.}, \bibinfo{author}{Jang, D.}, \bibinfo{author}{Roberts, T.G.}, \bibinfo{author}{Linares, R.}, \bibinfo{author}{Fletcher, J.}, \bibinfo{year}{2022}.
\newblock \bibinfo{title}{Cislunar space situational awareness sensor tasking using deep reinforcement learning agents}, in: \bibinfo{booktitle}{2022 Advanced Maui Optical and Space Surveillance Technologies Conference (AMOS), Maui, Hawaii}, p. \bibinfo{pages}{n/a}.
\bibitem[{Siew and Linares(2022)}]{Siew2022}
\bibinfo{author}{Siew, P.M.}, \bibinfo{author}{Linares, R.}, \bibinfo{year}{2022}.
\newblock \bibinfo{title}{Optimal tasking of ground-based sensors for space situational awareness using deep reinforcement learning}.
\newblock \bibinfo{journal}{Sensors} \bibinfo{volume}{22}.
\newblock \DOIprefix\doi{10.3390/s22207847}.
\bibitem[{Soldi et~al.(2021)Soldi, Gaglione, Forti, Simone, Daffinà, Bottini, Quattrociocchi, Millefiori, Braca, Carniel, Willett, Iodice, Riccio and Farina}]{Soldi2021}
\bibinfo{author}{Soldi, G.}, \bibinfo{author}{Gaglione, D.}, \bibinfo{author}{Forti, N.}, \bibinfo{author}{Simone, A.D.}, \bibinfo{author}{Daffinà, F.C.}, \bibinfo{author}{Bottini, G.}, \bibinfo{author}{Quattrociocchi, D.}, \bibinfo{author}{Millefiori, L.M.}, \bibinfo{author}{Braca, P.}, \bibinfo{author}{Carniel, S.}, \bibinfo{author}{Willett, P.}, \bibinfo{author}{Iodice, A.}, \bibinfo{author}{Riccio, D.}, \bibinfo{author}{Farina, A.}, \bibinfo{year}{2021}.
\newblock \bibinfo{title}{Space-based global maritime surveillance. part i: Satellite technologies}.
\newblock \bibinfo{journal}{IEEE Aerospace and Electronic Systems Magazine} \bibinfo{volume}{36}, \bibinfo{pages}{8--28}.
\bibitem[{Sutton and Barto(2018)}]{Sutton2018}
\bibinfo{author}{Sutton, R.S.}, \bibinfo{author}{Barto, A.G.}, \bibinfo{year}{2018}.
\newblock \bibinfo{title}{Reinforcement learning : an introduction}.
\newblock \bibinfo{edition}{second edition} ed., \bibinfo{publisher}{MIT press}.
\bibitem[{Tan et~al.(2021)Tan, Mohd-Mokhtar and Arshad}]{TanChee2021}
\bibinfo{author}{Tan, C.S.}, \bibinfo{author}{Mohd-Mokhtar, R.}, \bibinfo{author}{Arshad, M.R.}, \bibinfo{year}{2021}.
\newblock \bibinfo{title}{A comprehensive review of coverage path planning in robotics using classical and heuristic algorithms}.
\newblock \bibinfo{journal}{IEEE Access} \bibinfo{volume}{9}, \bibinfo{pages}{119310--119342}.
\newblock \DOIprefix\doi{10.1109/ACCESS.2021.3108177}.
\bibitem[{Tong et~al.(2020)Tong, Zhou, Zeng, Chen and Shahabi}]{Tong2020}
\bibinfo{author}{Tong, Y.}, \bibinfo{author}{Zhou, Z.}, \bibinfo{author}{Zeng, Y.}, \bibinfo{author}{Chen, L.}, \bibinfo{author}{Shahabi, C.}, \bibinfo{year}{2020}.
\newblock \bibinfo{title}{Spatial crowdsourcing: a survey}.
\newblock \bibinfo{journal}{VLDB Journal} \bibinfo{volume}{29}, \bibinfo{pages}{217--250}.
\newblock \DOIprefix\doi{10.1007/S00778-019-00568-7/METRICS}.
\bibitem[{{United Nations Trade and Development (UNCTAD)}(2024)}]{united2024review}
\bibinfo{author}{{United Nations Trade and Development (UNCTAD)}}, \bibinfo{year}{2024}.
\newblock \bibinfo{title}{Review of maritime transport 2024: Navigating maritime chokepoints}.
\newblock \bibinfo{publisher}{Stylus Publishing, LLC}.
\bibitem[{Vinyals et~al.(2015)Vinyals, Fortunato and Jaitly}]{Vinyals2015}
\bibinfo{author}{Vinyals, O.}, \bibinfo{author}{Fortunato, M.}, \bibinfo{author}{Jaitly, N.}, \bibinfo{year}{2015}.
\newblock \bibinfo{title}{Pointer networks}.
\newblock \bibinfo{journal}{Advances in neural information processing systems} \bibinfo{volume}{28}.
\bibitem[{Wu et~al.(2024)Wu, Cheng, Chu and Song}]{wu2024autonomous}
\bibinfo{author}{Wu, J.}, \bibinfo{author}{Cheng, L.}, \bibinfo{author}{Chu, S.}, \bibinfo{author}{Song, Y.}, \bibinfo{year}{2024}.
\newblock \bibinfo{title}{An autonomous coverage path planning algorithm for maritime search and rescue of persons-in-water based on deep reinforcement learning}.
\newblock \bibinfo{journal}{Ocean engineering} \bibinfo{volume}{291}, \bibinfo{pages}{116403}.
\bibitem[{Wu et~al.(2019)Wu, Xiong, Wu, He and She}]{Wu2019}
\bibinfo{author}{Wu, L.}, \bibinfo{author}{Xiong, Y.}, \bibinfo{author}{Wu, M.}, \bibinfo{author}{He, Y.}, \bibinfo{author}{She, J.}, \bibinfo{year}{2019}.
\newblock \bibinfo{title}{A task assignment method for sweep coverage optimization based on crowdsensing}.
\newblock \bibinfo{journal}{IEEE Internet of Things Journal} \bibinfo{volume}{6}, \bibinfo{pages}{10686--10699}.
\newblock \DOIprefix\doi{10.1109/JIOT.2019.2940717}.
\bibitem[{Xin et~al.(2021)Xin, Song, Cao and Zhang}]{Xin2021}
\bibinfo{author}{Xin, L.}, \bibinfo{author}{Song, W.}, \bibinfo{author}{Cao, Z.}, \bibinfo{author}{Zhang, J.}, \bibinfo{year}{2021}.
\newblock \bibinfo{title}{Multi-decoder attention model with embedding glimpse for solving vehicle routing problems}, in: \bibinfo{booktitle}{Proceedings of the AAAI Conference on Artificial Intelligence}, pp. \bibinfo{pages}{12042--12049}.
\bibitem[{Zelenka et~al.(2020)Zelenka, Kasanický, Bundzel and Andoga}]{Zelenka2020}
\bibinfo{author}{Zelenka, J.}, \bibinfo{author}{Kasanický, T.}, \bibinfo{author}{Bundzel, M.}, \bibinfo{author}{Andoga, R.}, \bibinfo{year}{2020}.
\newblock \bibinfo{title}{Self-adaptation of a heterogeneous swarm of mobile robots to a covered area}.
\newblock \bibinfo{journal}{Applied Sciences (Switzerland)} \bibinfo{volume}{10}.
\newblock \DOIprefix\doi{10.3390/app10103562}.
\bibitem[{Zhao and Bai(2024)}]{zhao2024joint}
\bibinfo{author}{Zhao, L.}, \bibinfo{author}{Bai, Y.}, \bibinfo{year}{2024}.
\newblock \bibinfo{title}{Joint-optimized coverage path planning framework for {USV}-assisted offshore bathymetric mapping: From theory to practice}.
\newblock \bibinfo{journal}{Knowledge-Based Systems} \bibinfo{volume}{304}, \bibinfo{pages}{112449}.
\newblock \DOIprefix\doi{10.1016/j.knosys.2024.112449}.
\bibitem[{Zhao et~al.(2024)Zhao, Bai and Paik}]{zhao2024optimal}
\bibinfo{author}{Zhao, L.}, \bibinfo{author}{Bai, Y.}, \bibinfo{author}{Paik, J.K.}, \bibinfo{year}{2024}.
\newblock \bibinfo{title}{Optimal coverage path planning for {USV}-assisted coastal bathymetric survey: Models, solutions, and lake trials}.
\newblock \bibinfo{journal}{Ocean Engineering} \bibinfo{volume}{296}, \bibinfo{pages}{116921}.
\newblock \DOIprefix\doi{10.1016/j.oceaneng.2024.116921}.
\bibitem[{Zhou et~al.(2019)Zhou, Liu, Feng, Zhang, Mumtaz and Rodriguez}]{ZhouZhen2019}
\bibinfo{author}{Zhou, Z.}, \bibinfo{author}{Liu, P.}, \bibinfo{author}{Feng, J.}, \bibinfo{author}{Zhang, Y.}, \bibinfo{author}{Mumtaz, S.}, \bibinfo{author}{Rodriguez, J.}, \bibinfo{year}{2019}.
\newblock \bibinfo{title}{Computation resource allocation and task assignment optimization in vehicular fog computing: A contract-matching approach}.
\newblock \bibinfo{journal}{IEEE Transactions on Vehicular Technology} \bibinfo{volume}{68}, \bibinfo{pages}{3113--3125}.
\newblock \DOIprefix\doi{10.1109/TVT.2019.2894851}.
\bibitem[{Zuo et~al.(2020)Zuo, Tharmarasa, Jassemi-Zargani, Kashyap, Thiyagalingam and Kirubarajan}]{Zuo2020}
\bibinfo{author}{Zuo, Y.}, \bibinfo{author}{Tharmarasa, R.}, \bibinfo{author}{Jassemi-Zargani, R.}, \bibinfo{author}{Kashyap, N.}, \bibinfo{author}{Thiyagalingam, J.}, \bibinfo{author}{Kirubarajan, T.T.}, \bibinfo{year}{2020}.
\newblock \bibinfo{title}{Milp formulation for aircraft path planning in persistent surveillance}.
\newblock \bibinfo{journal}{IEEE Transactions on Aerospace and Electronic Systems} \bibinfo{volume}{56}, \bibinfo{pages}{3796--3811}.
\newblock \DOIprefix\doi{10.1109/TAES.2020.2983532}.
\bibitem[{Zweig et~al.(2020)Zweig, Ahmed, Willke and Ma}]{zweig2020neural}
\bibinfo{author}{Zweig, A.}, \bibinfo{author}{Ahmed, N.}, \bibinfo{author}{Willke, T.L.}, \bibinfo{author}{Ma, G.}, \bibinfo{year}{2020}.
\newblock \bibinfo{title}{Neural algorithms for graph navigation}, in: \bibinfo{booktitle}{Learning Meets Combinatorial Algorithms at NeurIPS2020}, p. \bibinfo{pages}{n/a}.

\end{thebibliography}

\end{document}